\pdfoutput=1

\documentclass[11pt]{article}

\usepackage[]{acl}

\usepackage{times}
\usepackage{latexsym}

\usepackage[T1]{fontenc}

\usepackage[utf8]{inputenc}

\usepackage{comment}
\usepackage{microtype}
\usepackage[linewidth=1pt]{mdframed}
\usepackage{tikz-dependency}
\usepackage{amsmath,amsfonts,bm}
\usepackage{graphicx}
\usepackage{algorithmicx}
\usepackage[noend]{algpseudocode}
\usepackage{booktabs}
\usepackage[percent]{overpic}
\usepackage{subcaption}
\usepackage{siunitx}
\usepackage{wrapfig}
\usepackage{listings}

\usepackage{enumitem}

\usepackage{multirow}
\usepackage[page]{appendix}

\usepackage{verbatim}
\usepackage{hyperref}
\usepackage{url}
\definecolor{human}{RGB}{213,235,252}
\definecolor{bot}{RGB}{224,224,224}

\NewEnviron{elaboration}{
\begin{tikzpicture}
\node[rectangle,minimum width=\textwidth,fill=bot] (m) {\begin{minipage}{.98\textwidth}\BODY\end{minipage}};
\draw[rounded corners, dotted] (m.south west) rectangle (m.north east);
\end{tikzpicture}
}
\NewEnviron{elaborationr}{
\begin{tikzpicture}
\node[rectangle,minimum width=\textwidth,fill=human] (m) {
\raggedleft
\begin{minipage}{.98\textwidth}
\BODY
\end{minipage}};
\draw[rounded corners, dotted] (m.south west) rectangle (m.north east);
\end{tikzpicture}
}

\usepackage[export]{adjustbox}
\usepackage{tabularx}
\usepackage{printlen}

\title{Situated Dialogue Learning through Procedural Environment Generation}
 \author{Prithviraj Ammanabrolu$^{\dagger}$$^{\ddagger}$ \And Renee Jia$^{\dagger}$  \\ $^{\dagger}$Georgia Institute of Technology \hspace{3em} $^{\ddagger}$Allen Institute for AI \\
         \texttt{raja@allenai.org} \And Mark O. Riedl$^{\dagger}$
         }

\begin{document}
\maketitle
\begin{abstract}
We teach goal-driven agents to interactively act and speak in situated environments by training on generated curriculums.
Our agents operate in LIGHT~\citep{Urbanek2019}---a large-scale crowd-sourced fantasy text adventure game wherein an agent perceives and interacts with the world through textual natural language.
Goals in this environment take the form of character-based quests, consisting of personas and motivations.
We augment LIGHT by learning to procedurally generate additional novel textual worlds and quests to create a curriculum of steadily increasing difficulty for training agents to achieve such goals.
In particular, we measure curriculum difficulty in terms of the rarity of the quest in the original training distribution---an easier environment is one that is more likely to have been found in the unaugmented dataset.
An ablation study shows that this method of learning from the tail of a distribution results in significantly higher generalization abilities as measured by zero-shot performance on never-before-seen quests.
\end{abstract}

\section{Introduction}
\begin{figure}[!h]
    \centering
    \includegraphics[width=.75\linewidth]{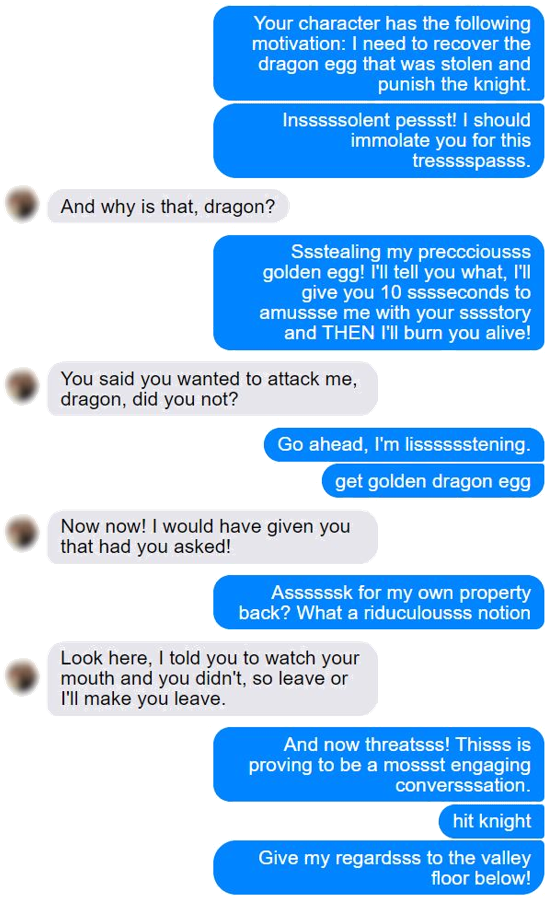}
    \caption{The LIGHT questing environment presented as a 2 player game deployed in Messenger.}
    \label{fig:lightquest}
\end{figure}
A key hypothesis in the pursuit towards creating goal-driven natural language-based agents posits that interactivity and environment grounding is critical for effective language learning~\citep{Barsalou2008,bisk-etal-2020-experience,ammanabrolu2021situated}.
Text games provide a platform on which to interactively train agents that can both act and speak in a situated manner---producing language that is both goal-driven and contextually relevant.
Agents in text games operate---perceiving, acting in, and speaking to others in a world---entirely using textual natural language.
These games are structured generally as sequential decision making problems in the form of puzzles or quests that must be completed to advance in the game.

As seen in Figure~\ref{fig:lightquest}, we focus on creating agents in LIGHT~\citep{Urbanek2019}, a large-scale crowdsourced fantasy text-adventure game, consisting of rich textual worlds---locations, objects, and characters with personas, and quests---motivations for each character.
To complete these quests, an agent must: 
(1) maintain character via its persona; and
(2) reason in a {\em partially observable} world about potential actions and utterances based on incomplete descriptions of the locations, objects, and other characters.
This requires several human like competencies such as commonsense reasoning, dynamic natural language understanding, and operating in combinatorially sized language-based state-action spaces.
Although recent work has provided evidence showing that interactive language learning via reinforcement learning (RL) in text games can be significantly more sample efficient than static supervised learning~\citep{Ammanabrolu2021} when creating goal-driven natural language agents, their ability to robustly generalize to novel scenarios is limited.

In sequential decision making problems in particular, this generalization gap is the result of an agent simply memorizing trajectories, e.g. the sequence of actions and dialogues required to finish a game, and thus being unable to react in novel scenarios---i.e. the agent learns from the head the training data and simply memorizes the long tail.
One way of decreasing this generalization gap is by training agents on procedurally generated environments---wherein the agent learns a family of parametrized tasks with a significantly larger state-action spaces than singular environments, thus effectively making the memorization of trajectories impossible~\citep{Justesen2018,Cobbe2020}.
Drawing inspiration from all of these ideas, we create a method that {\em learns} to create a training curriculum of increasingly more difficult novel procedurally generated environments.

Our contributions are threefold: 
(1) We present a method of parametrizing and generating a curriculum of environments in text games;
(2) We show how to effectively train reinforcement learning agents on this curriculum;
and (3) Provide an experimental study showing that our method enables significantly better generalization than those training on singular environments.

\begin{figure*}
    \centering
    \includegraphics[width=.9\linewidth]{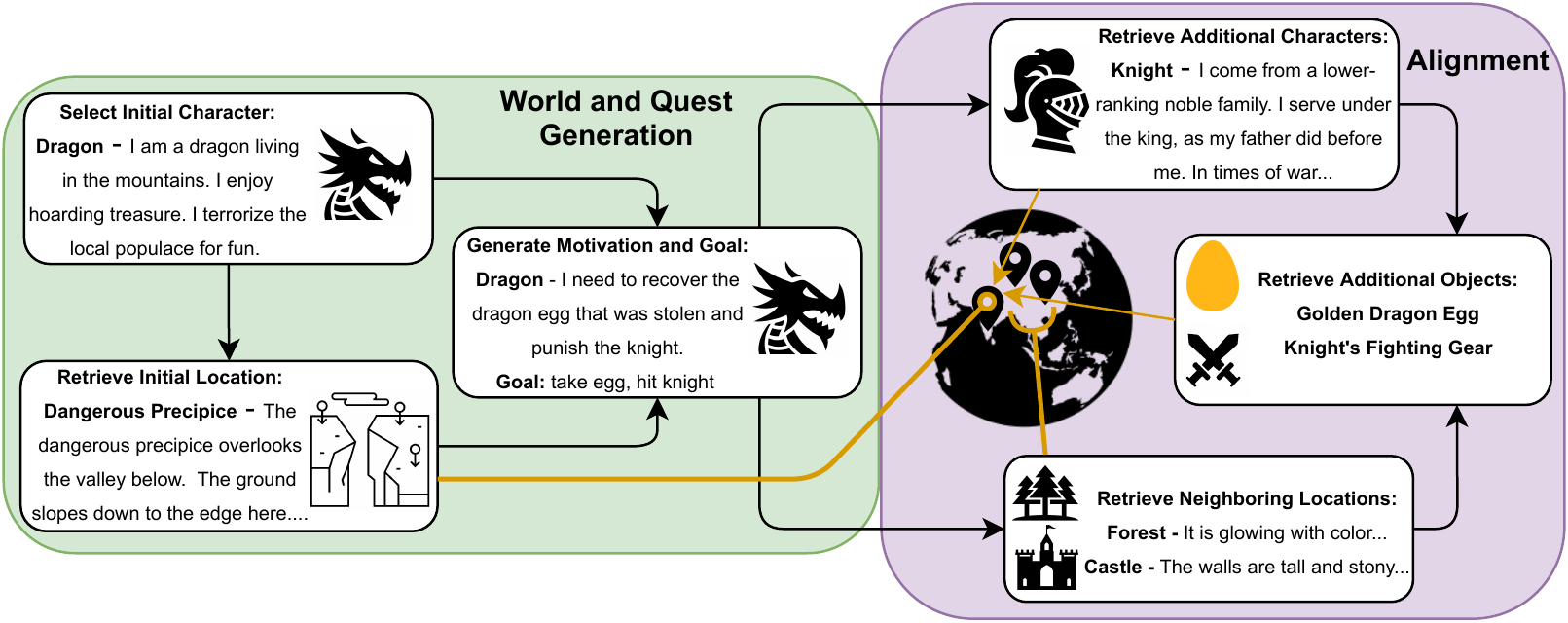}
    \caption{Procedural environment generation pipeline. Black lines indicate conditioning on all prior components. Gold lines indicate (adjacent) location placement.}
    \label{fig:pcgcurriculum}
\end{figure*}

\section{Procedural Environment Generation}
\label{sec:lightprocgen}
This section describes our procedural generation pipeline as seen in Figure~\ref{fig:pcgcurriculum}, starting with world and quest generation, followed by aligning both of them.
There are two main kinds of models that we use for the different modules in this pipeline: retrieval and generative.

\textbf{The LIGHT Questing Environment.}
The LIGHT game environment~\citep{Urbanek2019}\footnote{\url{https://parl.ai/projects/light}} is a multi-user fantasy text-adventure game consisting of a rich, diverse set of 1775 characters, 663 locations, and 3462 objects.
Characters are able to perform templated actions to interact with both objects and characters, and can speak to other characters through free form text dialogues.
Actions in text games generally consist of verb phrases (VP) followed optionally by prepositional phrases (VP PP).
For example, {\em get OBJ, put OBJ, give OBJ to CHAR}, etc..
These actions change the state of the world which is expressed through text descriptions.

Quests in LIGHT~\citep{Ammanabrolu2021} take the form of a short motivation and goal action that is required reach the world state required to finish the game.
For example, if the short motivation is {\em ``Your motivation is to acquire a sword''}, then the corresponding goal state would be for the character to have a sword in their inventory and goal action would be {\em get sword}.
This environment also contains a set of human expert demonstration of people speaking and acting in character while playing the quests mentioned above. 
Further details are found in Appendix~\ref{app:lightdetails}.

\subsection{World and Quest Creation}

\textbf{World Retrieval.} The first step of the pipeline involves choosing an initial character who will perform the quest.
For this, we uniformly randomly sample from the set of characters found in the LIGHT-Quest training set.
The corresponding character information includes a name and a description of the character's persona.
Given this character information, we further retrieve the location where the character is most likely to be found.

Retrieval models are trained to return the most highly correlated output for a given input in the dataset.
For example, a retrieval model can be asked to return the most likely character that can be found at a particular location.
These models compare a human annotated gold standard label with negative candidates drawn from the dataset.
The negative candidates provide noise that the model must filter out in order to learn representations that let it best predict the gold label.
These models are trained via a ranking loss that maximizes the scores of the gold label while simultaneously minimizing negative candidate score.
At test time, the highest ranked candidate based on the score is selected as the model prediction.

Specifically, we use a retrieval-based ranker model that checks for similarity of StarSpace~\citep{Wu2018} embeddings.
Our choice of model is influenced by \citet{Fan2019} who report state-of-the-art retrieval performance for locations in LIGHT using this model.
The overall ranker model first trains a randomly initialized StarSpace embedding model that is designed to correlate characters with the locations they are found in.
It learns a single bag-of-words embedding that takes into account all the individual words contained within the input---encoding character and location information as well as the previously mentioned negative retrieval candidates.
The rest of the training is similar to other retrieval models described earlier.
The retrieved location information consists of a location name as well as a description of the location.

\textbf{Quest Generation.} The quest is now generated using the existing character and location information.
The generation-based models used in this pipeline are trained to return the most likely output sequence given an input sequence.
Given a target sequence $Y=\{y_1,...,y_M\}$ and some input context vector via the encoders $\mathbf{X}$.
These models use autoregressive decoding techniques that factor the distribution over the target sequence into a chain of conditional probabilities with a causal left to right structure as $P(Y|\mathbf{X};\theta) =\prod_{i=1}^{M+1}p(y_i|y_{0:i-1},\mathbf{X};\theta)$ where $\theta$ represents the current network parameters.
At test time, a special start-of-sequence token is provided to the model which then proceeds to decode the rest of the output sequence using beam search.

We train two BART~\citep{lewis-etal-2020-bart} models that encodes input information via a bidirectional transformer encoder and decodes autoregressively: the first takes as input character and location information and produces a short motivation (Section~\ref{sec:lightprocgen}); the second takes as input character, location information, short motivation and produces the sequence of LIGHT game engine executable actions needed to achieve the motivation.
This sequence of actions is provided by the human expert demonstrations as mentioned in Section~\ref{sec:lightprocgen}.

\subsection{Aligning Worlds and Quests}
At this stage, the environment contains a motivated main character to perform a quest and a location for them to start in.
We now focus on aligning the world with the quest to ensure that the quest is playable and achievable.
Intuitively, to ensure that a quest is achievable, the world needs to contain all of the entities---locations, characters, and objects---mentioned within the quest.

To this end, the alignment process involves training three BERT-based~\citep{Devlin2018} biencoder retrieval models to retrieve the most likely characters, locations, and objects required flesh the environment out and make the quest achievable.
We use the same biencoder architecture proposed by \citet{Urbanek2019} which encodes context using one transformer and candidates with another---scoring candidates via inner product between the two encoded vectors.
The character retrieval model is conditioned on the initial character, quest, and location---producing additional characters required to complete the world.

We follow the setup in \citet{Ammanabrolu2021} and restrict worlds to only contains 2 characters at maximum but note that this method is extendable to greater numbers of characters.
Similarly, the location retrieval model is also conditioned on the same things---producing, in this case, 4 neighbors to the initial location (resulting in worlds that are 5 locations large).
These locations are connected to the initial location and a character can move between them by using commands such as {\em go west}, {\em go up} etc..
Once these characters and locations are added to the world, the object retrieval model predicts the set of objects that are required to be distributed for each location given all the character information present in it.
The final game environment instance is complete once this object set has been added.

\section{Curriculum Learning}
\label{sec:curriculumgen}



\begin{figure}
    \centering
    \includegraphics[width=.7\linewidth]{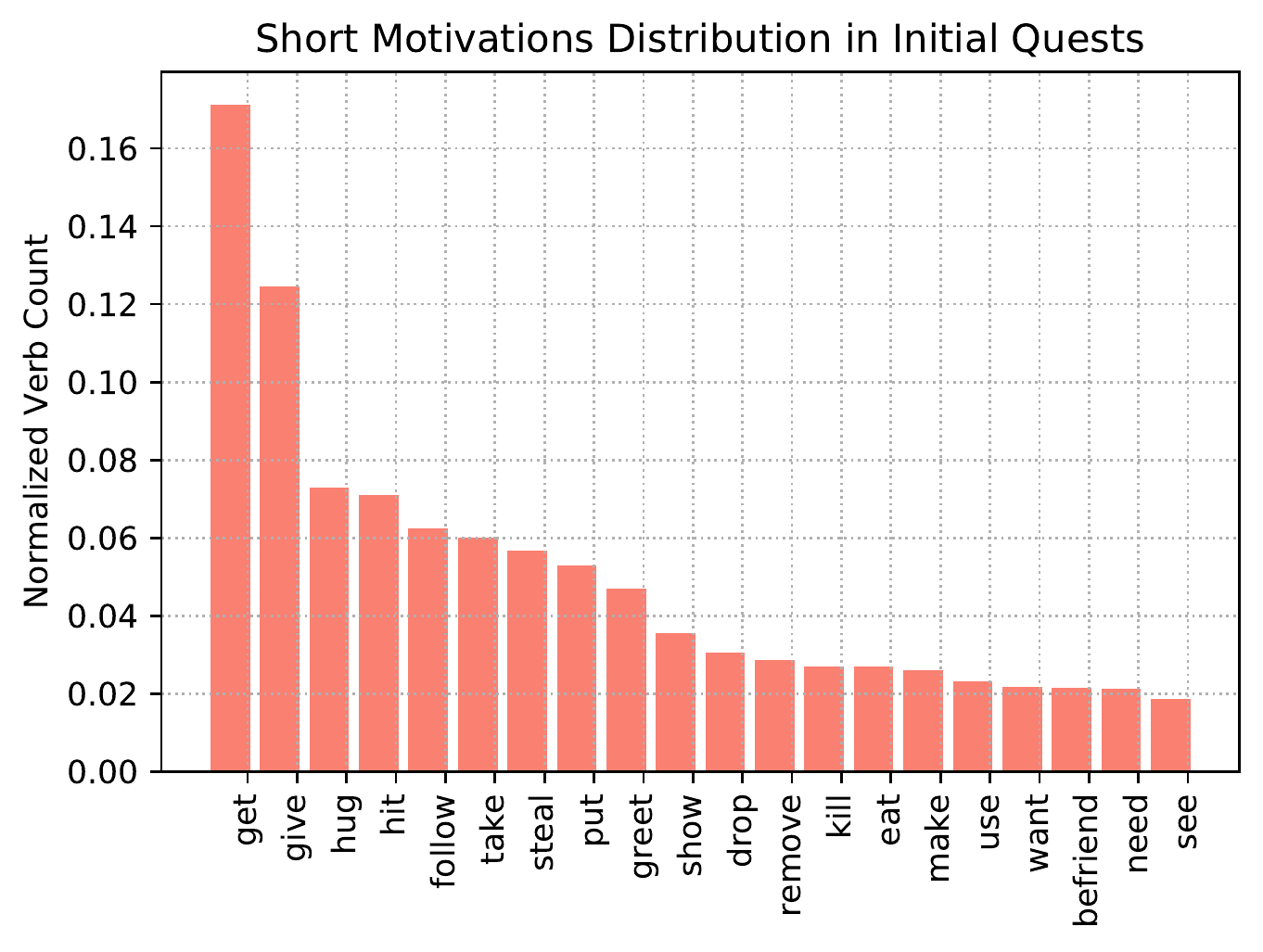}
    \caption{Normalized top-20 verb count distribution of short motivations of the LIGHT-Quests dataset.}
    \label{fig:originallightdistr}
\end{figure}

\textbf{Generating Curriculums.}
We generate curriculums by building off of our procedural LIGHT game instance generation pipeline.
We make the observation that the original quests in LIGHT are heavily skewed towards certain quest types---with the majority involving goals and short motivations that contain objectives related to getting and object, and hitting or hugging another character (Figure~\ref{fig:originallightdistr}).
We further note that the first verb in the short motivation forms the basis of the quest for that agent.

Actions in LIGHT, and more generally in text games, are executed in the game engines on the basis of verbs---engine subroutines are linked to verbs with nouns forming arguments---and as such are primarily responsible for changing the state of the world.
For example, {\em get sword} invokes the {\em get} subroutine that places an object, in this case a sword, in the character's surrounding into their inventory.
As the quest is generated early in the pipeline, with the world and the rest of the components being conditioned on it, we can say that the first verb in the short motivation is an important dimension along which we can assess the distribution of individual LIGHT game instances.
Thus, concretely, the verb counts from the short motivation aggregated over a set of quests represents the primary dimension along which we measure the distribution of quests.

\textbf{Parametrizing Curriculum Difficulty.}
Given the relative imbalance of this multinomial distribution, as seen in Figure~\ref{fig:originallightdistr}, we hypothesize that a LIGHT agent only learns to do well on certain types of objectives and not others---memorizing trajectories for less seen quest types, i.e. those found in the tail of the distribution.
Preliminary evidence for this hypothesis is also seen in \citet{Prabhumoye2020}, where they show a positive correlation between the number of instances of a particular type of quest during training and the final test goal-achievement performance.
Based on these observations and our initial hypothesis, we use this particular dimension to {\em parametrize curriculum difficulty} for training LIGHT agents---quest types that are rarer in the initial training data will be harder for the agent to generalize to in a zero-shot setting.

\begin{figure}
    \centering
    \includegraphics[width=.7\linewidth]{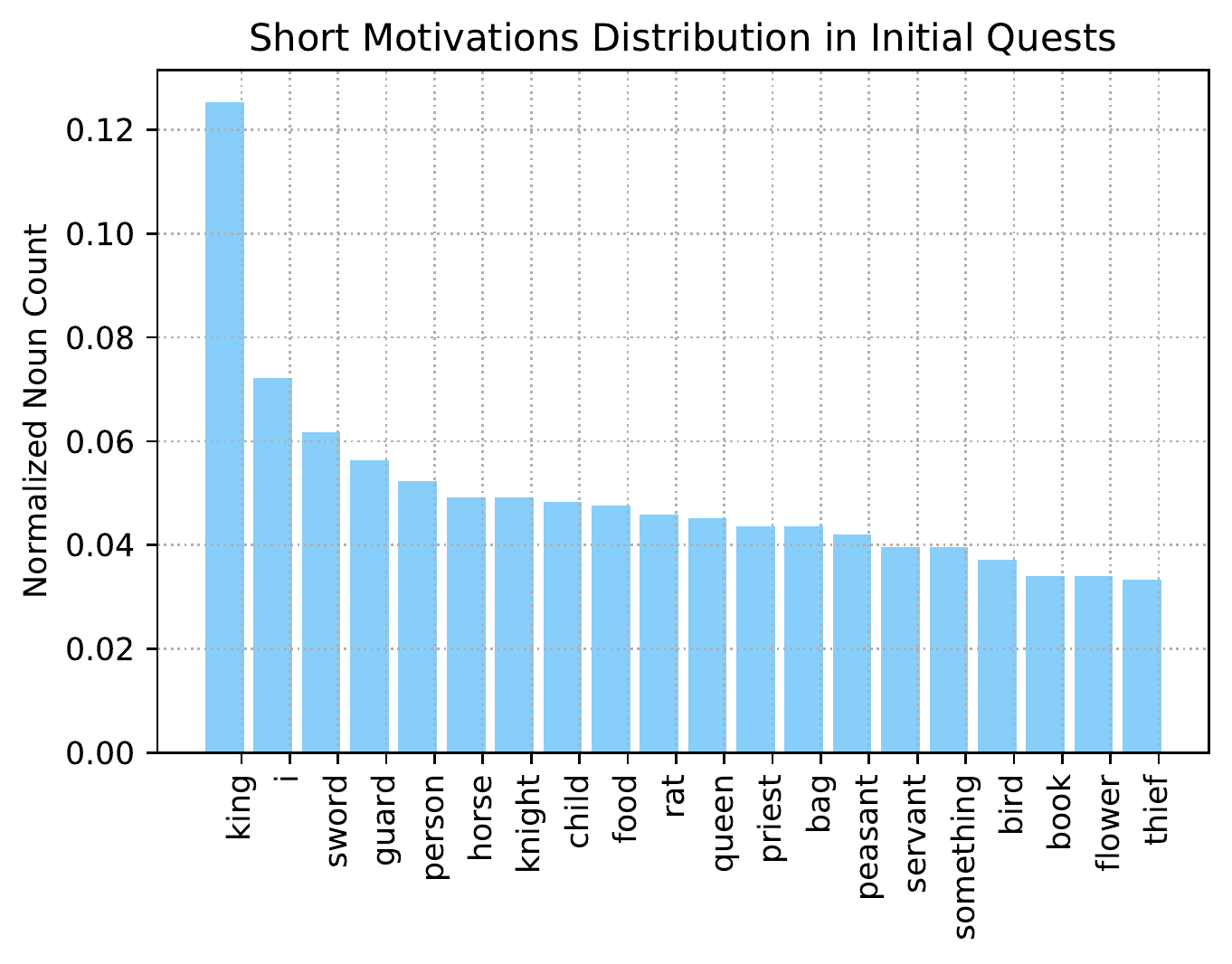}
    \caption{Normalized top-20 noun count distribution of short motivations of the LIGHT-Quests dataset.}
    \label{fig:originallightdistrnoun}
\end{figure}

Intuitively, we seek to create curriculums that contain a diverse set of game instances with quest types that are not often found in the initial training data.
Our earlier observations let us hypothesize that this will enable the LIGHT agent to more effectively learn from rare instances of quests as opposed to memorizing the corresponding trajectories.
To this end, the generated curriculums each consist of a pool of quests with steadily decreasing quest type imbalance.
In our case, this imply that the flatness of the multinomial distribution increases until it tends towards being uniform with respect to the categorical quest type variable.
This is done by running the procedural generation pipeline iteratively until the number of instances for the highest count quest type is within $n$ of the lowest count quest type.
The total number of additional generated instances is held fixed across curriculums, only the task distribution of quest types within each curriculum changes.
\begin{figure}
    \centering
    \includegraphics[width=.7\linewidth]{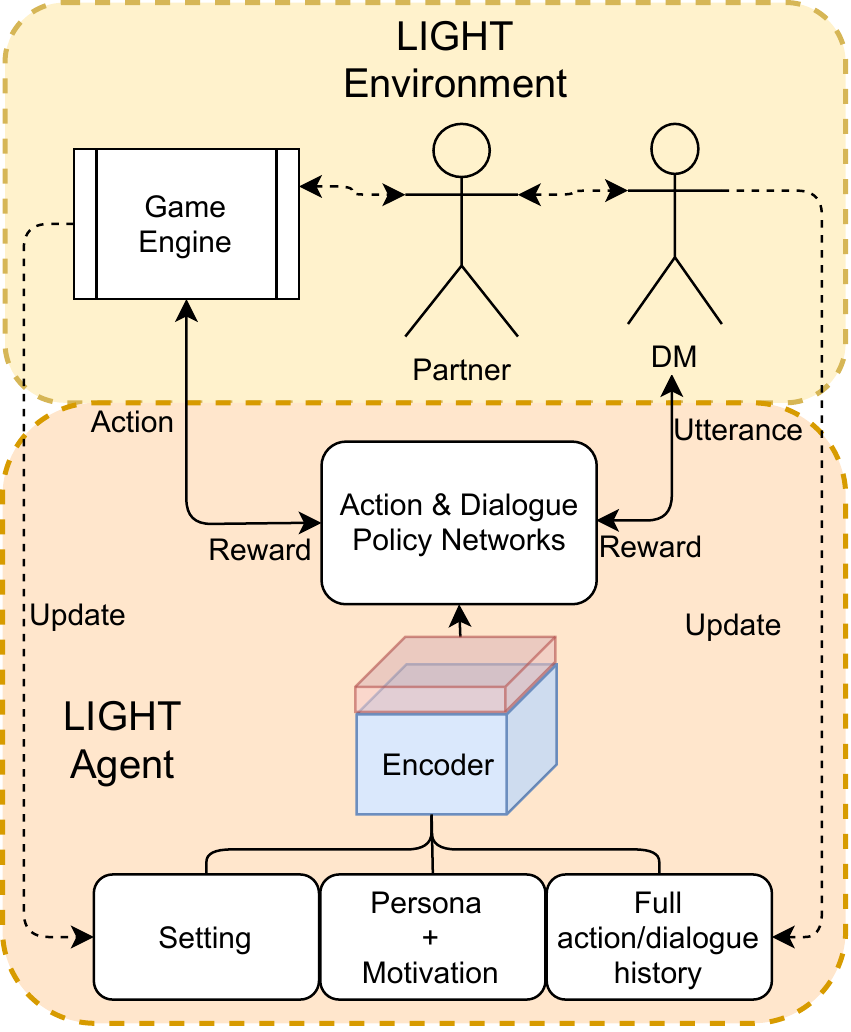}
    \caption{Architecture and training pipeline for the LIGHT RL Agent~\citep{Ammanabrolu2021}.}
    \label{fig:lightrlcurrmini}
\end{figure}

\begin{figure*}

\begin{minipage}{.24\textwidth}
    \centering
    \includegraphics[width=\linewidth]{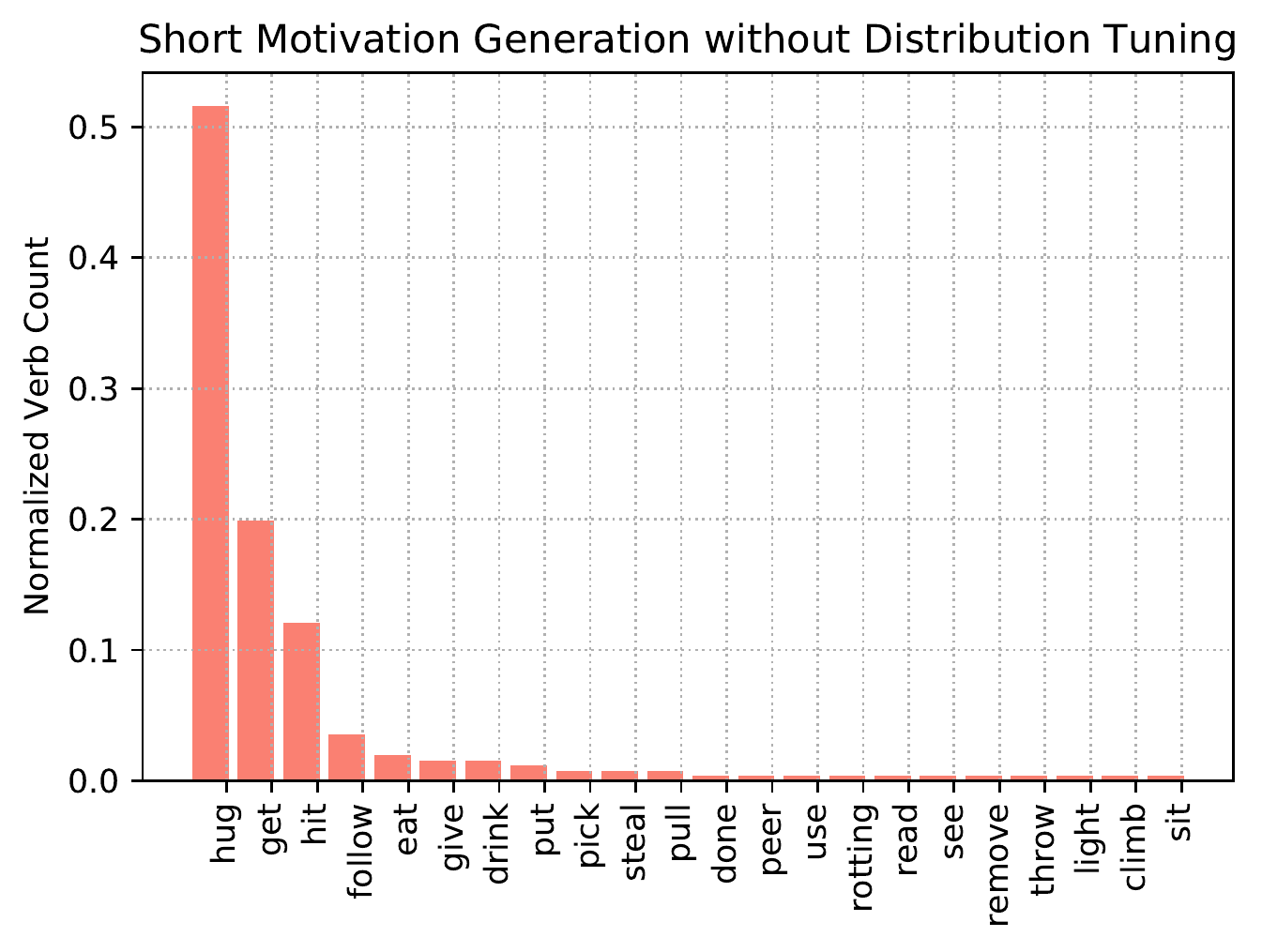}
\end{minipage}
\begin{minipage}{.24\textwidth}
    \centering
    \includegraphics[width=\linewidth]{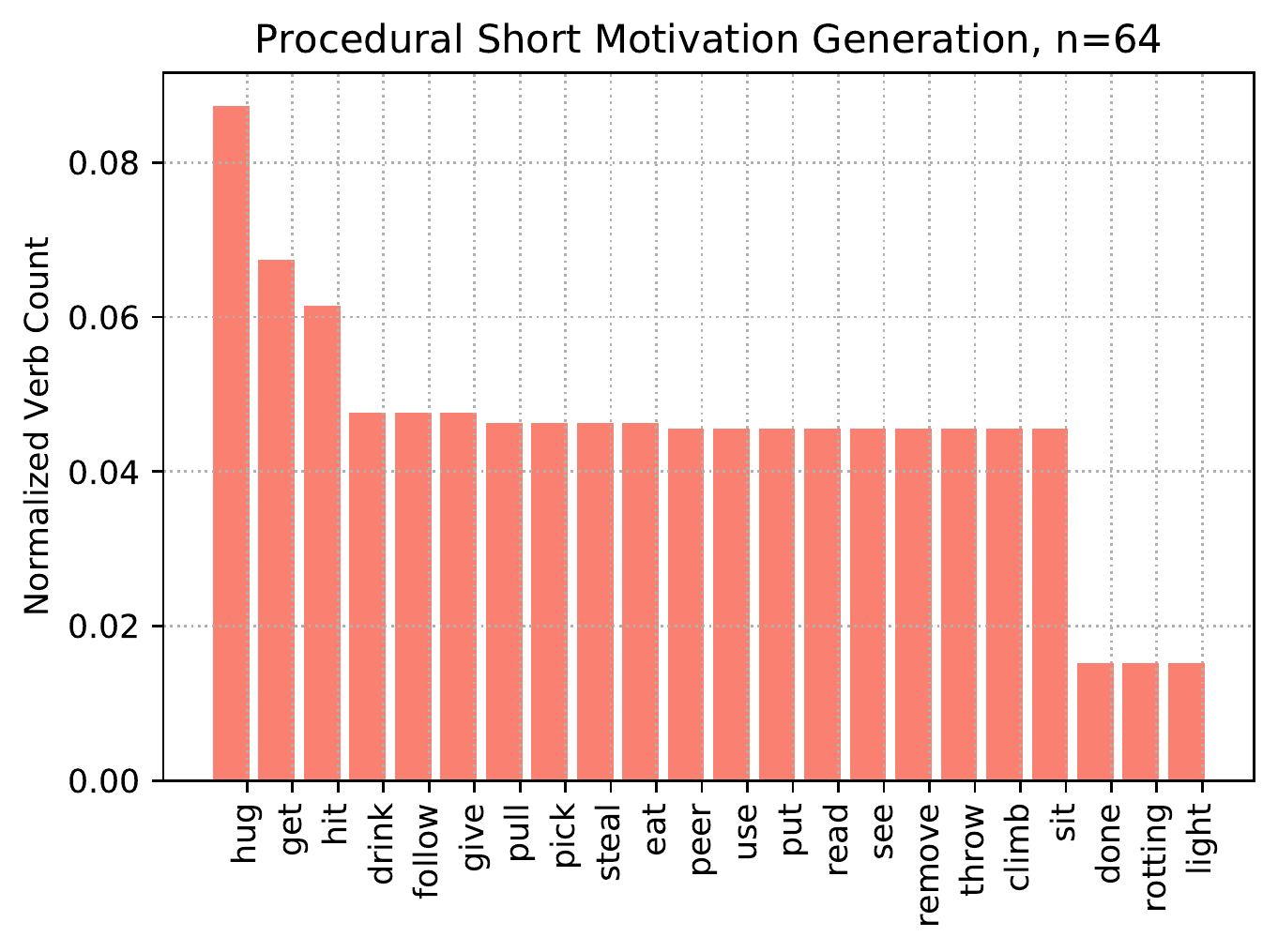}
\end{minipage}
\begin{minipage}[width=\linewidth]{.24\textwidth}
    \centering
    \includegraphics[width=\linewidth]{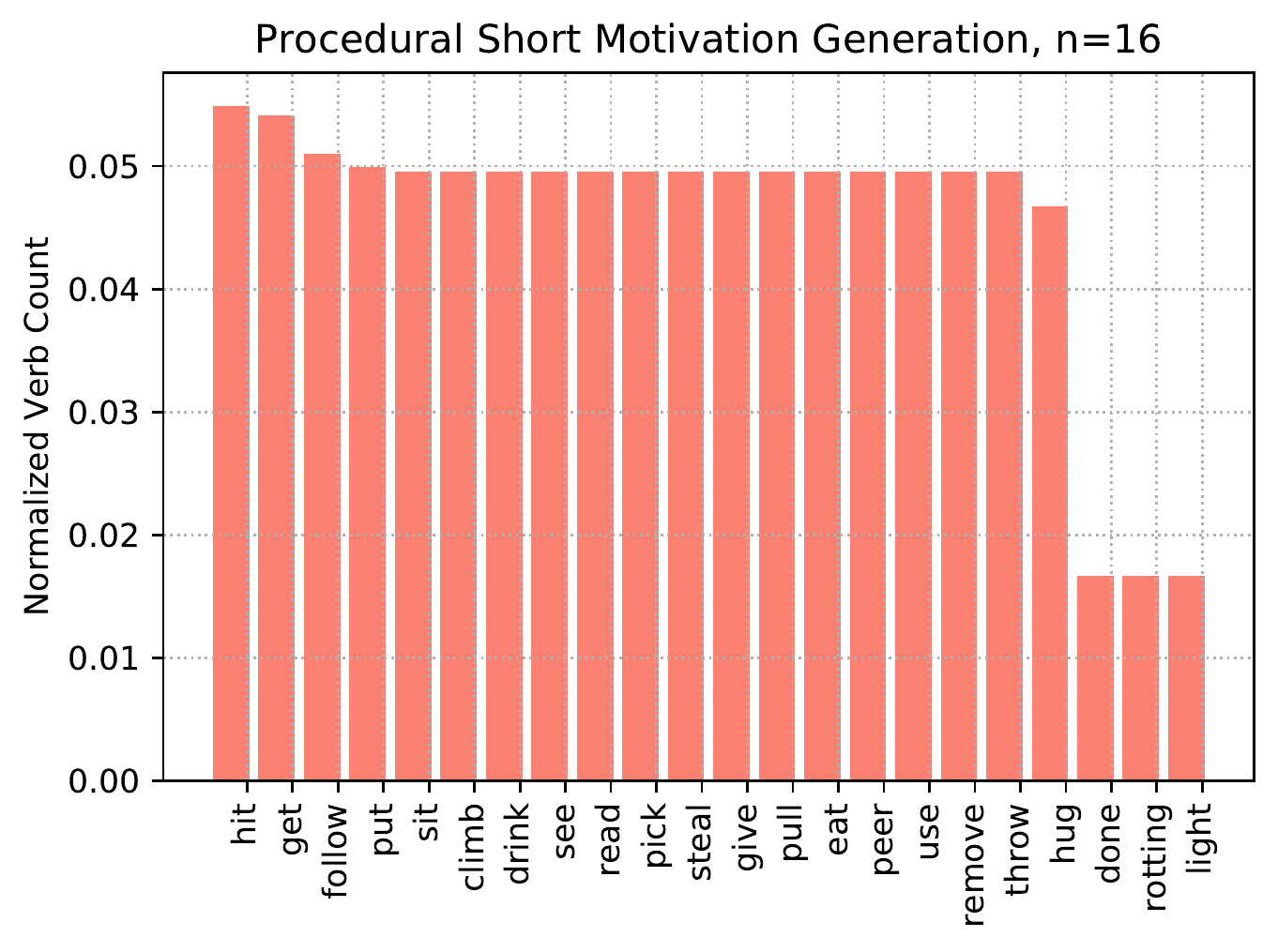}
\end{minipage}
\begin{minipage}[width=\linewidth]{.24\textwidth}
    \centering
    \includegraphics[width=\linewidth]{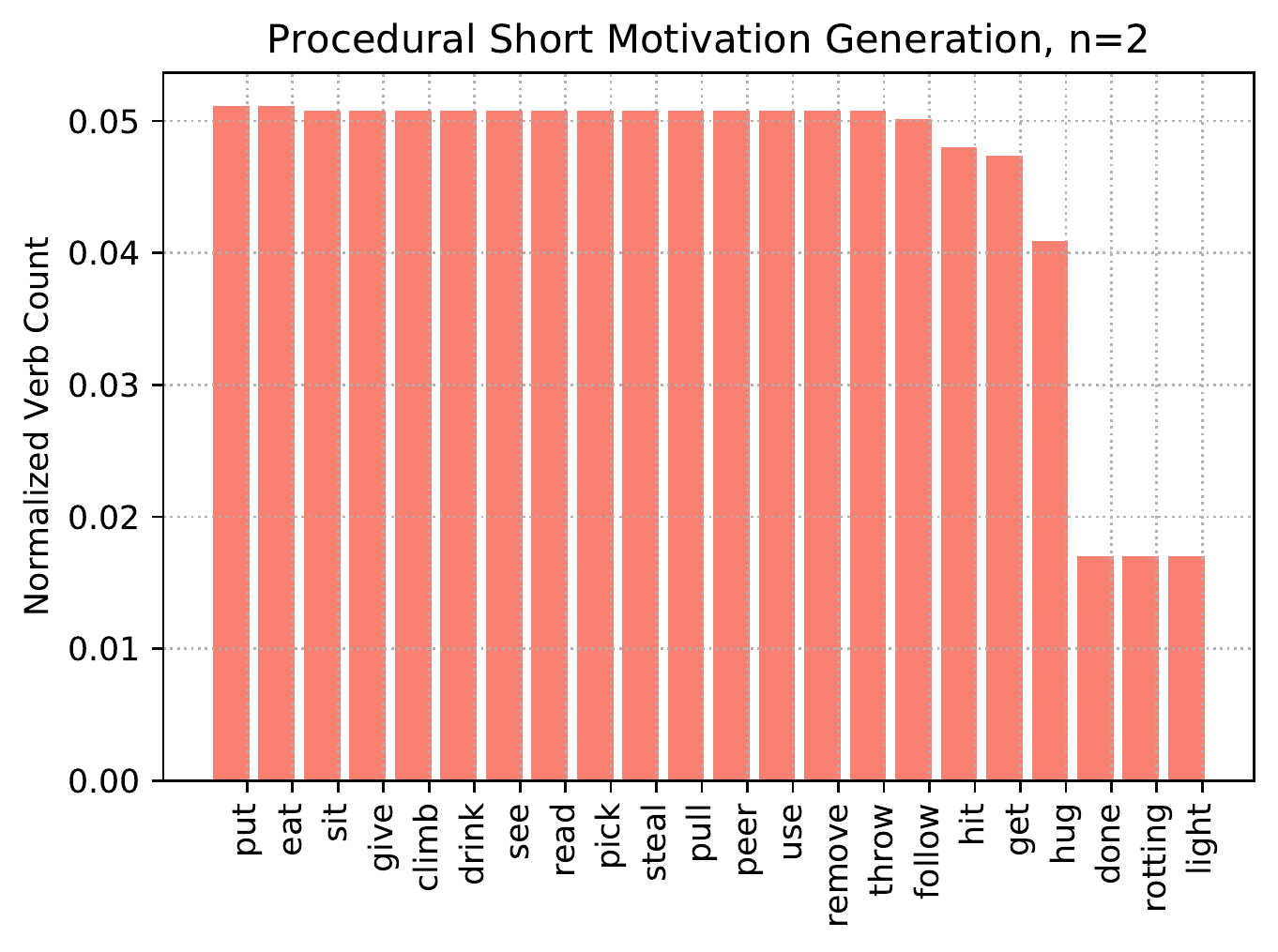}
\end{minipage}
\end{figure*}

\begin{figure*}
\begin{minipage}{.24\textwidth}
    \centering
    \includegraphics[width=\linewidth]{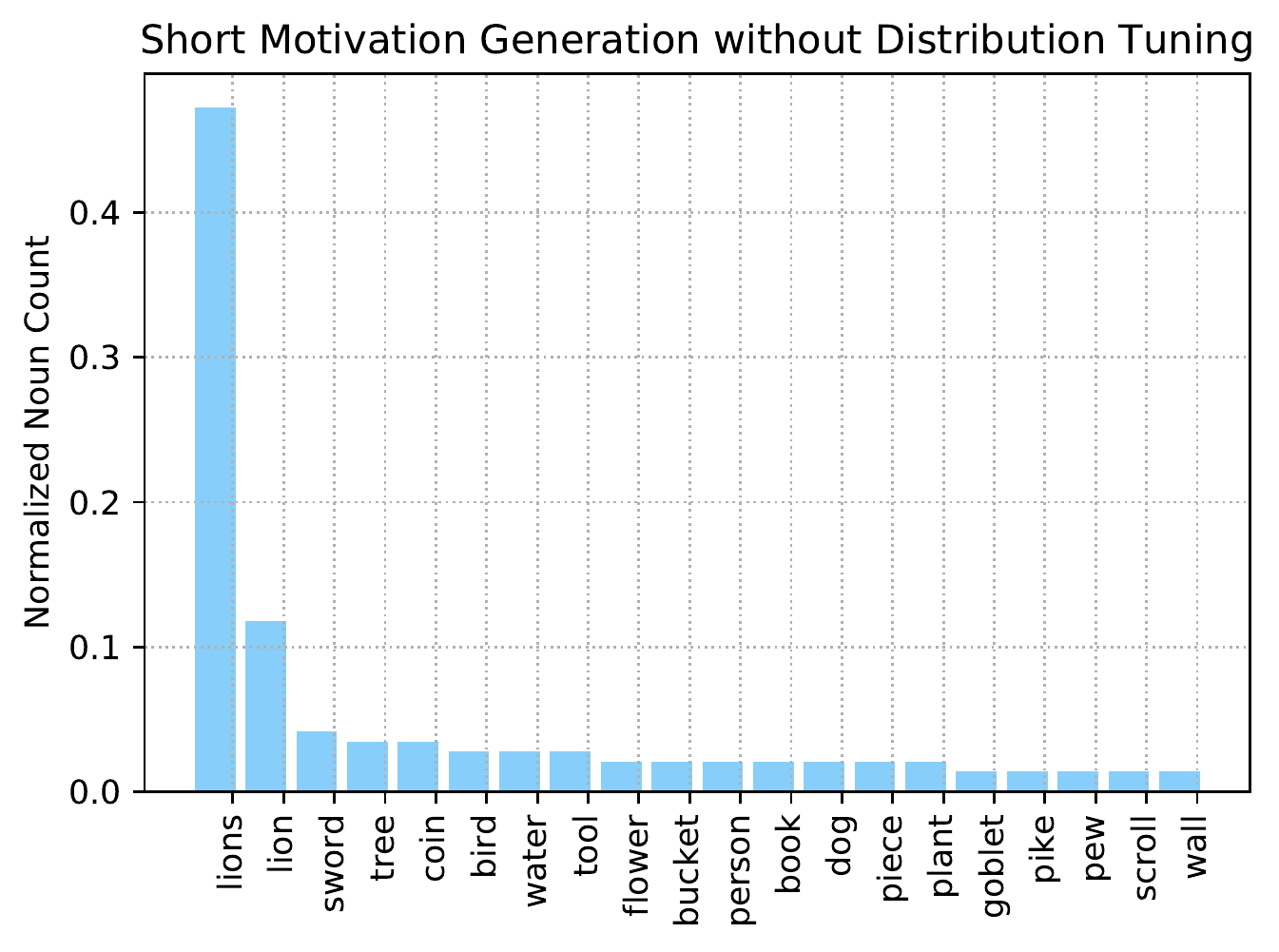}
\end{minipage}
\begin{minipage}{.24\textwidth}
    \centering
    \includegraphics[width=\linewidth]{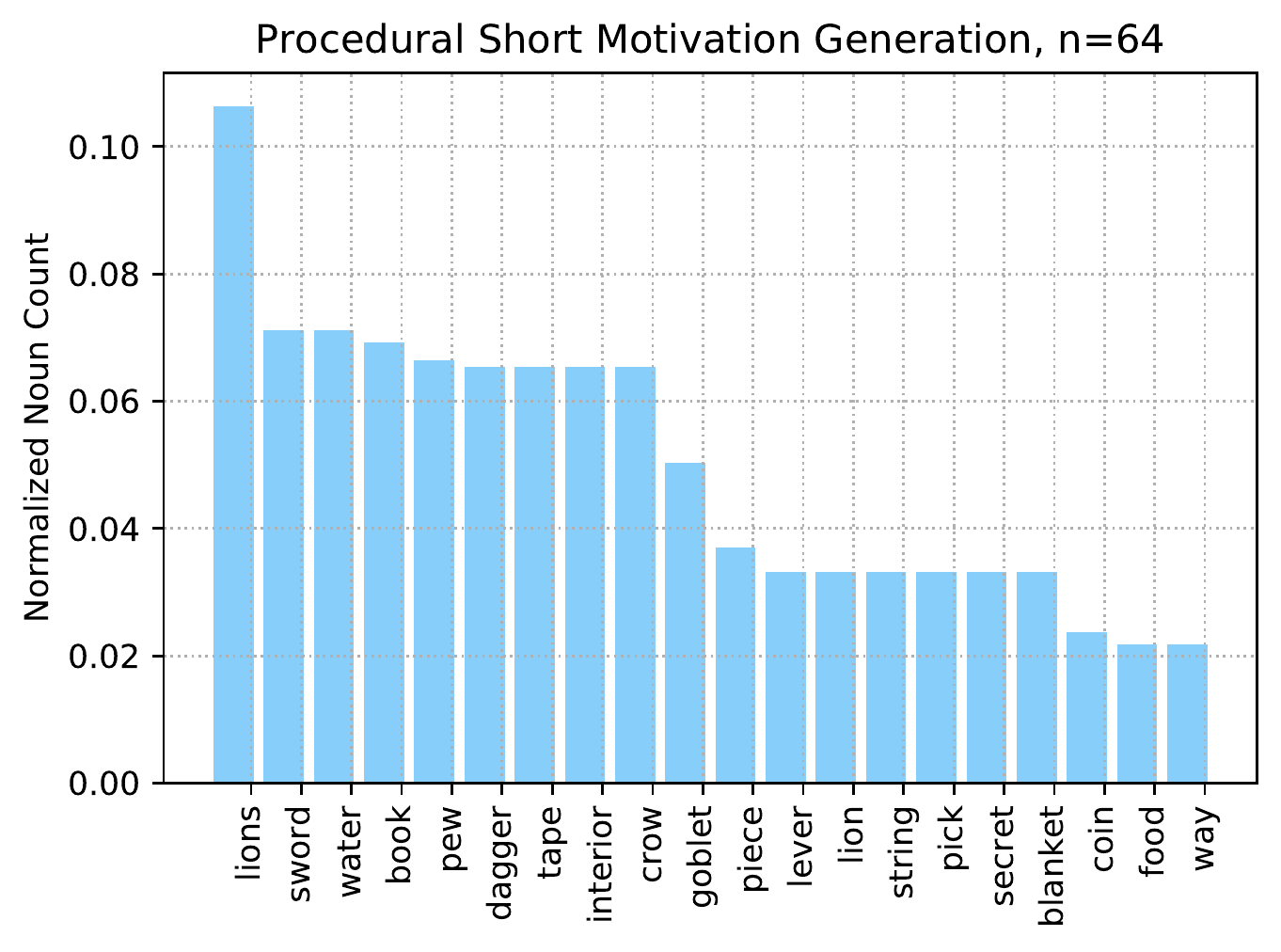}
\end{minipage}
\begin{minipage}[width=\linewidth]{.24\textwidth}
    \centering
    \includegraphics[width=\linewidth]{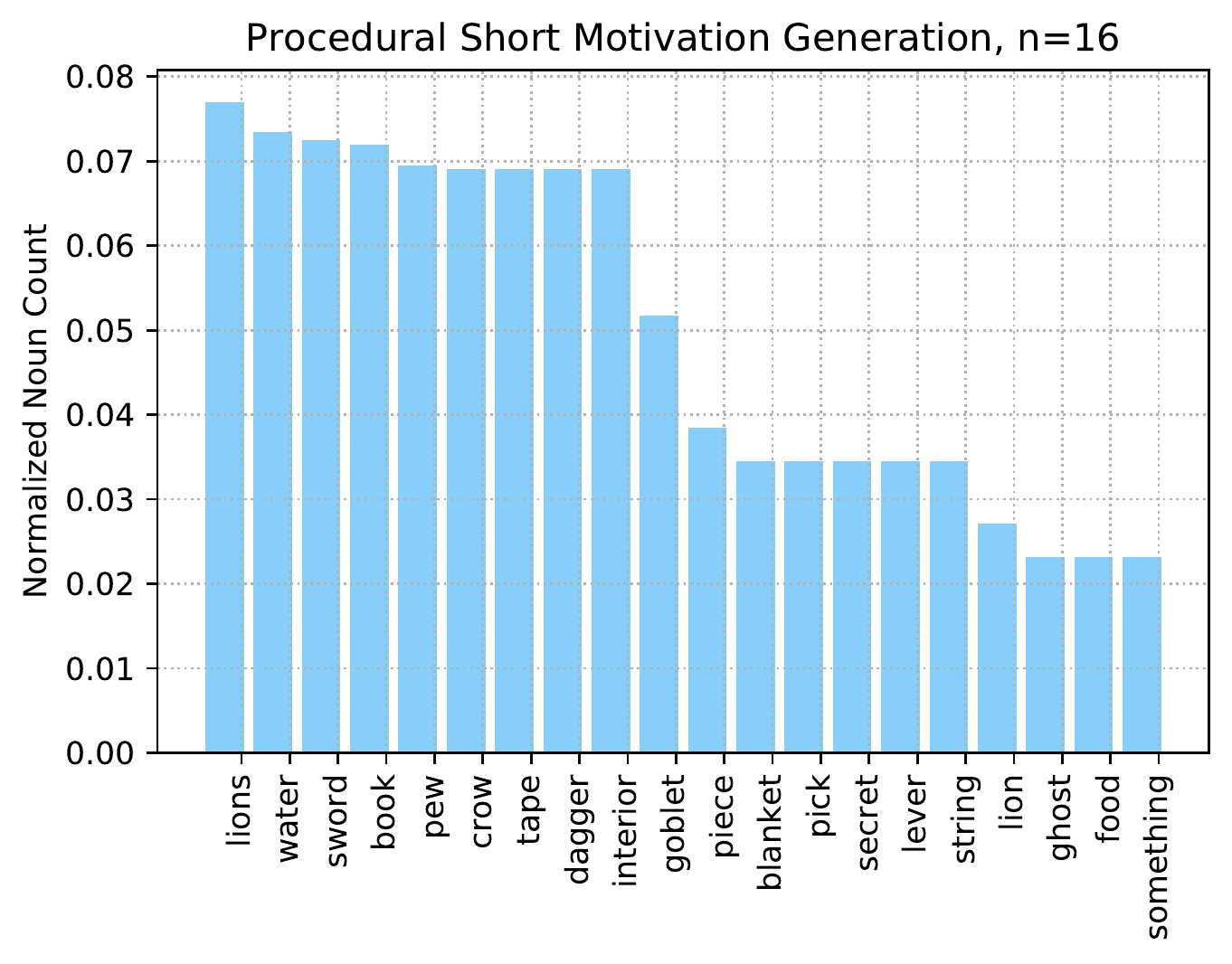}
\end{minipage}
\begin{minipage}[width=\linewidth]{.24\textwidth}
    \centering
    \includegraphics[width=\linewidth]{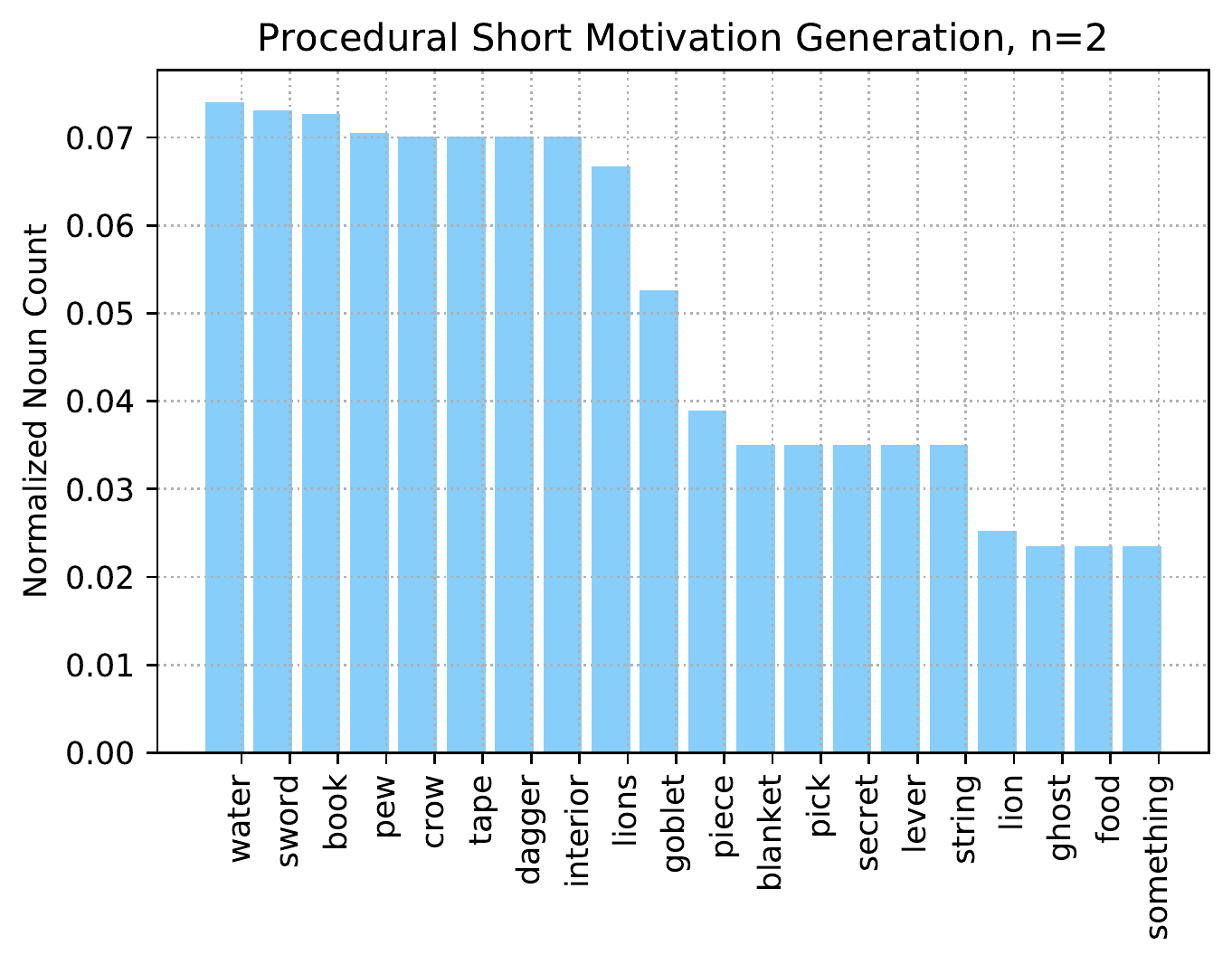}
\end{minipage}
\caption{Top-20 distribution of verbs (top) and nouns (bottom) in the short motivation of the curriculum of quests starting from the original generated curriculum on the left to the flattened, {\bf generated} curriculum on the right as a function of $n$ (Section~\ref{sec:curriculumgen}). The y-axis of the reflects normalized overall count in the pool of quests.}
\label{fig:generatedcurrnoun}
\end{figure*}

Figure~\ref{fig:generatedcurrnoun} shows that decreasing $n$ has the intended effect of decreasing imbalance with respect to verb types.
Generating using this pipeline has the added effect of increasing diversity within the pool of each available quest type.
One measure of diversity within the pool of a single quest type is the types of nouns contained within the short motivations---these generally correspond to the characters, locations, and objects mentioned.
Figure~\ref{fig:generatedcurrnoun} shows that decreasing imbalance in the verb types for a short motivation also results in decreasing imbalance in noun types, once again corresponding to decreasing $n$.
Short motivation generation is one of the first steps in the pipeline, i.e. the rest of the pipeline is conditioned on it, and as such increasing the flatness of the distribution there has the effects of increasing distribution for downstream components.

\textbf{A2C Curriculum Training.} Overall training is done via A2C~\citep{Mnih2016} a policy gradient algorithm that maximizes long-term expected reward by comparing the advantage $A(s_t,a^*_t)$ of taking an action $a_t$ in a state $s_t$ to the average value of taking any valid action as predicted by the critic $V(s_t)$. 
The setup and network architectures used are similar to \citet{Ammanabrolu2021} and are summarized in Figure~\ref{fig:lightrlcurrmini}.
At every step, the LIGHT agent receives as input the text describing the setting, the character's persona~\&~motivation, and the full dialogue history.
This is then encoded using a transformer based encoder and sent to the action and dialogue policy networks which output an action/dialogue utterance.
These are then passed into the LIGHT environment which process them and returns rewards to be used by the agent.

\textbf{Rewards.}
As seen in Figure~\ref{fig:lightrlcurrmini}, all actions, either those of the agent-in-training or the partner agent, are processed by the engine, checking for goal state completion---hence known as {\em act goals}.
For example, if the LIGHT agent had the motivation to acquire a sword, the goal could be completed via a: {\em self act completion}: where the agent acquires a sword itself by picking it up, stealing it, convincing the partner to drop theirs so you can pick it up, etc.
{\em partner act completion}: where the agent uses dialogue utterances to convince their partner to achieve the goal for them (e.g., by persuading the partner to give them the sword).
The naturalness of the dialogue utterances is further rated by a learned Dungeon Master (DM), a transformer-based ranker model trained on human demonstrations to score how relevant the utterance is given the character's persona and motivation.
Further training details are provided in Appendix~\ref{app:lightdetails}.

\section{Evaluation}

We conduct two separate evaluations: the first measures the effectiveness of the various models in the procedural environment generation pipeline as well as the effectiveness of the pipeline as a whole.
The second provides zero-shot ablations of the LIGHT RL agents trained on the resulting curriculums and answers the questions (1) how does the relative difficulty of the training quests effect test performance?; (2) how does the diversity of the environments during training effect test performance?; and (3) how are the results of the previous questions affected by pre-training?

\begin{table}[]
    \footnotesize
      \centering
        \begin{tabular}{lllll}
        \textbf{Pipeline Step} & \textbf{Model} & \textbf{Hits@10} & \textbf{F1} & \textbf{Ppl} \\ \hline
        \multicolumn{5}{c}{\textbf{World Generation}} \\ \hline
        \textbf{Location} & Biencoder & 0.543 & 0.153 & - \\
        \textbf{Object} & Biencoder & 0.563 & 0.154 & - \\ 
        \textbf{Character} & Starspace & 0.653 & 0.289 & - \\ 
        \hline
        \multicolumn{5}{c}{\textbf{Quest Generation}} \\ \hline
        \textbf{Short Motive} & BART & - & 0.488 & 7.55 \\
        \textbf{Goal Action} & BART & - & 0.763 & 3.75
        \end{tabular}
        \caption{Procedural generation evaluation showing metrics for each individual model in the pipeline. }
        \label{tab:procgeneval}
\end{table}

\subsection{Procedural Generation Evaluation}
All of the models in the pipeline described in Section~\ref{sec:lightprocgen} are trained using only the training set of the original LIGHT and LIGHT-Quests data.
LIGHT-Quests inherits characters, locations, and objects from the original LIGHT dataset and adds on motivations and goals in the form of quests.
Thus, the character, location, and object retrieval models are evaluated on the LIGHT unseen test set and the motivation and goal generation models are evaluated on the LIGHT-Quests test set.
We report the standard array of metrics: hits@10 and F1 ranking prediction score for retrieval models; and F1 (as a harmonic average of BLEU-1~\citep{papineni-etal-2002-bleu} and ROUGE-1~\citep{Lin2004}) and perplexity for generative models.
Hyperparameters for all models are found in Appendix~\ref{app:lightprocegenhyper}.

\textbf{Analysis.}
Table~\ref{tab:procgeneval} presents the results of this evaluation.
There are two primary trends to note: (1) character retrieval is easier than retrieving location and objects; and (2) goal action generation is easier than motivation generation.
We hypothesize that the first trend is a direct consequence of the fact that generated motivations and goals regularly contain the names of the characters involved but mostly leave implicit information such as the objects required---e.g. the action {\em hit dragon} as a knight would require a weapon such as a sword to be equipped first.
The second trend stems from the fact that goal actions can often be thought of as condensed version of the short motivation---number of tokens required to generate goal actions is far less than short motivations.
This implies that the goal action model is akin to a summarization model as opposed to the short motivation model which has the more difficult task of generating the motivation with only initial persona and location information.

\subsection{Curriculum Learning Evaluation}
\label{sec:curreval}
This evaluation tests the LIGHT RL agent's ability to zero-shot generalize to unseen environments.
For all experiments in this study, agents were each zero-shot evaluated on $211$ human demonstrations from the LIGHT-Quests test set for a single episode per quest across three independent runs.
They were measured on the basis of whether or not they were able to achieve their goals in the environments conditioned on their personas: {\em act goals} measuring their ability to act consistently, and {\em speech goals} reflecting their ability to speak naturally.
The study ablates across three dimensions in order to answer the posed research questions relating to: (1) curriculum difficulty, (2) curriculum diversity, and (3) agent pre-training.

\textbf{Curriculum Difficulty.} To measure the overall effectiveness of the distribution tuning technique shown in Section~\ref{sec:curriculumgen}, we vary the parameter $n$ used to measure curriculum difficulty---note that a lower $n$ corresponds to a flatter distribution and so is higher difficulty.
As seen in Fig.~\ref{fig:generatedcurrnoun}, we generate pools of quests with steadily increasing difficulty with varying $n$ based on the range of the original untuned distribution---with the agents being trained on each pool separately as well as all of them in sequence through a curriculum.
Agents received $10^7$ total environment interactions per parallel A2C agent in a batch of 16.
For the curriculum learning method, the agent received $2.5\times10^6$ interactions per pool of quests starting with the initial pool of untuned quests and then sequentially with $n=64,16,2$ resulting in a total of  $10^7$ total environment interactions per parallel A2C agent.

\textbf{Curriculum Diversity.} The variations in the combinations of quests and worlds themselves seen at training time has potential to effect zero-shot performance~\citep{samvelyan2021minihack}. 
We introduce two baselines that change the relative diversities of resulting quests in the curriculums, to contrast with our proposed procedural generation pipeline.
Generated quest details are found in Appendix~\ref{app:currstats}.
\setlist{nolistsep}
\begin{itemize}[noitemsep]
    \item \textit{Sampled Curriculums.}
Inspired by \citet{Chawla2002,Graves2017}, we explore an alternate method of creating curriculums by simply oversampling the same rare quests found in the tails of the distributions.
This method does not generate new environments via the pipeline, instead choosing to sample rarer instances of quests with a higher weight when initializing each parallel A2C actor.
This means that the distribution of verbs looks similar to what it is in Figure~\ref{fig:generatedcurrnoun} but the quests within a pool are repeated multiple times and so contain no new diversity.

    \item \textit{Randomly Generated Curriculums.}
On the other side of the diversity spectrum, we test a method that follows the same steps as the pipeline proposed in Section~\ref{sec:lightprocgen} with the modification that the selection process for each step in the pipeline is random.
The characters, objects, location are randomly selected and the generated motivations per character are conditioned on these randomly created worlds.
This results in a significantly higher diversity of quests per pool---at the expense of the relative coherence of the overall environment.
\end{itemize}


\textbf{Pre-training.}
We test two model types, drawing from \citet{Ammanabrolu2021}, to determine if pre-training effects curriculums learning.

\begin{itemize}[noitemsep]
    \item \textit{Scratch.} No pre-training is done, the encoder is a 3-layer randomly initialized transformer and trained along with the policy networks.
    \item \textit{Adaptive.} 
Pre-training is done on the tasks introduced in \citet{Ammanabrolu2021} by training a 12 layer transformer with $256$ million parameters using a cross-entropy loss as seen in~\citet{Humeau2020}.
These weights are then transferred to the encoder used during RL training then frozen with 3 randomly initialized-layers appended. %
The encoder is multi-task trained on both pushshift.io Reddit~\citep{Baumgartner2020} and the commonsense dataset ATOMIC-LIGHT~\citep{Ammanabrolu2021}, giving the agent general priors on how to act and speak.
It is then fine-tuned in LIGHT, giving the agent further domain specific priors.
Specific task details are provided in Appendix~\ref{app:lightdetails}.
\end{itemize}

\begin{table}[!h]
    \scriptsize
      \centering
        \begin{tabular}{rlll}
        \multicolumn{1}{l}{\textbf{Expt.}} & \textbf{Act Goals} & \textbf{Speech Goals} & \textbf{All Goals} \\ \hline
        \multicolumn{4}{c}{\textbf{Scratch Encoder}} \\ \hline
        \multicolumn{1}{l}{\textbf{No Curr.}} & 0.418 & 0.118 & 0.103 \\
        \multicolumn{1}{l}{\textbf{Sampled}} &  &  &  \\
        only n=64 & 0.392 & 0.113 & 0.097 \\
        only n=16 & 0.431 & 0.116 & 0.099 \\
        only n=2 & 0.435 & 0.124 & 0.111 \\
        curriculum & 0.460 & 0.145 & 0.138 \\
        \multicolumn{1}{l}{\textbf{Randomly Generated}} &  &  &  \\
        only n=64 &  0.221 &	0.011 &	0.009  \\
        only n=16 &  0.223 &	0.011 &	0.009  \\
        only n=2 &  0.257 &	0.016 &	0.012  \\
        curriculum & 0.263	& 0.024 &	0.017 \\
        \multicolumn{1}{l}{\textbf{Generated}} &  &  &  \\
        only n=64 & 0.426 & 0.121 & 0.107 \\
        only n=16 & 0.433 & 0.129 & 0.112 \\
        only n=2 & 0.432 & 0.130 & 0.112 \\
        curriculum & \textbf{0.477} & \textbf{0.163} & \textbf{0.155} \\ \hline
        \multicolumn{4}{c}{\textbf{Adaptive Encoder}} \\ \hline
        \multicolumn{1}{l}{\textbf{No Curr.}} & 0.420 & 0.330 & 0.303 \\
        \multicolumn{1}{l}{\textbf{Sampled}} &  &  &  \\
        only n=64 & 0.431 & 0.336 & 0.312 \\
        only n=16 & 0.450 & 0.340 & 0.317 \\
        only n=2 & 0.456 & 0.339 & 0.321 \\
        curriculum & 0.473 & 0.358 & 0.344 \\
        \multicolumn{1}{l}{\textbf{Randomly Generated}} &  &  &  \\
        only n=64 &  0.267 &	0.110 &	0.092  \\
        only n=16 &  0.271	& 0.125	& 0.116  \\
        only n=2 &  0.289 &	0.168 &	0.153  \\
        curriculum & 0.335 &	0.221 &	0.207 \\
        \multicolumn{1}{l}{\textbf{Generated}} &  &  &  \\
        only n=64 & 0.445 & 0.341 & 0.330 \\
        only n=16 & 0.469 & 0.367 & 0.359 \\
        only n=2 & 0.471 & 0.366 & 0.357 \\
        curriculum & \textbf{0.506} & \textbf{0.382} & \textbf{0.373}
        \end{tabular}
        \caption{Zero-shot goal achievement rates on a scale of 0-1, averaged over 3 random seeds with standard deviations not exceeding $0.02$. The ``All Goals'' column refers to quests where the agent has simultaneously achieved both types of goals within the allotted one episode. The parameter $n$ refers to the difference between the number of instances for the highest and lowest count quest types. All pair-wise comparisons made are statistically significant.}
        \label{tab:currlearningeval}
\end{table}


\textbf{Analysis.}
Table~\ref{tab:currlearningeval} presents the results of this evaluation.
We first report that the overall proportion of a pool of procedurally generated environments that contain achievable quests or goals for a single curriculum is $0.89$.
This metric provides a proxy for measuring the accuracy of the alignment process and the overall error rate of the pipeline.
The high achievability rate means that only a small proportion of LIGHT RL A2C agents will waste environment interactions learning from quests that cannot be completed---increasing this rate even further would likely also improve sample efficiency.

Further, we see that just the distribution tuning by itself shows no significant gains in performance over the baselines trained on the original data and in fact loses performance in certain cases.
In contrast, learning from the individually tuned quest pools in a sequential curriculum increases performance significantly.
This appears to indicate that LIGHT RL agents need to be trained with quests pools of steadily increasing difficulty---starting immediately on a set of quests with a high proportion of rare, generated quests can degrade performance.

The significantly increased performance of the procedurally generated curriculums over the sampled and randomly generated curriculums indicates the relative importance of diversity within a single quest type---but only up to a certain extent.
The sampled quests contain multiple instances of the same quest type but the generated ones have higher variability---leading to an increased observation space, ensuring that the agent cannot simply memorize trajectories.
On the other hand, randomly generated quests have even higher variability but sacrifice relative coherence---it is more likely that the world contains unlikely scenarios, e.g. a desert and swamp being located right next to each other---resulting in significantly decreased performance.

We'd finally like to note that the adaptive pre-trained model takes advantage of the generated curriculums and distribution tuning more than the non-pre-trained scratch encoder, showing consistenly higher performance across the board.
We hypothesize that this is likely a consequence of the adaptive model having greater model capacity---the pre-training enabling it to learn generalizable representations of the generated environments.
Overall, trends in performance are independent of pre-training---both the scratch and the adaptive pre-trained model benefit significantly from learning from the procedurally generated curriculums.

\section{Related Work}

\textbf{Text-based Game Playing and Generation.}
Recent text game playing works have focused on tackling three primary challenges: (1) how to represent agent knowledge to effectively operate in partially observable environments~\citep{Adhikari2020,sautier2020}; (2) scaling RL algorithms to handle combinatorial natural language state-action spaces~\citep{Zahavy2018,Ammanabrolu2020c,Jang2021}; and (3) giving agents commonsense priors to better reason about the world~\citep{Murugesan2020, murugesan2021textbased}

On the flip side, we have procedural generation of games with works such as \citet{Short2017,Risi2019,Khalifa2020} that focus on creating content especially for 2D visual games via search or reinforcement learning based methods.
\citet{Ammanabrolu2020a,ammanabrolu2020towards} use knowledge graphs to ground language and produce worlds and quests separately for text games from existing corpora such as stories.
\citet{Fan2019} leverage LIGHT to learn to generate interactive fiction worlds on the basis of locations, characters, and objects---this work is closest in spirit to our own World Generation module later on.
They all focus on either generating or playing games. 

\textbf{Goal oriented Dialogue.}
Sub-tasks within the overall task of goal oriented dialogue, such as dialogue state management~\citep{Singh2000,Pietquin2011,Fatemi2016} and response generation~\citep{Li2016} have used
RL to boost performance.
As noted by \citet{Ammanabrolu2021}, the negotiation tasks of \cite{Yarats2017,Lewis2017}, where two agents are trying to convince each other to perform certain actions, are related to the tasks in LIGHT-Quests.
These works all lack environment grounding. 

\textbf{Curriculum Learning.}
Curriculums in reinforcement learning have traditionally been used to set goals of steadily increasing difficulty for an agent~\citep{Bengio2009,Schmidhuber2013}.
The difficulty of these curriculums are generally measured difficulty via proxy of agent performance~\citep{Narvekar2020}---methods either choose to adversarially set goals of steadily increasing difficulty~\citep{Sukhbaatar2018,Racaniere2019,dennis2020emergent,Campero2021} or to maximize learning performance based on environment instances an agent finds difficult historically~\citep{Graves2017,Portelas2020}.
While we were inspired by these works, they all focus on searching for goals for agents which can be difficult to scale to complex tasks such our own natural language motivation-based goals.
We'd also like to note that most works using procedural generation to benchmark RL agents such as \citet{Cobbe2020,kuettler2020nethack,samvelyan2021minihack} rely on the underlying richness of the game engines to generate novel environments as opposed to learning to generate. 

\section{Conclusions}
We focus on the problem of improving zero-shot generalization abilities of goal-driven RL agents to act and speak via natural language.
An (obviously) key component of achieving this is to train the RL agents on a balanced training dataset that matches the test data in distribution.
As this is an unlikely scenario in most real-world applications, we make the observation that we can artificially augment our pool of training environments by generating curriculums to mimic this.
In our text game domain, with goal-driven situated natural language agents, we hypothesize---and gather supporting evidence suggesting---that an effective way to parametrize such distributions is by looking at the primary verbs within an agent's motivation and bringing the distribution of verb types as close to uniform as possible.
Curriculum training significantly increases an agent's ability to generalize to novel scenarios.

\section{Broader Impacts}
As noted by \citet{Urbanek2019} and \citet{Ammanabrolu2021}, the ability to speak and act in these textual fantasy worlds has implications for domains beyond text-games.
Text games are a platform where agents can interact in a relatively isolated environment and learn to interactively communicate effectively through natural language in a situated manner.
Our methods use both large language models and deep reinforcement learning and are prone to the pitfalls that other contemporary methods using these techniques face, especially in the areas of dialogue and text game systems.
We mitigate this first pitfall by restricting our current system to a retrieval based dialogue, ensuring that we can filter out non-normative dialogue usages beforehand, though we will note that the system can be extended to generative systems as described in \citet{Prabhumoye2020}.
Further, the LIGHT dataset is crowdsourced and contains data biases that can be attributed to the crowdworkers tasked with creating the data.
\citet{dinan2019queens} provides an in depth discussion regarding the inherent dataset biases, such as gender bias in the distribution of characters, in LIGHT and techniques to mitigate them---we follow these methods to reduce their effects on both the environment generation and agent training procedures.

\bibliography{acl}

\begin{thebibliography}{56}
\expandafter\ifx\csname natexlab\endcsname\relax\def\natexlab#1{#1}\fi

\bibitem[{Adhikari et~al.(2020)Adhikari, Yuan, C{\^o}t{\'e}, Zelinka, Rondeau,
  Laroche, Poupart, Tang, Trischler, and Hamilton}]{Adhikari2020}
Ashutosh Adhikari, Xingdi Yuan, Marc-Alexandre C{\^o}t{\'e}, Mikul{\'a}{\v{s}}
  Zelinka, Marc-Antoine Rondeau, Romain Laroche, Pascal Poupart, Jian Tang,
  Adam Trischler, and Will Hamilton. 2020.
\newblock Learning dynamic belief graphs to generalize on text-based games.
\newblock \emph{Advances in Neural Information Processing Systems}, 33.

\bibitem[{Ammanabrolu et~al.(2020{\natexlab{a}})Ammanabrolu, Broniec, Mueller,
  Paul, and Riedl}]{ammanabrolu2020towards}
Prithviraj Ammanabrolu, William Broniec, Alex Mueller, Jeremy Paul, and Mark~O.
  Riedl. 2020{\natexlab{a}}.
\newblock \href {https://arxiv.org/abs/1909.06283} {Toward automated quest
  generation in text-adventure games}.
\newblock In \emph{International Conference on Computational Creativity
  (ICCC)}.

\bibitem[{Ammanabrolu et~al.(2020{\natexlab{b}})Ammanabrolu, Cheung, Tu,
  Broniec, and Riedl}]{Ammanabrolu2020a}
Prithviraj Ammanabrolu, Wesley Cheung, Dan Tu, William Broniec, and Mark~O
  Riedl. 2020{\natexlab{b}}.
\newblock \href {https://www.aaai.org/ojs/index.php/AIIDE/article/view/7400}
  {Bringing stories alive: Generating interactive fiction worlds}.
\newblock In \emph{Proceedings of the Sixteenth AAAI Conference on Artificial
  Intelligence and Interactive Digital Entertainment (AIIDE-20)}.

\bibitem[{Ammanabrolu and Hausknecht(2020)}]{Ammanabrolu2020c}
Prithviraj Ammanabrolu and Matthew Hausknecht. 2020.
\newblock \href {https://openreview.net/forum?id=B1x6w0EtwH} {{Graph
  Constrained Reinforcement Learning for Natural Language Action Spaces}}.
\newblock In \emph{International Conference on Learning Representations}.

\bibitem[{Ammanabrolu and Riedl(2021)}]{ammanabrolu2021situated}
Prithviraj Ammanabrolu and Mark~O Riedl. 2021.
\newblock \href {https://www.cell.com/patterns/fulltext/S2666-3899(21)00159-8}
  {Situated language learning via interactive narratives}.
\newblock \emph{Patterns, Cell Press}.

\bibitem[{Ammanabrolu et~al.(2021)Ammanabrolu, Urbanek, Li, Szlam,
  Rockt{\"a}schel, and Weston}]{Ammanabrolu2021}
Prithviraj Ammanabrolu, Jack Urbanek, Margaret Li, Arthur Szlam, Tim
  Rockt{\"a}schel, and Jason Weston. 2021.
\newblock \href {https://arxiv.org/abs/2010.00685} {How to motivate your
  dragon: Teaching goal-driven agents to speak and act in fantasy worlds}.
\newblock In \emph{Proceedings of 2021 Annual Conference of the North American
  Chapter of the Association for Computational Linguistics: Human Language
  Technologies, NAACL-HLT 2021}.

\bibitem[{Barsalou(2008)}]{Barsalou2008}
Lawrence~W. Barsalou. 2008.
\newblock \href {https://doi.org/10.1146/annurev.psych.59.103006.093639}
  {Grounded cognition}.
\newblock \emph{Annual Review of Psychology}, 59(1):617--645.
\newblock PMID: 17705682.

\bibitem[{Baumgartner et~al.(2020)Baumgartner, Zannettou, Keegan, Squire, and
  Blackburn}]{Baumgartner2020}
Jason Baumgartner, Savvas Zannettou, Brian Keegan, Megan Squire, and Jeremy
  Blackburn. 2020.
\newblock The pushshift reddit dataset.
\newblock In \emph{Proceedings of the International AAAI Conference on Web and
  Social Media}, volume~14, pages 830--839.

\bibitem[{Bengio et~al.(2009)Bengio, Louradour, Collobert, and
  Weston}]{Bengio2009}
Yoshua Bengio, J{\'e}r{\^o}me Louradour, Ronan Collobert, and Jason Weston.
  2009.
\newblock Curriculum learning.
\newblock In \emph{Proceedings of the 26th annual international conference on
  machine learning}, pages 41--48.

\bibitem[{Bisk et~al.(2020)Bisk, Holtzman, Thomason, Andreas, Bengio, Chai,
  Lapata, Lazaridou, May, Nisnevich, Pinto, and
  Turian}]{bisk-etal-2020-experience}
Yonatan Bisk, Ari Holtzman, Jesse Thomason, Jacob Andreas, Yoshua Bengio, Joyce
  Chai, Mirella Lapata, Angeliki Lazaridou, Jonathan May, Aleksandr Nisnevich,
  Nicolas Pinto, and Joseph Turian. 2020.
\newblock \href {https://doi.org/10.18653/v1/2020.emnlp-main.703} {Experience
  grounds language}.
\newblock In \emph{Proceedings of the 2020 Conference on Empirical Methods in
  Natural Language Processing (EMNLP)}, pages 8718--8735, Online. Association
  for Computational Linguistics.

\bibitem[{Campero et~al.(2021)Campero, Raileanu, Kuttler, Tenenbaum,
  Rockt{\"a}schel, and Grefenstette}]{Campero2021}
Andres Campero, Roberta Raileanu, Heinrich Kuttler, Joshua~B. Tenenbaum, Tim
  Rockt{\"a}schel, and Edward Grefenstette. 2021.
\newblock \href {https://openreview.net/forum?id=ETBc_MIMgoX} {Learning with
  {\{}amig{\}}o: Adversarially motivated intrinsic goals}.
\newblock In \emph{International Conference on Learning Representations}.

\bibitem[{Chawla et~al.(2002)Chawla, Bowyer, Hall, and Kegelmeyer}]{Chawla2002}
Nitesh~V Chawla, Kevin~W Bowyer, Lawrence~O Hall, and W~Philip Kegelmeyer.
  2002.
\newblock Smote: synthetic minority over-sampling technique.
\newblock \emph{Journal of artificial intelligence research}, 16:321--357.

\bibitem[{Cobbe et~al.(2020)Cobbe, Hesse, Hilton, and Schulman}]{Cobbe2020}
Karl Cobbe, Chris Hesse, Jacob Hilton, and John Schulman. 2020.
\newblock Leveraging procedural generation to benchmark reinforcement learning.
\newblock In \emph{International conference on machine learning}, pages
  2048--2056. PMLR.

\bibitem[{C\^ot\'e et~al.(2018)C\^ot\'e, K\'ad\'ar, Yuan, Kybartas, Barnes,
  Fine, Moore, Hausknecht, Asri, Adada, Tay, and Trischler}]{Cote2018}
Marc-Alexandre C\^ot\'e, \'Akos K\'ad\'ar, Xingdi Yuan, Ben Kybartas, Tavian
  Barnes, Emery Fine, James Moore, Matthew Hausknecht, Layla~El Asri, Mahmoud
  Adada, Wendy Tay, and Adam Trischler. 2018.
\newblock Textworld: A learning environment for text-based games.
\newblock \emph{CoRR}, abs/1806.11532.

\bibitem[{Degris and Sigaud(2013)}]{Degris2013}
Thomas Degris and Olivier Sigaud. 2013.
\newblock Factored markov decision processes.
\newblock \emph{Markov Decision Processes in Artificial Intelligence}, pages
  99--126.

\bibitem[{Dennis et~al.(2020)Dennis, Jaques, Vinitsky, Bayen, Russell, Critch,
  and Levine}]{dennis2020emergent}
Michael Dennis, Natasha Jaques, Eugene Vinitsky, Alexandre Bayen, Stuart
  Russell, Andrew Critch, and Sergey Levine. 2020.
\newblock Emergent complexity and zero-shot transfer via unsupervised
  environment design.
\newblock \emph{Neural Information Processing Systems (NeurIPS)}.

\bibitem[{Devlin et~al.(2018)Devlin, Chang, Lee, and Toutanova}]{Devlin2018}
Jacob Devlin, Ming{-}Wei Chang, Kenton Lee, and Kristina Toutanova. 2018.
\newblock \href {http://arxiv.org/abs/1810.04805} {{BERT:} pre-training of deep
  bidirectional transformers for language understanding}.
\newblock \emph{CoRR}, abs/1810.04805.

\bibitem[{Dinan et~al.(2020)Dinan, Fan, Williams, Urbanek, Kiela, and
  Weston}]{dinan2019queens}
Emily Dinan, Angela Fan, Adina Williams, Jack Urbanek, Douwe Kiela, and Jason
  Weston. 2020.
\newblock Queens are powerful too: Mitigating gender bias in dialogue
  generation.
\newblock In \emph{Proceedings of Empirical Methods in Natural Language
  Processing (EMNLP-20)}.

\bibitem[{Fan et~al.(2019)Fan, Urbanek, Ringshia, Dinan, Qian, Karamcheti,
  Prabhumoye, Kiela, Rocktaschel, Szlam, and Others}]{Fan2019}
Angela Fan, Jack Urbanek, Pratik Ringshia, Emily Dinan, Emma Qian, Siddharth
  Karamcheti, Shrimai Prabhumoye, Douwe Kiela, Tim Rocktaschel, Arthur Szlam,
  and Others. 2019.
\newblock {Generating Interactive Worlds with Text}.
\newblock In \emph{In proceedings of the Thirty-Third AAAI Conference on AI
  (AAAI-19)}.

\bibitem[{Fatemi et~al.(2016)Fatemi, Asri, Schulz, He, and
  Suleman}]{Fatemi2016}
Mehdi Fatemi, Layla~El Asri, Hannes Schulz, Jing He, and Kaheer Suleman. 2016.
\newblock Policy networks with two-stage training for dialogue systems.
\newblock \emph{arXiv preprint arXiv:1606.03152}.

\bibitem[{Graves et~al.(2017)Graves, Bellemare, Menick, Munos, and
  Kavukcuoglu}]{Graves2017}
Alex Graves, Marc~G Bellemare, Jacob Menick, R{\'e}mi Munos, and Koray
  Kavukcuoglu. 2017.
\newblock Automated curriculum learning for neural networks.
\newblock In \emph{International Conference on Machine Learning}, pages
  1311--1320.

\bibitem[{Hausknecht et~al.(2020)Hausknecht, Ammanabrolu, C{\^{o}}t{\'{e}}, and
  Yuan}]{Hausknecht2020}
Matthew Hausknecht, Prithviraj Ammanabrolu, Marc-Alexandre C{\^{o}}t{\'{e}},
  and Xingdi Yuan. 2020.
\newblock \href {https://arxiv.org/abs/1909.05398} {Interactive fiction games:
  A colossal adventure}.
\newblock In \emph{Thirty-Fourth AAAI Conference on Artificial Intelligence
  (AAAI)}.

\bibitem[{Humeau et~al.(2020)Humeau, Shuster, Lachaux, and Weston}]{Humeau2020}
Samuel Humeau, Kurt Shuster, Marie-Anne Lachaux, and Jason Weston. 2020.
\newblock \href {https://openreview.net/forum?id=SkxgnnNFvH} {Poly-encoders:
  Architectures and pre-training strategies for fast and accurate
  multi-sentence scoring}.
\newblock In \emph{International Conference on Learning Representations}.

\bibitem[{Jang et~al.(2021)Jang, Seo, Lee, and Kim}]{Jang2021}
Youngsoo Jang, Seokin Seo, Jongmin Lee, and Kee-Eung Kim. 2021.
\newblock \href {https://openreview.net/forum?id=7_G8JySGecm} {Monte-carlo
  planning and learning with language action value estimates}.
\newblock In \emph{International Conference on Learning Representations}.

\bibitem[{Justesen et~al.(2018)Justesen, Torrado, Bontrager, Khalifa, Togelius,
  and Risi}]{Justesen2018}
Niels Justesen, Ruben~Rodriguez Torrado, Philip Bontrager, Ahmed Khalifa,
  Julian Togelius, and Sebastian Risi. 2018.
\newblock Illuminating generalization in deep reinforcement learning through
  procedural level generation.
\newblock \emph{arXiv preprint arXiv:1806.10729}.

\bibitem[{Khalifa et~al.(2020)Khalifa, Bontrager, Earle, and
  Togelius}]{Khalifa2020}
Ahmed Khalifa, Philip Bontrager, Sam Earle, and Julian Togelius. 2020.
\newblock Pcgrl: Procedural content generation via reinforcement learning.
\newblock In \emph{Proceedings of the AAAI Conference on Artificial
  Intelligence and Interactive Digital Entertainment}, volume~16, pages
  95--101.

\bibitem[{K{\"{u}}ttler et~al.(2020)K{\"{u}}ttler, Nardelli, Miller, Raileanu,
  Selvatici, Grefenstette, and Rockt{\"{a}}schel}]{kuettler2020nethack}
Heinrich K{\"{u}}ttler, Nantas Nardelli, Alexander~H. Miller, Roberta Raileanu,
  Marco Selvatici, Edward Grefenstette, and Tim Rockt{\"{a}}schel. 2020.
\newblock {The NetHack Learning Environment}.
\newblock In \emph{Proceedings of the Conference on Neural Information
  Processing Systems (NeurIPS)}.

\bibitem[{Lee et~al.(2019)Lee, Cho, and Kiela}]{Lee2019}
Jason Lee, Kyunghyun Cho, and Douwe Kiela. 2019.
\newblock Countering language drift via visual grounding.
\newblock \emph{arXiv preprint arXiv:1909.04499}.

\bibitem[{Lewis et~al.(2020)Lewis, Liu, Goyal, Ghazvininejad, Mohamed, Levy,
  Stoyanov, and Zettlemoyer}]{lewis-etal-2020-bart}
Mike Lewis, Yinhan Liu, Naman Goyal, Marjan Ghazvininejad, Abdelrahman Mohamed,
  Omer Levy, Veselin Stoyanov, and Luke Zettlemoyer. 2020.
\newblock \href {https://doi.org/10.18653/v1/2020.acl-main.703} {{BART}:
  Denoising sequence-to-sequence pre-training for natural language generation,
  translation, and comprehension}.
\newblock In \emph{Proceedings of the 58th Annual Meeting of the Association
  for Computational Linguistics}, pages 7871--7880, Online. Association for
  Computational Linguistics.

\bibitem[{Lewis et~al.(2017)Lewis, Yarats, Dauphin, Parikh, and
  Batra}]{Lewis2017}
Mike Lewis, Denis Yarats, Yann~N Dauphin, Devi Parikh, and Dhruv Batra. 2017.
\newblock Deal or no deal? end-to-end learning for negotiation dialogues.
\newblock \emph{arXiv preprint arXiv:1706.05125}.

\bibitem[{Li et~al.(2016)Li, Monroe, Ritter, Galley, Gao, and
  Jurafsky}]{Li2016}
Jiwei Li, Will Monroe, Alan Ritter, Michel Galley, Jianfeng Gao, and Dan
  Jurafsky. 2016.
\newblock \href {http://arxiv.org/abs/1606.01541} {Deep reinforcement learning
  for dialogue generation}.
\newblock \emph{CoRR}, abs/1606.01541.

\bibitem[{Lin(2004)}]{Lin2004}
Chin-Yew Lin. 2004.
\newblock \href {https://www.aclweb.org/anthology/W04-1013} {{ROUGE}: A package
  for automatic evaluation of summaries}.
\newblock In \emph{Text Summarization Branches Out}, pages 74--81, Barcelona,
  Spain. Association for Computational Linguistics.

\bibitem[{Mazar{\'e} et~al.(2018)Mazar{\'e}, Humeau, Raison, and
  Bordes}]{Mazare2018}
Pierre-Emmanuel Mazar{\'e}, Samuel Humeau, Martin Raison, and Antoine Bordes.
  2018.
\newblock \href {https://doi.org/10.18653/v1/D18-1298} {Training millions of
  personalized dialogue agents}.
\newblock In \emph{Proceedings of the 2018 Conference on Empirical Methods in
  Natural Language Processing}, pages 2775--2779, Brussels, Belgium.
  Association for Computational Linguistics.

\bibitem[{Mnih et~al.(2016)Mnih, Badia, Mirza, Graves, Lillicrap, Harley,
  Silver, and Kavukcuoglu}]{Mnih2016}
Volodymyr Mnih, Adria~Puigdomenech Badia, Mehdi Mirza, Alex Graves, Timothy
  Lillicrap, Tim Harley, David Silver, and Koray Kavukcuoglu. 2016.
\newblock Asynchronous methods for deep reinforcement learning.
\newblock In \emph{International conference on machine learning}, pages
  1928--1937.

\bibitem[{Murugesan et~al.(2021)Murugesan, Atzeni, Kapanipathi, Shukla,
  Kumaravel, Tesauro, Talamadupula, Sachan, and
  Campbell}]{murugesan2021textbased}
Keerthiram Murugesan, Mattia Atzeni, Pavan Kapanipathi, Pushkar Shukla, Sadhana
  Kumaravel, Gerald Tesauro, Kartik Talamadupula, Mrinmaya Sachan, and Murray
  Campbell. 2021.
\newblock {Text-based RL Agents with Commonsense Knowledge: New Challenges,
  Environments and Baselines}.
\newblock In \emph{Thirty Fifth AAAI Conference on Artificial Intelligence}.

\bibitem[{Murugesan et~al.(2020)Murugesan, Atzeni, Shukla, Sachan, Kapanipathi,
  and Talamadupula}]{Murugesan2020}
Keerthiram Murugesan, Mattia Atzeni, Pushkar Shukla, Mrinmaya Sachan, Pavan
  Kapanipathi, and Kartik Talamadupula. 2020.
\newblock Enhancing text-based reinforcement learning agents with commonsense
  knowledge.
\newblock \emph{arXiv preprint arXiv:2005.00811}.

\bibitem[{Narvekar et~al.(2020)Narvekar, Peng, Leonetti, Sinapov, Taylor, and
  Stone}]{Narvekar2020}
Sanmit Narvekar, Bei Peng, Matteo Leonetti, Jivko Sinapov, Matthew~E Taylor,
  and Peter Stone. 2020.
\newblock Curriculum learning for reinforcement learning domains: A framework
  and survey.
\newblock \emph{Journal of Machine Learning Research}, 21(181):1--50.

\bibitem[{Papineni et~al.(2002)Papineni, Roukos, Ward, and
  Zhu}]{papineni-etal-2002-bleu}
Kishore Papineni, Salim Roukos, Todd Ward, and Wei-Jing Zhu. 2002.
\newblock \href {https://doi.org/10.3115/1073083.1073135} {{B}leu: a method for
  automatic evaluation of machine translation}.
\newblock In \emph{Proceedings of the 40th Annual Meeting of the Association
  for Computational Linguistics}, pages 311--318, Philadelphia, Pennsylvania,
  USA. Association for Computational Linguistics.

\bibitem[{Pietquin et~al.(2011)Pietquin, Geist, Chandramohan, and
  Frezza-Buet}]{Pietquin2011}
Olivier Pietquin, Matthieu Geist, Senthilkumar Chandramohan, and Herv{\'e}
  Frezza-Buet. 2011.
\newblock Sample-efficient batch reinforcement learning for dialogue management
  optimization.
\newblock \emph{ACM Transactions on Speech and Language Processing (TSLP)},
  7(3):7.

\bibitem[{Portelas et~al.(2020)Portelas, Colas, Hofmann, and
  Oudeyer}]{Portelas2020}
R{\'e}my Portelas, C{\'e}dric Colas, Katja Hofmann, and Pierre-Yves Oudeyer.
  2020.
\newblock Teacher algorithms for curriculum learning of deep rl in continuously
  parameterized environments.
\newblock In \emph{Conference on Robot Learning}, pages 835--853. PMLR.

\bibitem[{Prabhumoye et~al.(2020)Prabhumoye, Li, Urbanek, Dinan, Kiela, Weston,
  and Szlam}]{Prabhumoye2020}
Shrimai Prabhumoye, Margaret Li, Jack Urbanek, Emily Dinan, Douwe Kiela, Jason
  Weston, and Arthur Szlam. 2020.
\newblock I love your chain mail! making knights smile in a fantasy game world:
  Open-domain goal-orientated dialogue agents.
\newblock \emph{arXiv preprint arXiv:2002.02878}.

\bibitem[{Racaniere et~al.(2019)Racaniere, Lampinen, Santoro, Reichert, Firoiu,
  and Lillicrap}]{Racaniere2019}
Sebastien Racaniere, Andrew~K Lampinen, Adam Santoro, David~P Reichert, Vlad
  Firoiu, and Timothy~P Lillicrap. 2019.
\newblock Automated curricula through setter-solver interactions.
\newblock \emph{arXiv preprint arXiv:1909.12892}.

\bibitem[{Risi and Togelius(2019)}]{Risi2019}
Sebastian Risi and Julian Togelius. 2019.
\newblock Procedural content generation: From automatically generating game
  levels to increasing generality in machine learning.
\newblock \emph{arXiv preprint arXiv:1911.13071}.

\bibitem[{Roller et~al.(2020)Roller, Dinan, Goyal, Ju, Williamson, Liu, Xu,
  Ott, Shuster, Smith et~al.}]{Roller2020}
Stephen Roller, Emily Dinan, Naman Goyal, Da~Ju, Mary Williamson, Yinhan Liu,
  Jing Xu, Myle Ott, Kurt Shuster, Eric~M Smith, et~al. 2020.
\newblock Recipes for building an open-domain chatbot.
\newblock \emph{arXiv preprint arXiv:2004.13637}.

\bibitem[{Samvelyan et~al.(2021)Samvelyan, Kirk, Kurin, Parker-Holder, Jiang,
  Hambro, Petroni, Kuttler, Grefenstette, and
  Rockt{\"a}schel}]{samvelyan2021minihack}
Mikayel Samvelyan, Robert Kirk, Vitaly Kurin, Jack Parker-Holder, Minqi Jiang,
  Eric Hambro, Fabio Petroni, Heinrich Kuttler, Edward Grefenstette, and Tim
  Rockt{\"a}schel. 2021.
\newblock \href {https://openreview.net/forum?id=skFwlyefkWJ} {Minihack the
  planet: A sandbox for open-ended reinforcement learning research}.
\newblock In \emph{Thirty-fifth Conference on Neural Information Processing
  Systems Datasets and Benchmarks Track (Round 1)}.

\bibitem[{Sautier et~al.(2020)Sautier, Agravante, and Tatsubori}]{sautier2020}
Corentin Sautier, Don~Joven Agravante, and Michiaki Tatsubori. 2020.
\newblock \href {https://kbrl.github.io/papers/08-KBRL.pdf} {{State Prediction
  in TextWorld with a Predicate-Logic Pointer Network Architecture}}.
\newblock In \emph{In Workshop on Knowledge-based Reinforcment Learning at
  IJCAI-20}.

\bibitem[{Schmidhuber(2013)}]{Schmidhuber2013}
J{\"u}rgen Schmidhuber. 2013.
\newblock Powerplay: Training an increasingly general problem solver by
  continually searching for the simplest still unsolvable problem.
\newblock \emph{Frontiers in psychology}, 4:313.

\bibitem[{Short and Adams(2017)}]{Short2017}
Tanya Short and Tarn Adams. 2017.
\newblock \emph{Procedural generation in game design}.
\newblock CRC Press.

\bibitem[{Singh et~al.(2000)Singh, Kearns, Litman, and Walker}]{Singh2000}
Satinder~P Singh, Michael~J Kearns, Diane~J Litman, and Marilyn~A Walker. 2000.
\newblock Reinforcement learning for spoken dialogue systems.
\newblock In \emph{Advances in Neural Information Processing Systems}, pages
  956--962.

\bibitem[{Sukhbaatar et~al.(2018)Sukhbaatar, Lin, Kostrikov, Synnaeve, Szlam,
  and Fergus}]{Sukhbaatar2018}
Sainbayar Sukhbaatar, Zeming Lin, Ilya Kostrikov, Gabriel Synnaeve, Arthur
  Szlam, and Rob Fergus. 2018.
\newblock \href {https://openreview.net/forum?id=SkT5Yg-RZ} {Intrinsic
  motivation and automatic curricula via asymmetric self-play}.
\newblock In \emph{International Conference on Learning Representations}.

\bibitem[{Urbanek et~al.(2019)Urbanek, Fan, Karamcheti, Jain, Humeau, Dinan,
  Rocktäschel, Kiela, Szlam, and Weston}]{Urbanek2019}
Jack Urbanek, Angela Fan, Siddharth Karamcheti, Saachi Jain, Samuel Humeau,
  Emily Dinan, Tim Rocktäschel, Douwe Kiela, Arthur Szlam, and Jason Weston.
  2019.
\newblock Learning to speak and act in a fantasy text adventure game.
\newblock In \emph{Empirical Methods in Natural Language Processing (EMNLP)}.

\bibitem[{{Wang} et~al.(2017){Wang}, {Mao}, {Wang}, and {Guo}}]{Wang2017}
Q.~{Wang}, Z.~{Mao}, B.~{Wang}, and L.~{Guo}. 2017.
\newblock Knowledge graph embedding: A survey of approaches and applications.
\newblock \emph{IEEE Transactions on Knowledge and Data Engineering},
  29(12):2724--2743.

\bibitem[{Wu et~al.(2018)Wu, Fisch, Chopra, Adams, Bordes, and Weston}]{Wu2018}
Ledell Wu, Adam Fisch, Sumit Chopra, Keith Adams, Antoine Bordes, and Jason
  Weston. 2018.
\newblock Starspace: Embed all the things!
\newblock In \emph{Proceedings of the AAAI Conference on Artificial
  Intelligence}, volume~32.

\bibitem[{Yang et~al.(2018)Yang, Yuan, Cer, Kong, Constant, Pilar, Ge, Sung,
  Strope, and Kurzweil}]{Yang2018}
Yinfei Yang, Steve Yuan, Daniel Cer, Sheng-Yi Kong, Noah Constant, Petr Pilar,
  Heming Ge, Yun-Hsuan Sung, Brian Strope, and Ray Kurzweil. 2018.
\newblock Learning semantic textual similarity from conversations.
\newblock \emph{arXiv preprint arXiv:1804.07754}.

\bibitem[{Yarats and Lewis(2017)}]{Yarats2017}
Denis Yarats and Mike Lewis. 2017.
\newblock Hierarchical text generation and planning for strategic dialogue.
\newblock \emph{arXiv preprint arXiv:1712.05846}.

\bibitem[{Zahavy et~al.(2018)Zahavy, Haroush, Merlis, Mankowitz, and
  Mannor}]{Zahavy2018}
Tom Zahavy, Matan Haroush, Nadav Merlis, Daniel~J Mankowitz, and Shie Mannor.
  2018.
\newblock Learn what not to learn: Action elimination with deep reinforcement
  learning.
\newblock In S.~Bengio, H.~Wallach, H.~Larochelle, K.~Grauman, N.~Cesa-Bianchi,
  and R.~Garnett, editors, \emph{Advances in Neural Information Processing
  Systems 31}, pages 3562--3573. Curran Associates, Inc.

\end{thebibliography}
\bibliographystyle{acl_natbib}

\clearpage

\appendix

\section{Appendix}
\label{sec:appendix}
\begin{figure*}
    \centering
    \includegraphics[width=.8\linewidth]{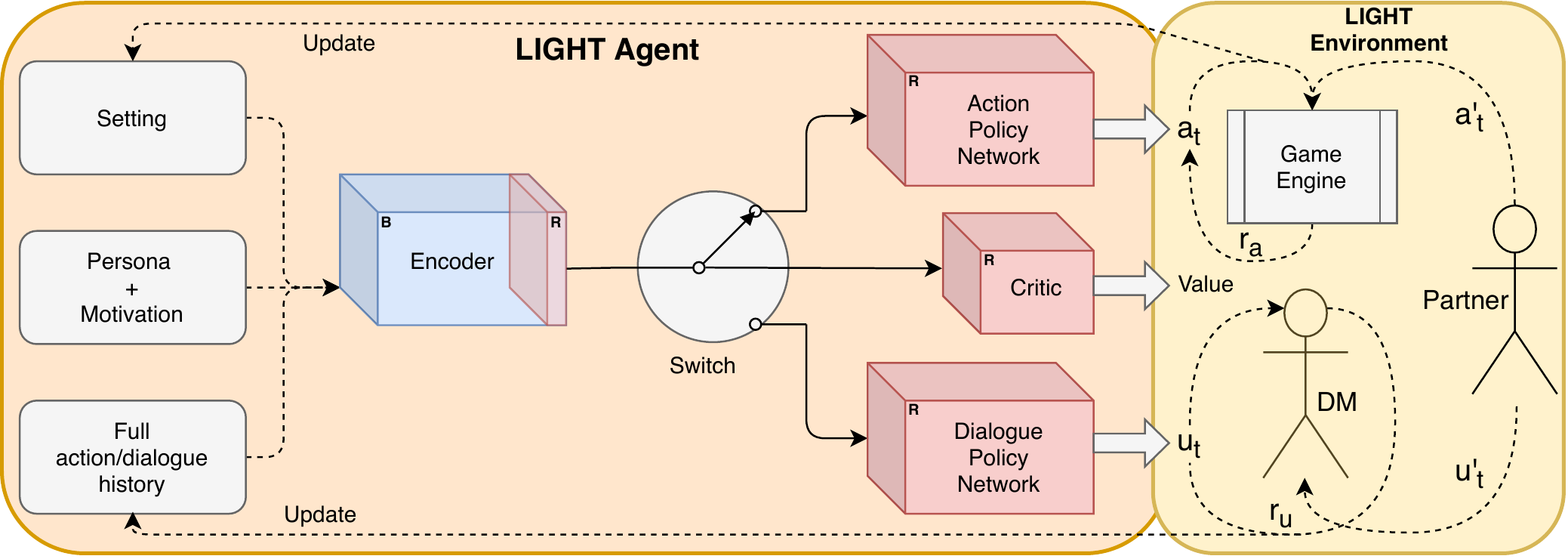}
    \caption{Expanded overall architecture and training pipeline diagram for the LIGHT RL Agent~\citep{Ammanabrolu2021}.}
    \label{fig:lightrlcurr}
\end{figure*}
\subsection{LIGHT Environment Details}
\label{app:lightdetails}

Formally, we adapt the definition of text-based games as seen in~\cite{Cote2018,Hausknecht2020} to LIGHT. 
They are partially observable Markov decision processes (POMDPs), represented as a 7-tuple of $\langle S,T,\mathcal{A},\Omega , O,\mathcal{R}, \gamma\rangle$ representing the set of environment states, conditional transition probabilities between states, the vocabulary or words used to compose action commands or dialogue utterances (e.g. {\em get sword} or {\em Hey, give me that sword!} respectively), observations returned by the game, observation conditional probabilities, reward function, and the discount factor respectively.

There are $5982$ training, $756$ validation, and $748$ test quests.
The average sequence of a human demonstration is $12.92$, with an average action sequence length of $2.18$ and dialogue of $10.74$.
There are $1800$ training, $100$ validation, and $211$ test human expert demonstrations corresponding to the same splits as the quests themselves.

The LIGHT environment further allows us to factorize the overall action space $\mathcal{A}$ into $A$ as the set of possible textual actions or commands (e.g. {\em get sword, steal coins from merchant}), and $U$ as the set of possible dialogues that can be uttered by an agent, thus making it a factored POMDP~\citep{Degris2013}.
This in turn means that, for a given quest $q$, each expert human demonstration  $\mathcal{D}(q)=\alpha^*_0,\alpha^*_1...\alpha^*_n$ can be factorized into two sub-sequences of expert demonstrations of actions and dialogue $\mathcal{D}_{A}(q)=a^*_0,a^*_1,...a^*_n$ and $\mathcal{D}_{U}(q)=u^*_0,u^*_1,...u^*_m$ respectively.
The factorized action spaces $A$ and $U$ are constructed by enumerating all possible actions/dialogue utterances in the all human demonstrations in LIGHT-quests. 

Figure~\ref{fig:lightrlcurr} shows the overall architecture and training pipeline---our reinforcement learning pipeline is unchanged from that shown in \citet{Ammanabrolu2021} with the exception of the curriculum of quests performed by the agent and the way the speech rewards are designed.
An encoder first takes in information about setting, persona, motivation for a single character then passes it onto a switch module.
This switch module is a meta policy that decides if an agent should act or talk and is trained to mimic how often human experts act or talk while performing quests via demonstrations.
Two separate policy networks make a decision on which action to perform or dialogue to say given the current context and a single shared critic attempts to measure the value of taking an action in a particular state.

Once an agent acts or talks, the partner agent---in this case also a polyencoder~\citep{Humeau2020} trained to react to agents with motivations---also acts or talks and this information is processed by the environment.
As recommended by \citet{Prabhumoye2020,Ammanabrolu2021}, we keep the partner model fixed during the episodes where the LIGHT agent trains to ensure that it retains natural English semantics---avoiding the problem of language drift by learning an emergent language with that must agree with the partner's usage~\citep{Lee2019}.

\textbf{A2C Training.}
Each parallel A2C agent samples from the the current pool of available quests---i.e. the curriculum---for a fixed number of steps $k$ before switching to the quest pool corresponding to the next higher level difficulty curriculum.
The initial pool of quests is the training set of LIGHT-Quests as seen in \citet{Ammanabrolu2021} and all pools after that correspond to decreasing values of $n$ used when generating the curriculums (as seen in Figure~\ref{fig:generatedcurrnoun}). 

\textbf{Rewards.}
Following \citet{Ammanabrolu2021}, we use a learned model--the Dungeon Master (DM)---to score the agent's ability to speak.
The DM used here is a poly-encoder model trained on collected human quest demonstrations as well as the original conversations in LIGHT. 
It is conditioned on quests and motivations and thus able to provide a (noisy) indication of how natural the agent's dialogue utterances are given its immediate context, similarly to the function of the DM during data collection.

Given the dialogue portion of a human quest demonstration $\mathcal{D}_{U}(q)=u^*_0,u^*_1,...u^*_n$, 
of length $n$, the DM returns a reward $r_u$ of $\frac{1}{2n}$ if
an utterance was in the demonstration $u \in \mathcal{D}_{U}(q)$
(for a maximum of one time per episode for each utterance from the demonstration).
A further $\frac{1}{2n}$ is given each time the utterance is scored as being within the top-$k$ most likely utterances by the DM.
The original quests all have human demonstrations but the procedurally generated ones do not.
During training, in cases where a particular LIGHT game instance does not have corresponding human demonstration, only the latter reward resulting from an utterance being within the top-$k$ most likely utterances by the DM is used.
This naturalness objective will be hence referred to as a {\em speech goal}.
These rewards thus also denser than {\em act goals}, helping the agent learn overall.
Further, similarly to the game engine, the DM also provides a set of $M$ valid utterances
which are the $M$ most likely dialogue candidates from the candidate set for the current context.

\subsection{Encoder Pre-training Tasks}
Here, we summarize the pre-training tasks for the encoders mentioned in Section~\ref{sec:curreval}.
These tasks are unchanged from those described in \citet{Ammanabrolu2021}.

\textbf{ATOMIC-LIGHT.} 
ATOMIC-LIGHT is a (domain-adapted) fantasy commonsense knowledge graph, and as such provides priors for an agent on how to act consistently in the world.
For example, given a clause such as
``The knight wishes to slay the dragon, as a result the knight {\em needs} to \underline{acquire a sword},'' the task would be to predict the underlined text---a form of knowledge graph completion~\citep{Wang2017}.

\textbf{Reddit.} A further tuning dataset is derived from an existing Reddit dataset, pushshift.io ~\citep{Baumgartner2020} as seen in \citet{Roller2020}.
This dataset has been used in several existing dialogue-based studies and has been shown to result in more natural conversations~\citep{Yang2018,Mazare2018}.

\textbf{LIGHT-Original.} 
The task itself dervied from the original LIGHT dataset~\citep{Urbanek2019} and involves predicting the next action or utterance given the prior dialogue history as well as the current setting and persona for a character.
They are collected in a chit-chat fashion, with no notion of objectives, and so provide priors on how to generally act consistently and speak in a fantasy world, but not directly how to complete quests.

\textbf{LIGHT-Quests.}
This dataset provides two pre-training tasks.
(1) {\em Bag-of-action timeline prediction} in which, given a quest consisting of setting, persona, and motivations, any one of the actions in the timeline must be predicted.
(2) {\em Sequential timeline prediction} in which, given a quest consisting of setting, persona, motivations, and the first $n$ actions in the timeline, the $n+1^{th}$ action must be predicted.
(3) Predict the next dialogue utterance given a human demonstration in a manner similar to the LIGHT-original tasks.

\subsection{Sampled and Randomly Generated Curriculum Distributions}
This section contains the verb and noun distributions for the {\em sampled} and {\em randomly generated} curriculums as described in Section~\ref{sec:curreval}, presented in the same fashion as Figure~\ref{fig:generatedcurrnoun}.

For the {\em randomly generated} curriculums, we present 5 different curriculums---varying the proportion of randomly generated quests per pool from 0\% (corresponding to the full procedurally generated pipeline), to 100\% randomly generated, in increments of 20\%.
Sections after this present ablation results after training agents on these curriculums to better analyze the effects of randomness and diversity in zero-shot generalization. 

\begin{figure}
\centering
\begin{minipage}{.49\textwidth}
    \centering
    \includegraphics[width=.8\linewidth]{figures/Short_Motivation_Generation_without_Distribution_Tuning_verb_normalized.pdf}
\end{minipage}
\begin{minipage}{.49\textwidth}
    \centering
    \includegraphics[width=.8\linewidth]{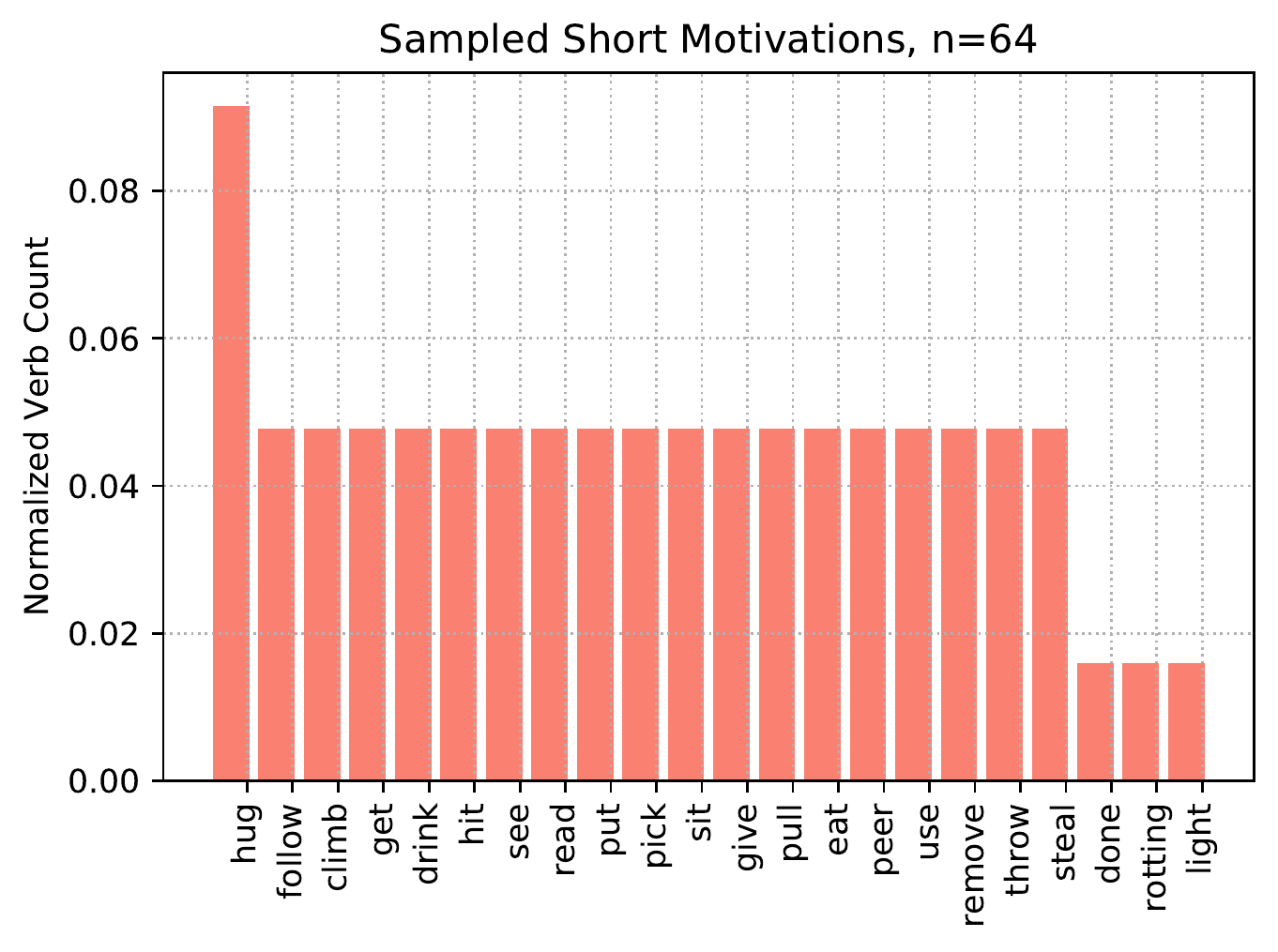}
\end{minipage}
\begin{minipage}{.49\textwidth}
    \centering
    \includegraphics[width=.8\linewidth]{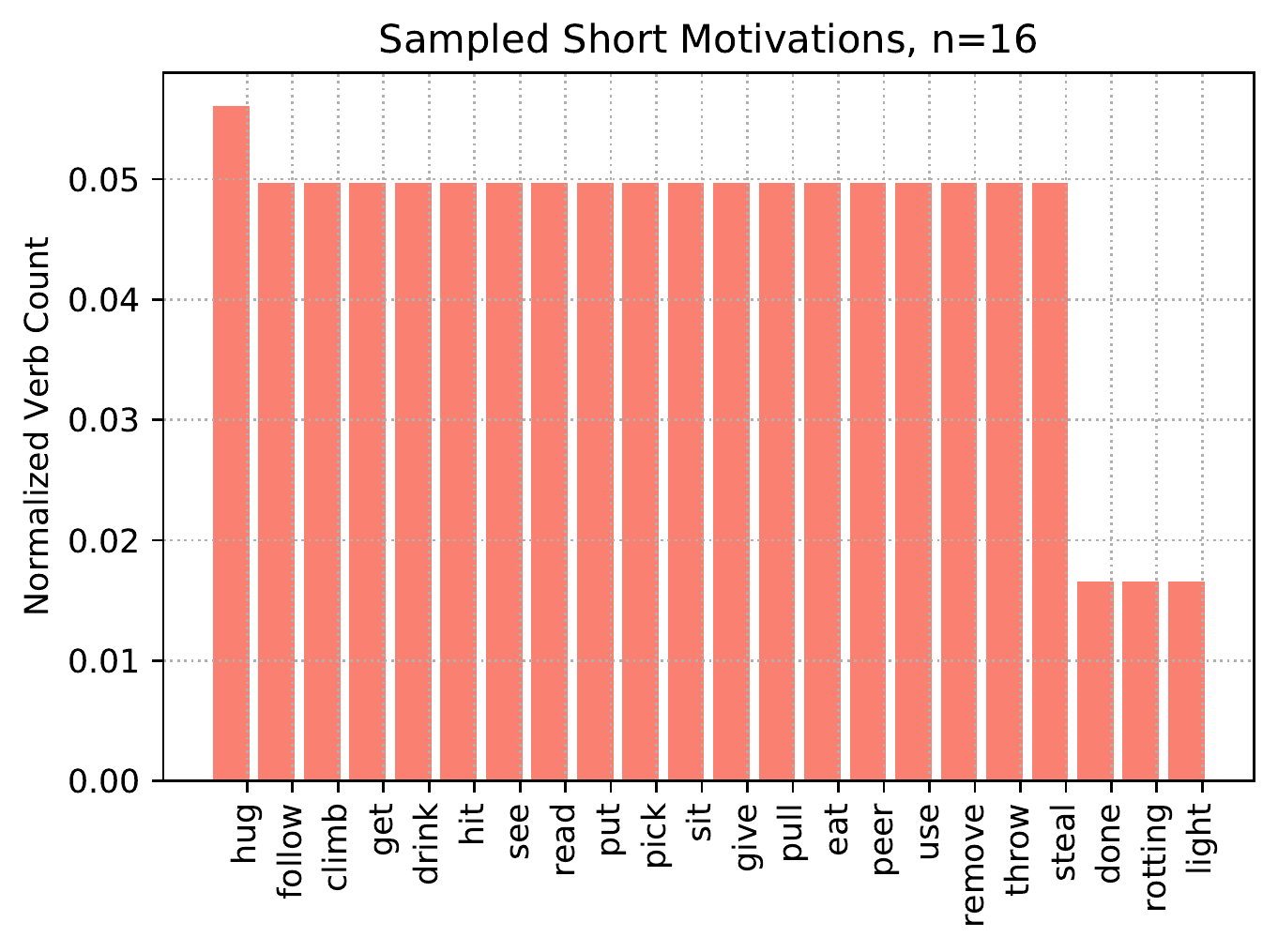}
\end{minipage}
\begin{minipage}{.49\textwidth}
    \centering
    \includegraphics[width=.8\linewidth]{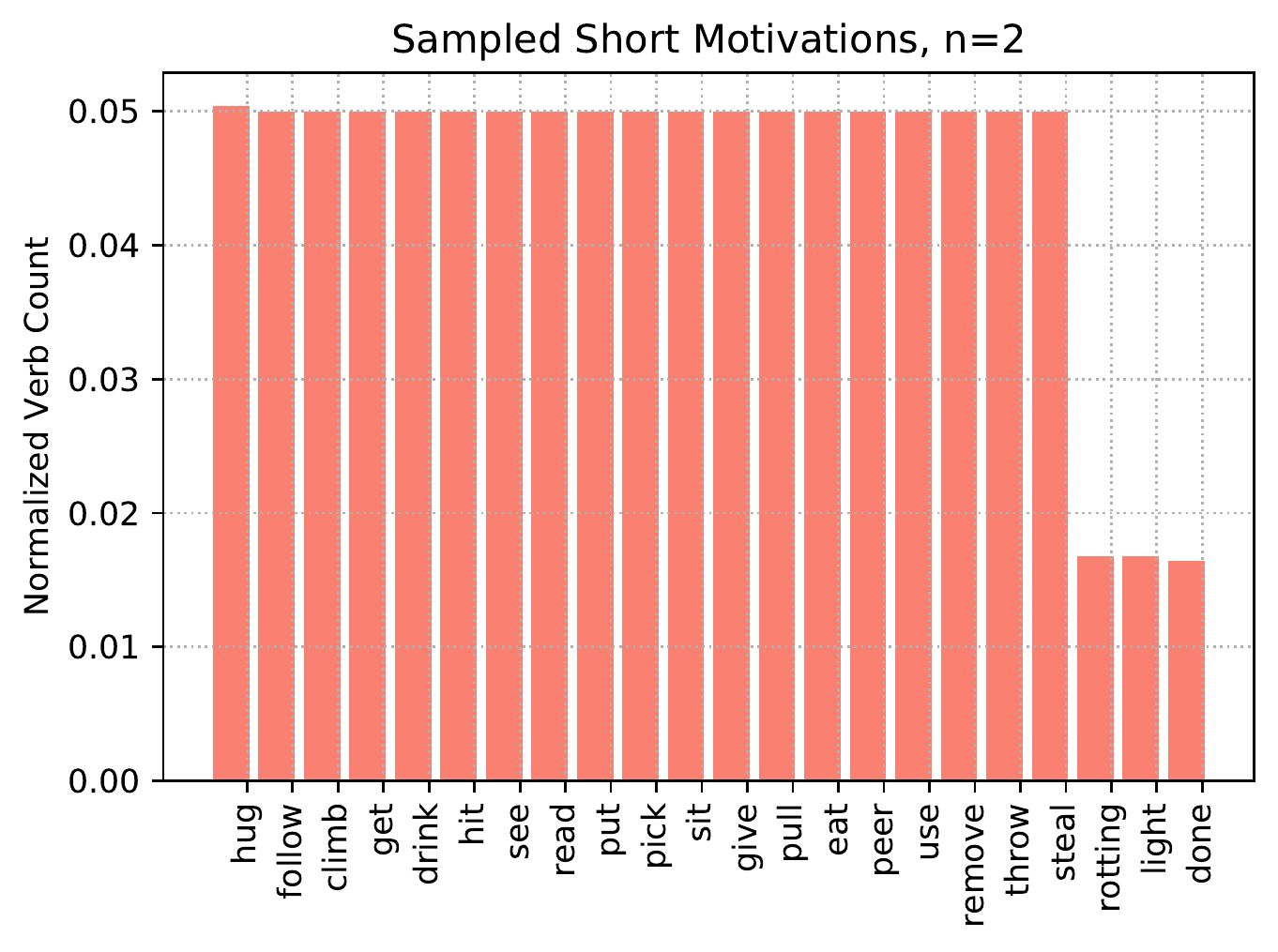}
\end{minipage}
\caption{Distribution of verbs in the short motivation of the curriculum of quests starting from the original distribution on top to the flattened and {\bf sampled} curriculum on the bottom as a function of $n$ (Section~\ref{sec:curriculumgen}). The y-axis of the different nouns reflect their relative proportion in the pool of quests.}
\label{fig:sampledcurr}
\end{figure}

\begin{figure}
\centering
\begin{minipage}{.49\textwidth}
    \centering
    \includegraphics[width=.8\linewidth]{figures/Short_Motivation_Generation_without_Distribution_Tuning_noun_normalized_top_20.pdf}
\end{minipage}
\begin{minipage}{.49\textwidth}
    \centering
    \includegraphics[width=.8\linewidth]{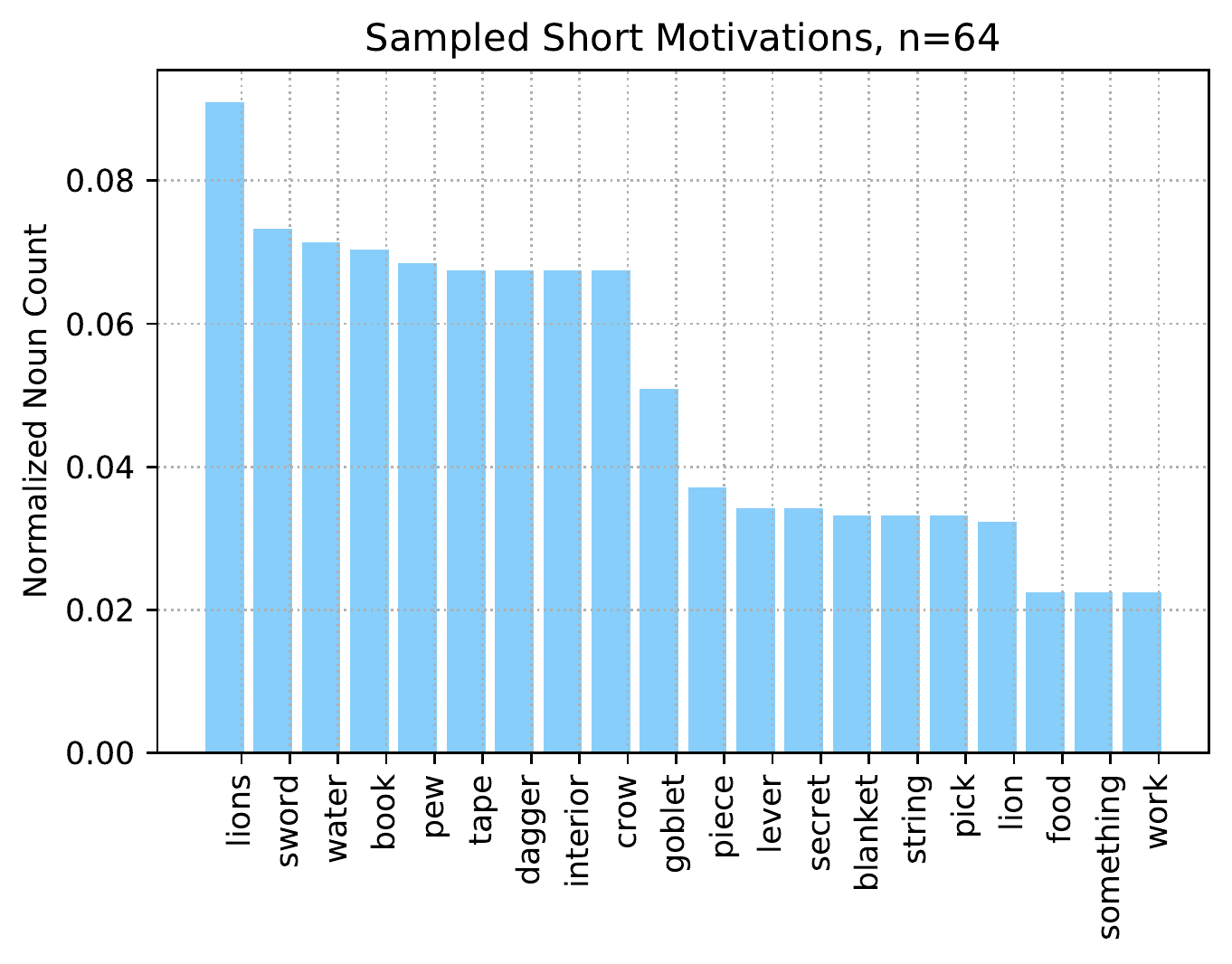}
\end{minipage}
\begin{minipage}{.49\textwidth}
    \centering
    \includegraphics[width=.8\linewidth]{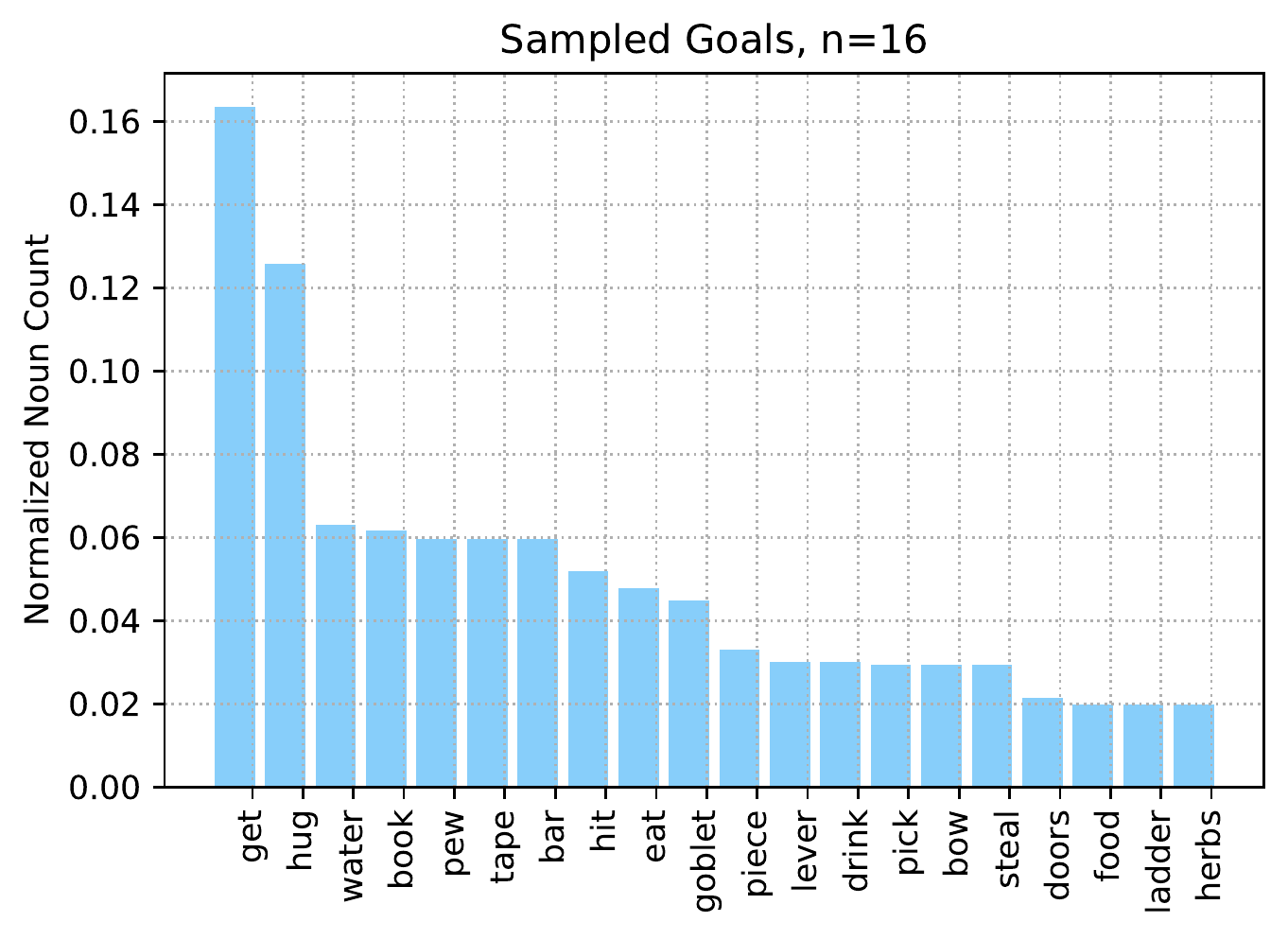}
\end{minipage}
\begin{minipage}{.49\textwidth}
    \centering
    \includegraphics[width=.8\linewidth]{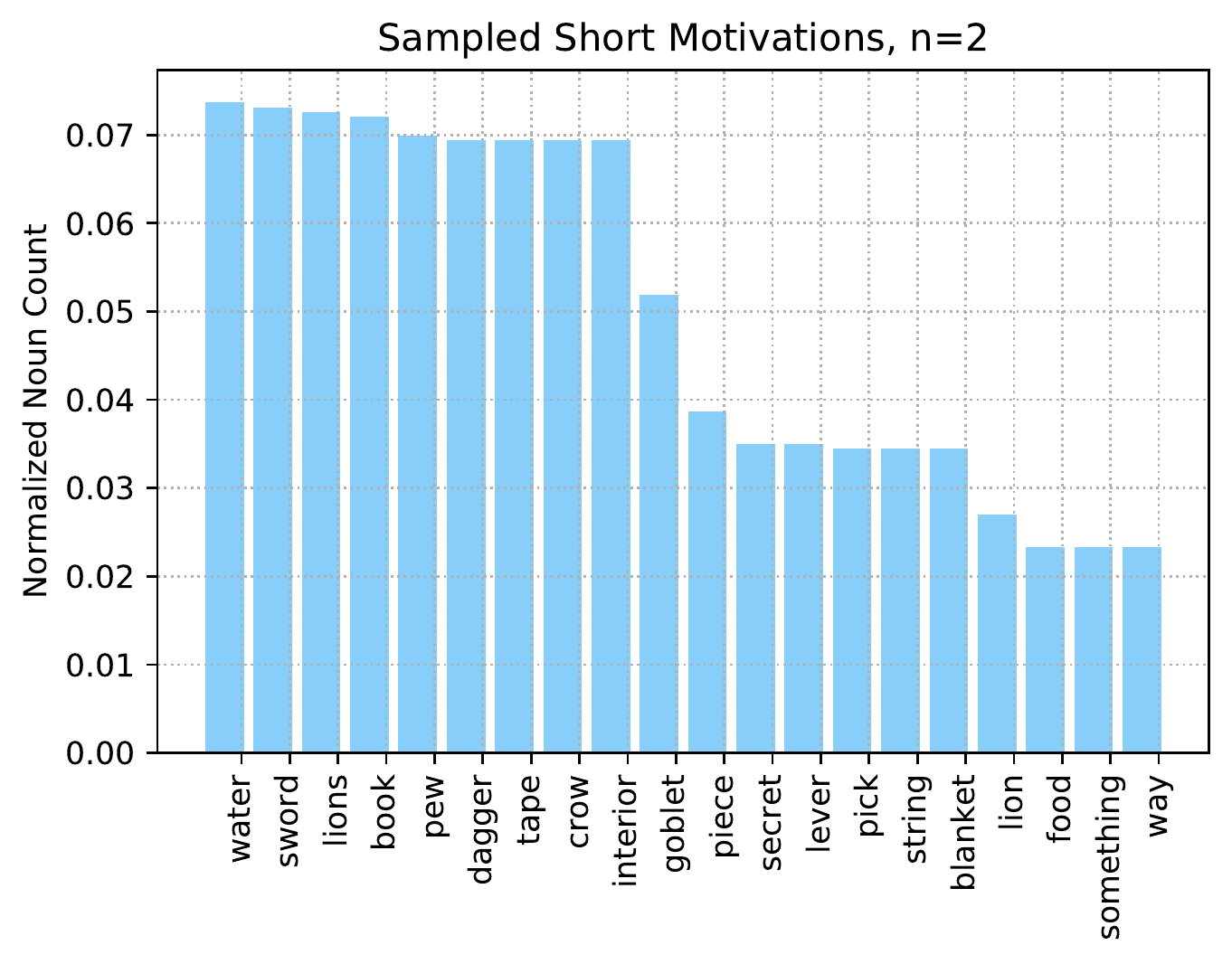}
\end{minipage}
\caption{Distribution of nouns in the short motivation of the curriculum of quests starting from the original distribution on top to the flattened and {\bf sampled} curriculum on the bottom as a function of $n$ (Section~\ref{sec:curriculumgen}). The y-axis of the different nouns reflect their relative proportion in the pool of quests.}
\label{fig:sampledcurrnoun}
\end{figure}

\newpage
\clearpage



\begin{figure*}
\begin{tabular}{cccc}
\subfloat{\includegraphics[width=0.225\linewidth]{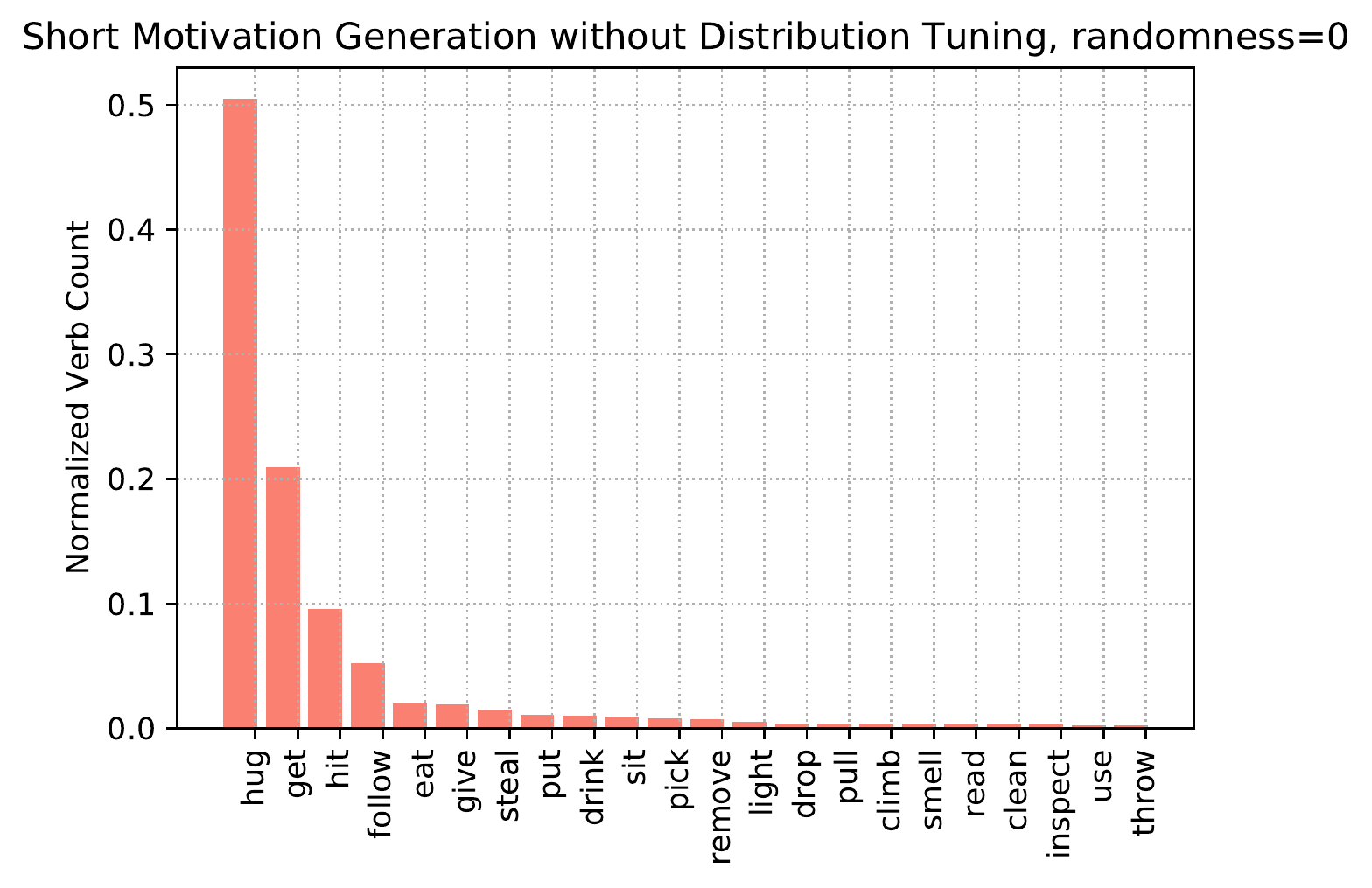}} &
\subfloat{\includegraphics[width=0.225\linewidth]{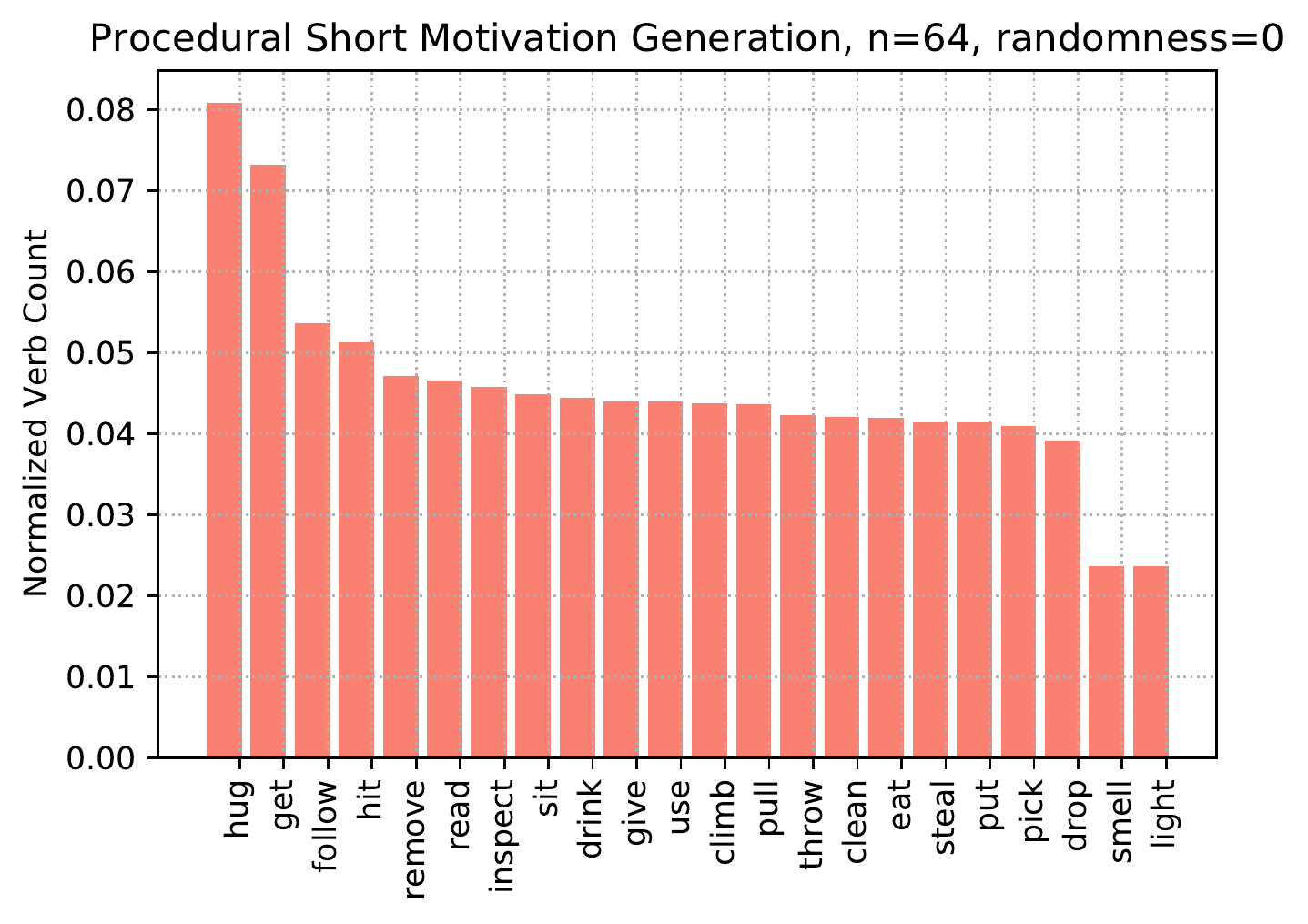}} &
\subfloat{\includegraphics[width=0.225\linewidth]{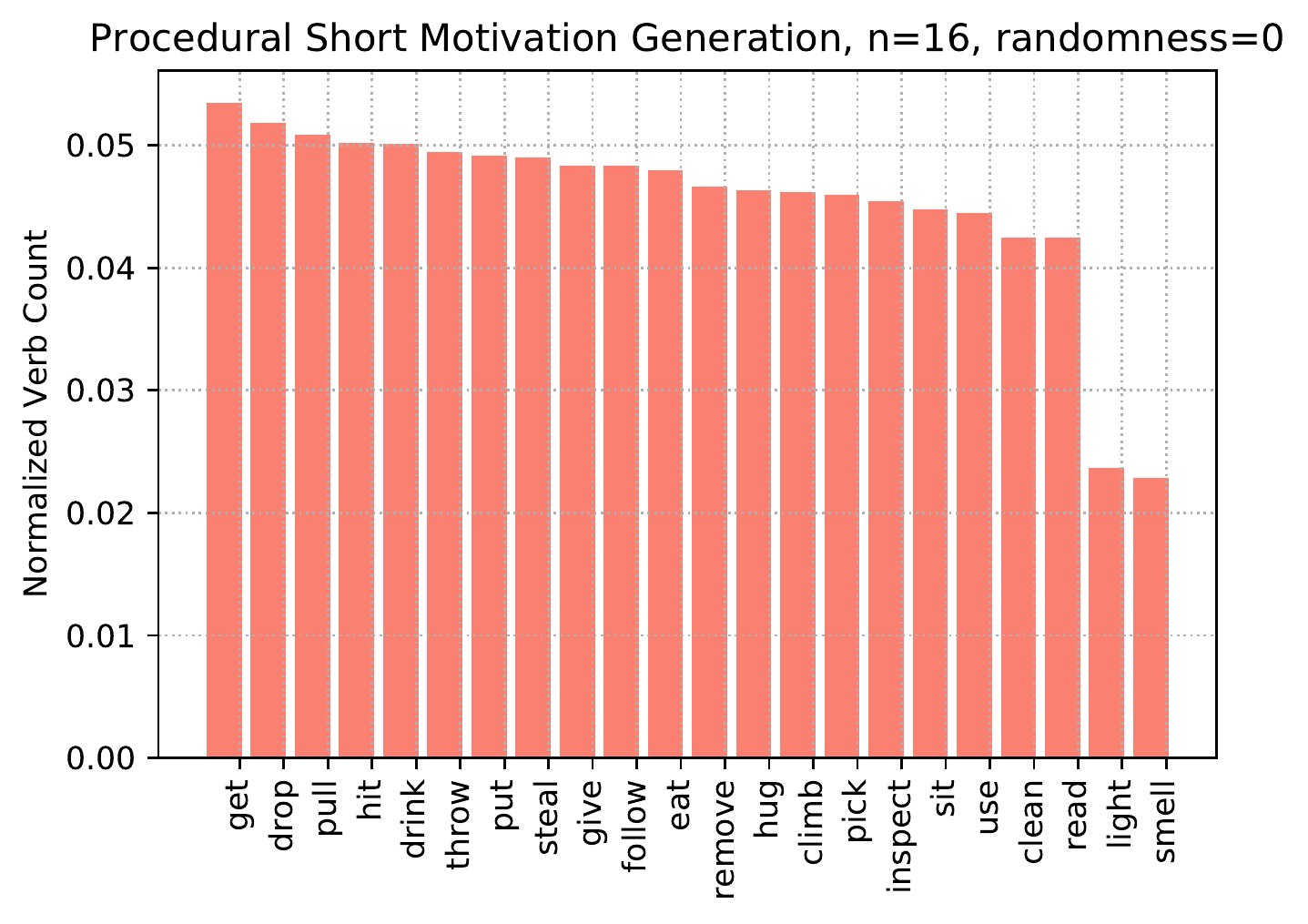}} &
\subfloat{\includegraphics[width=0.225\linewidth]{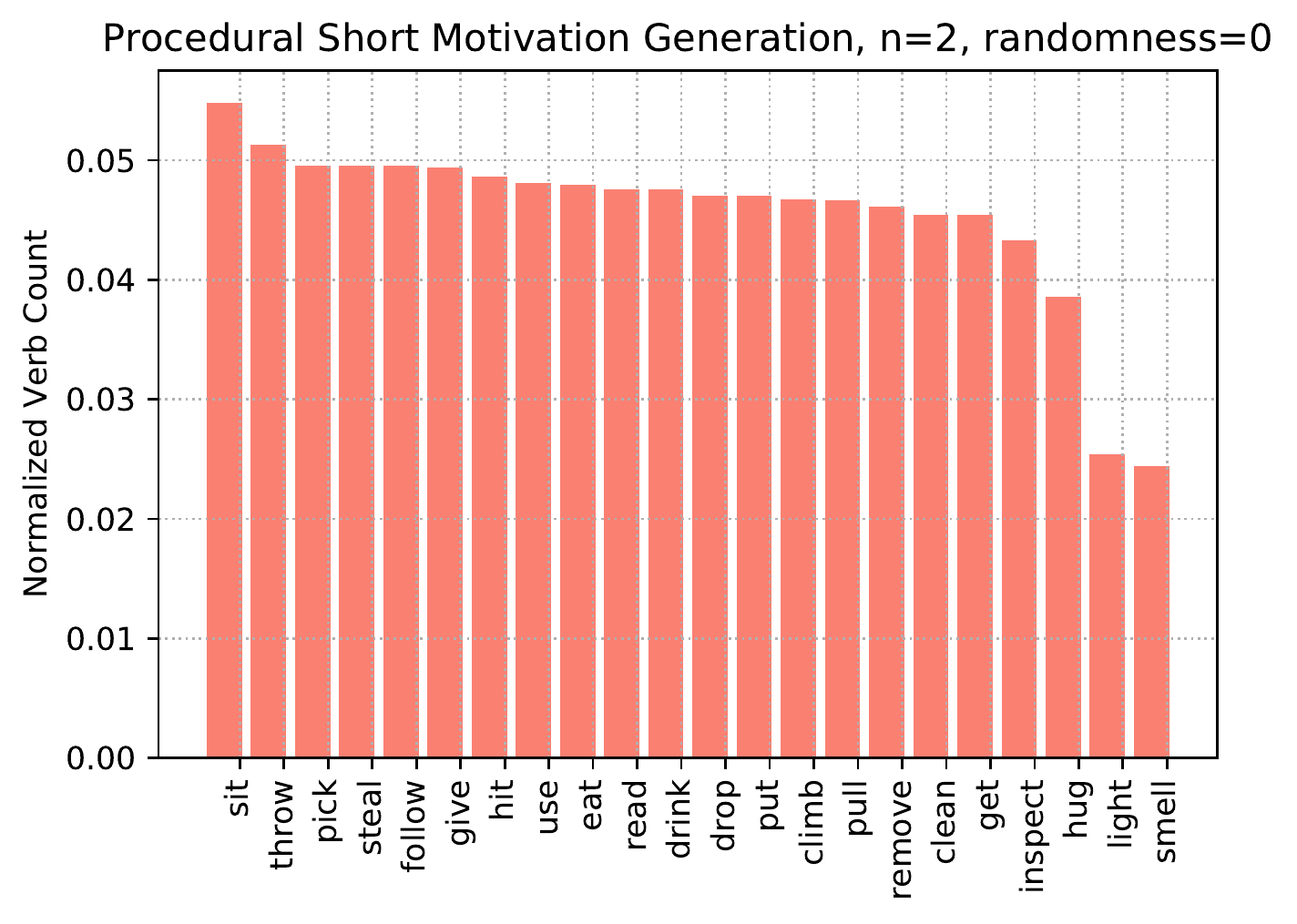}}\\
\subfloat{\includegraphics[width=0.225\linewidth]{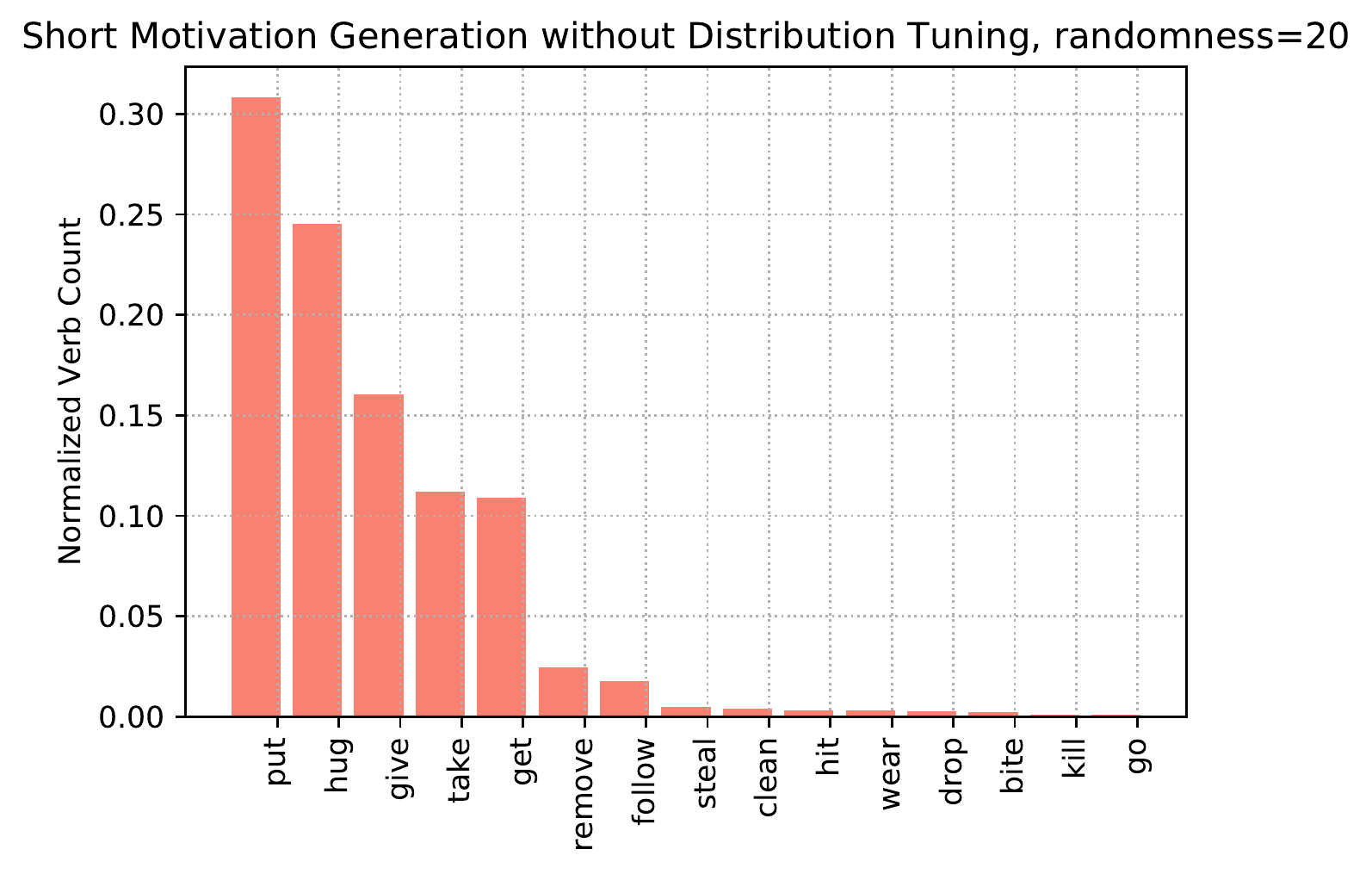}} &
\subfloat{\includegraphics[width=0.225\linewidth]{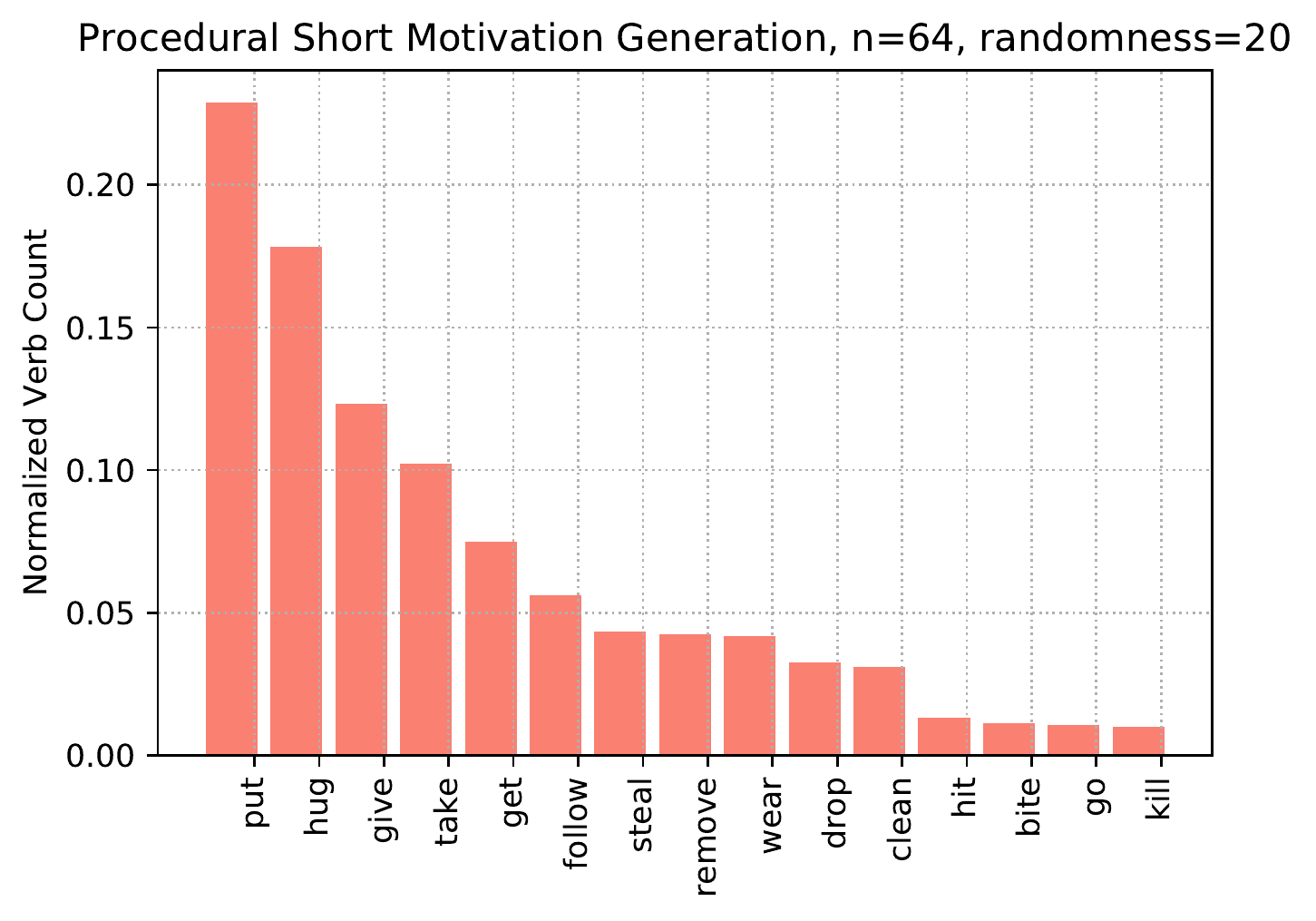}} &
\subfloat{\includegraphics[width=0.225\linewidth]{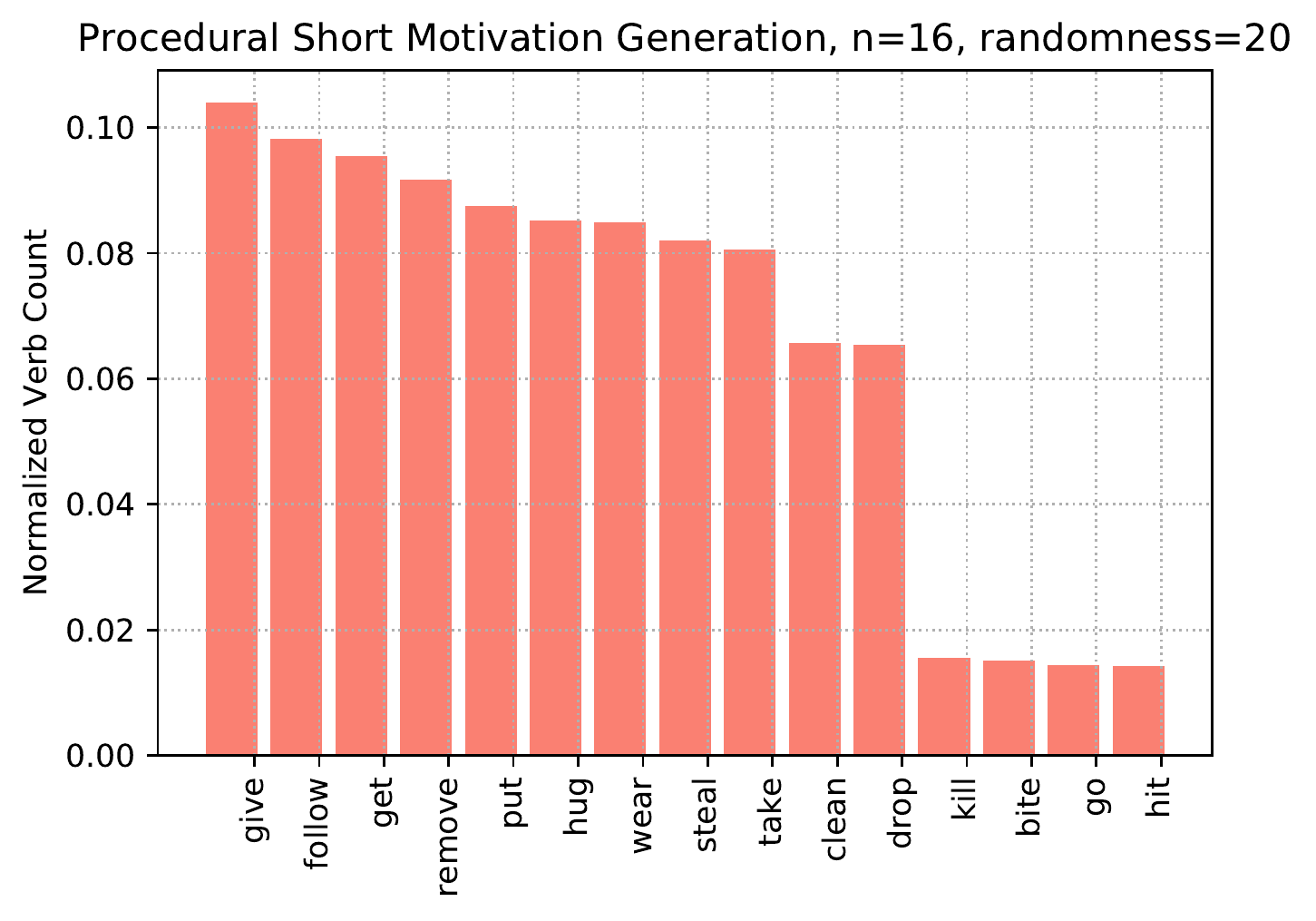}} &
\subfloat{\includegraphics[width=0.225\linewidth]{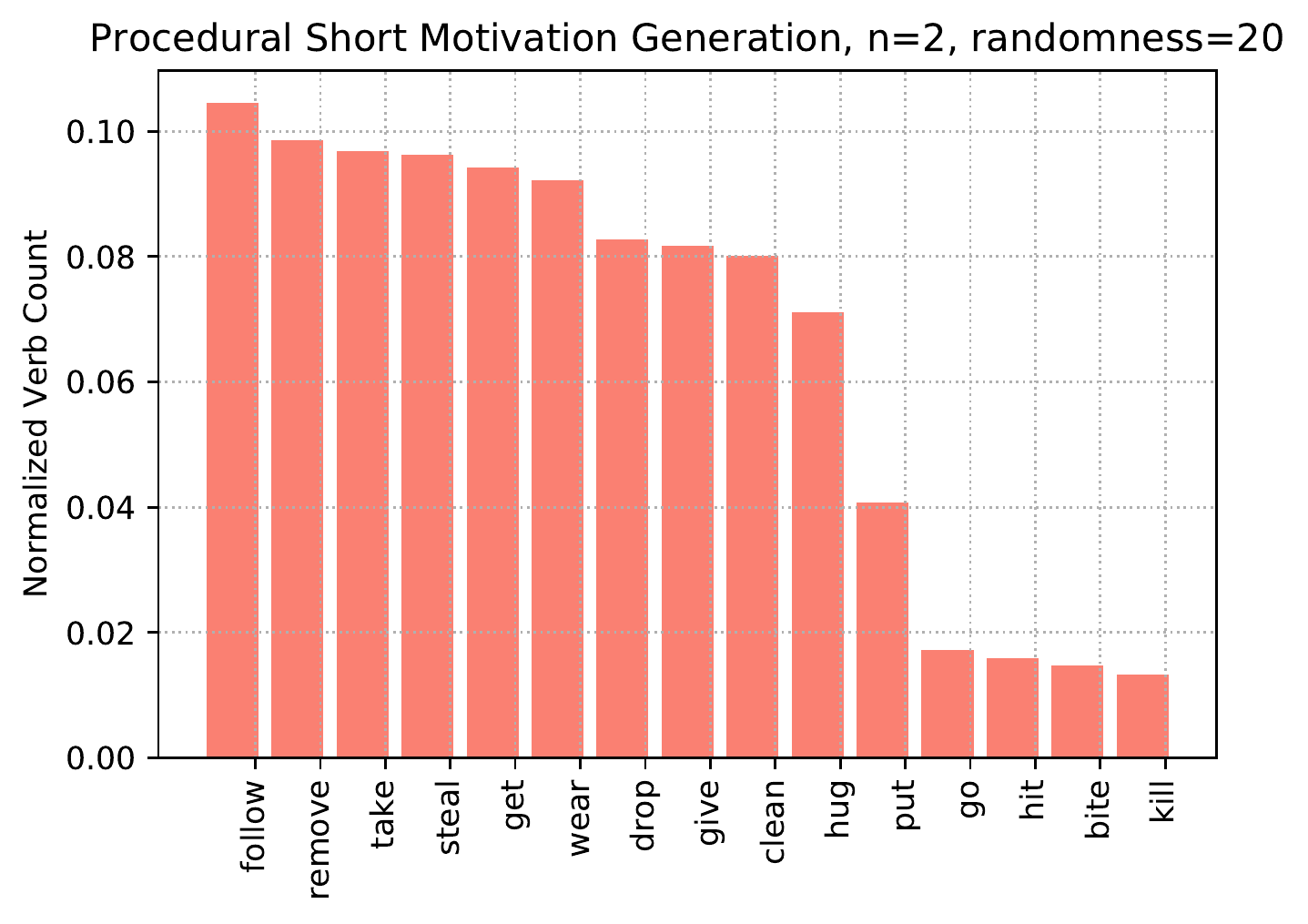}}\\
\subfloat{\includegraphics[width=0.225\linewidth]{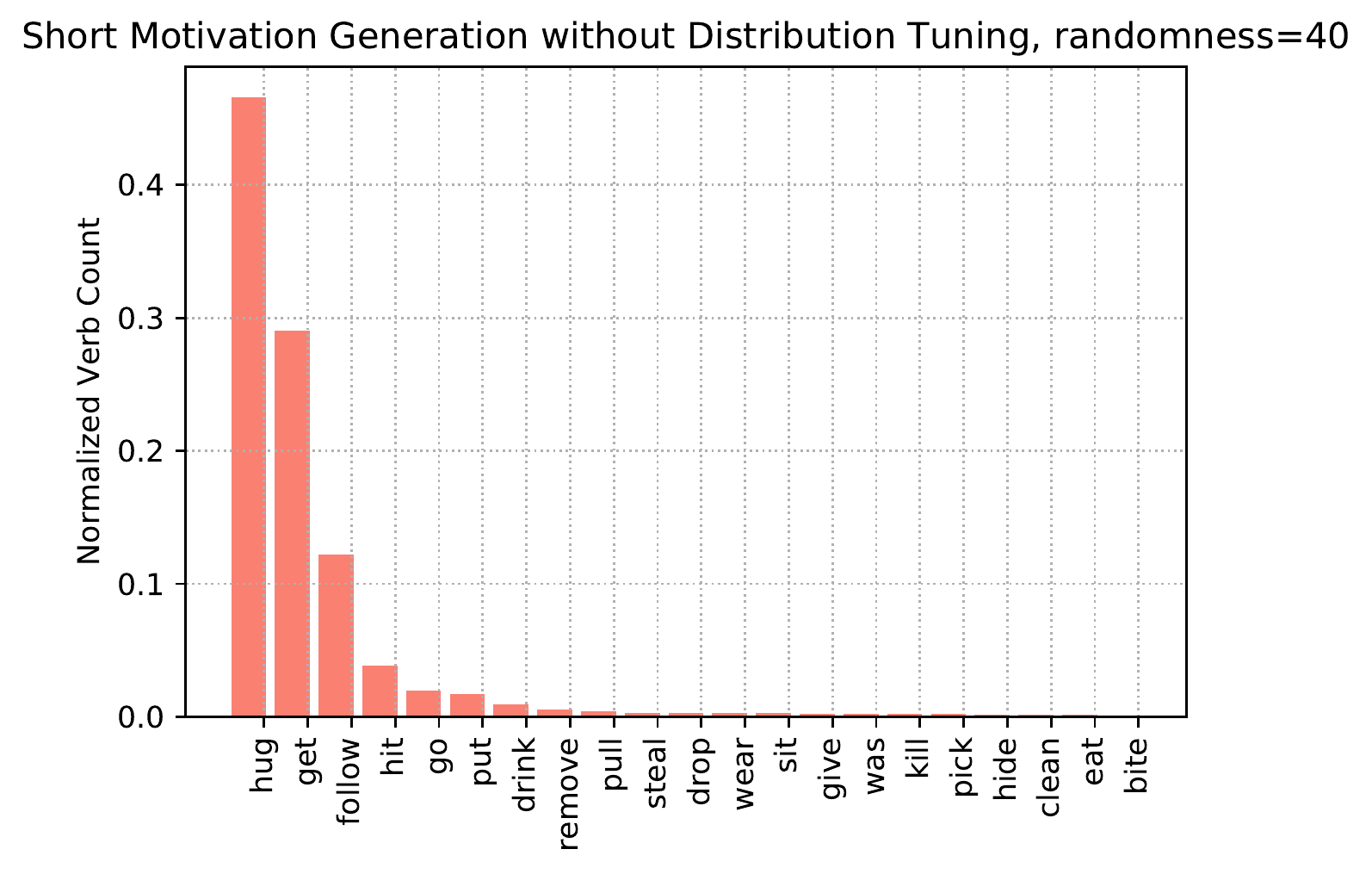}} &
\subfloat{\includegraphics[width=0.225\linewidth]{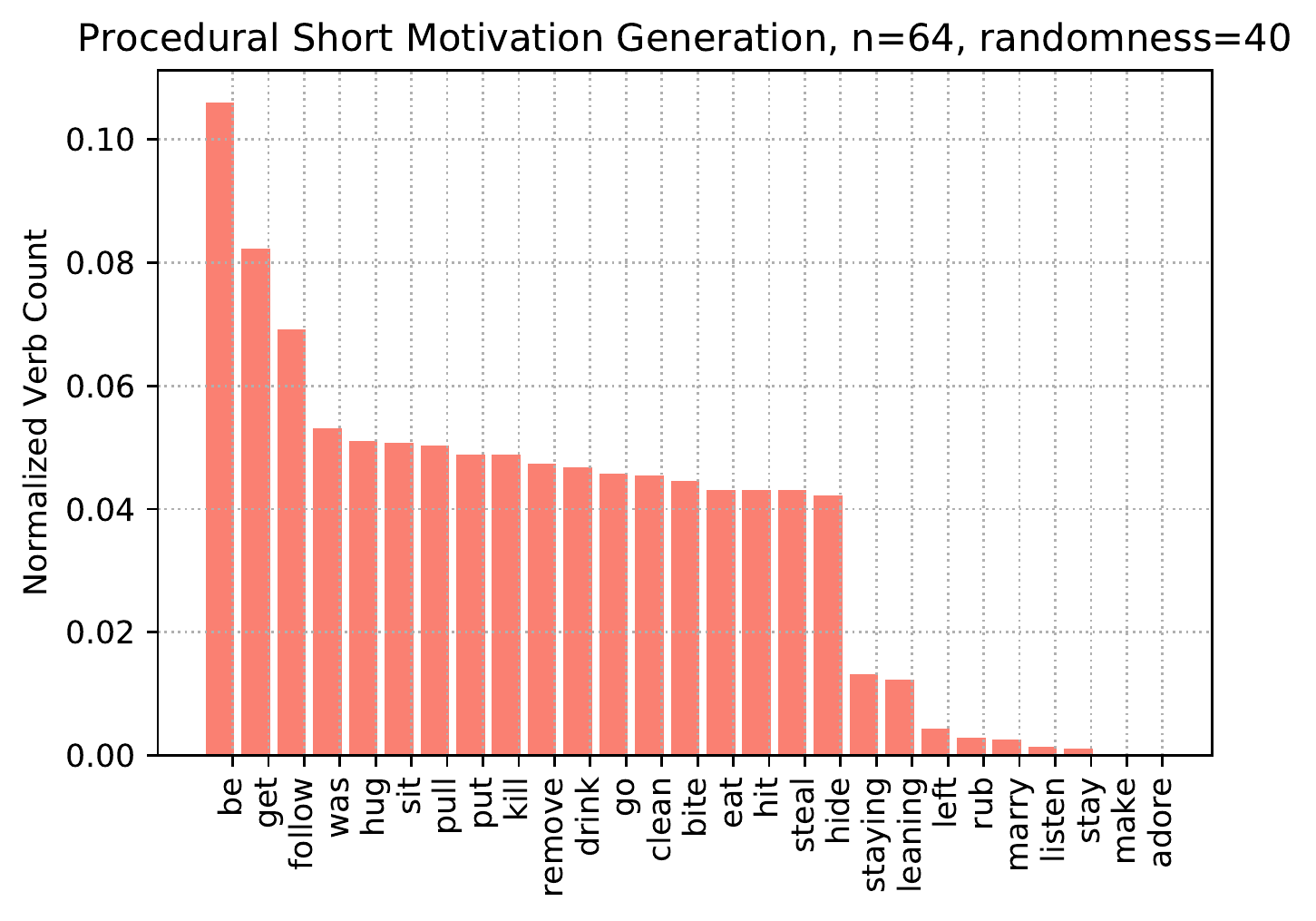}} &
\subfloat{\includegraphics[width=0.225\linewidth]{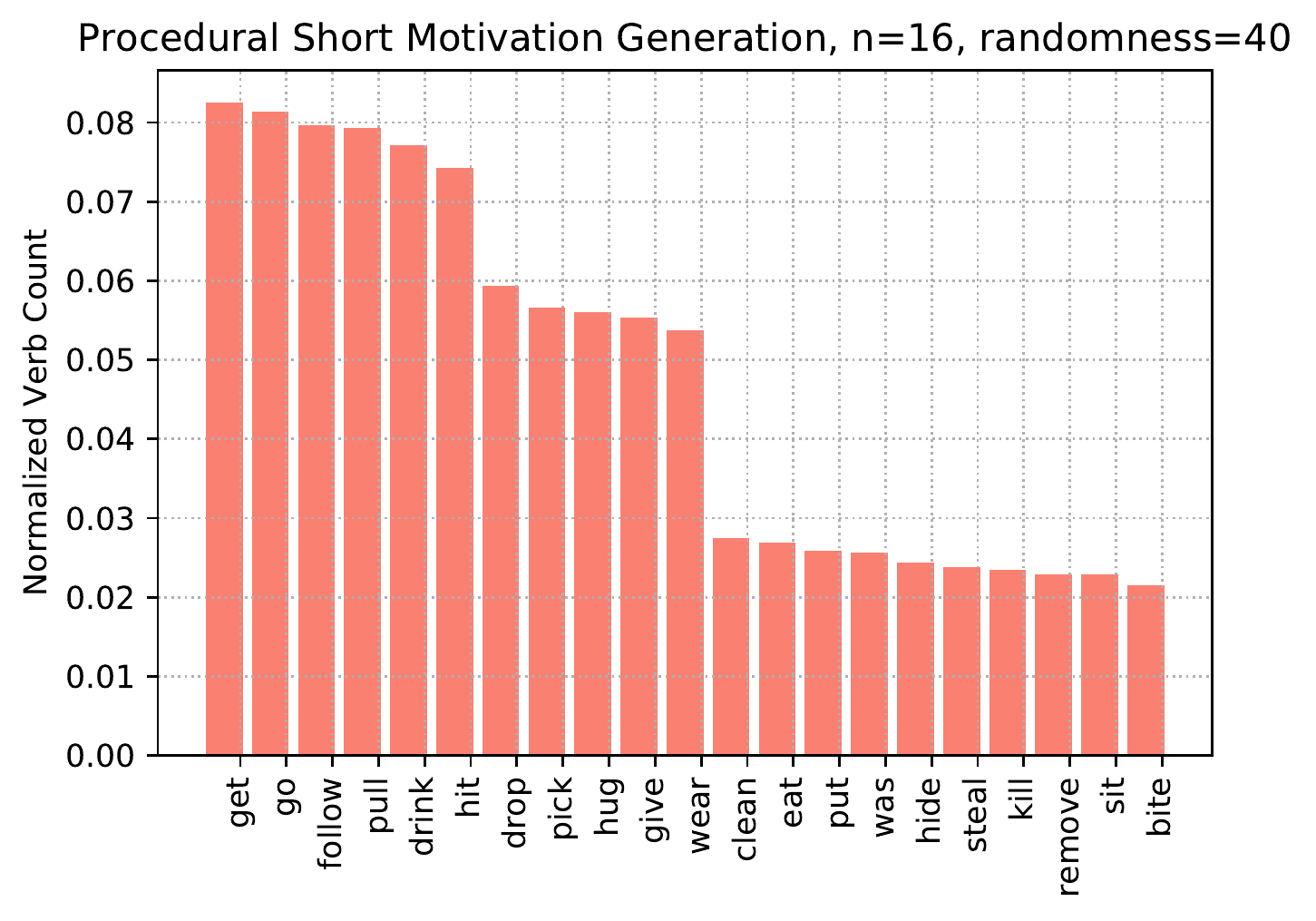}} &
\subfloat{\includegraphics[width=0.225\linewidth]{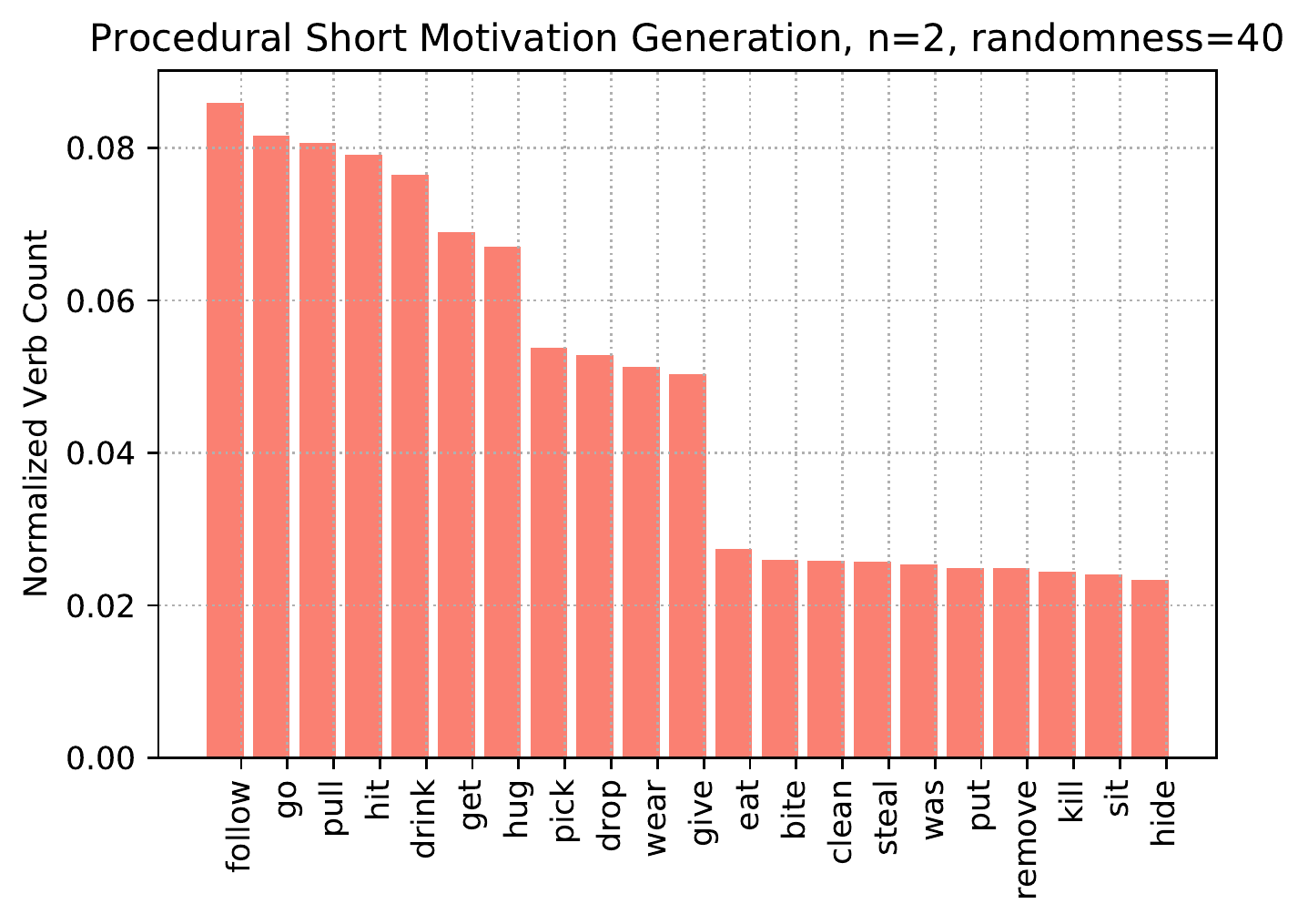}}\\
\subfloat{\includegraphics[width=0.225\linewidth]{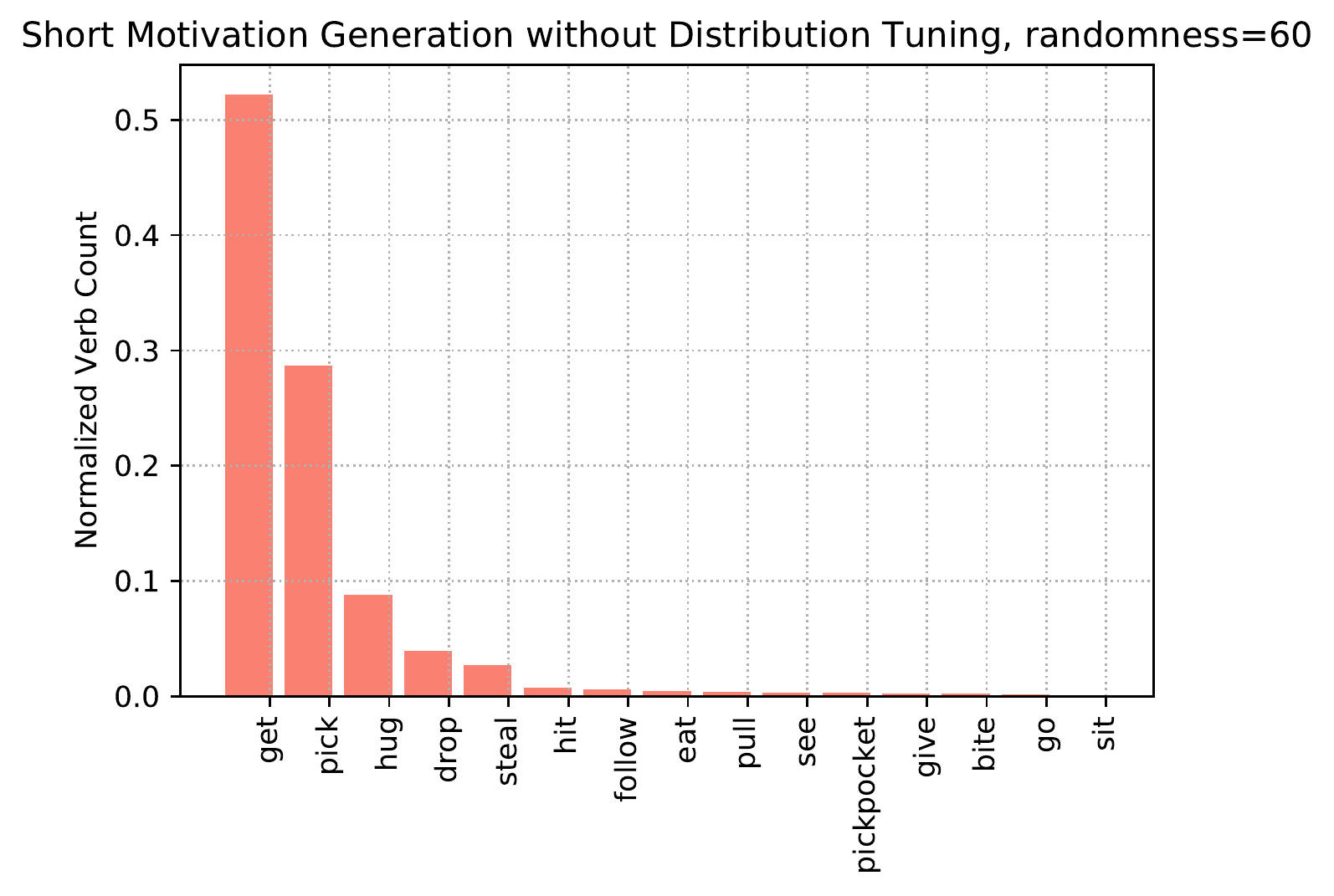}} &
\subfloat{\includegraphics[width=0.225\linewidth]{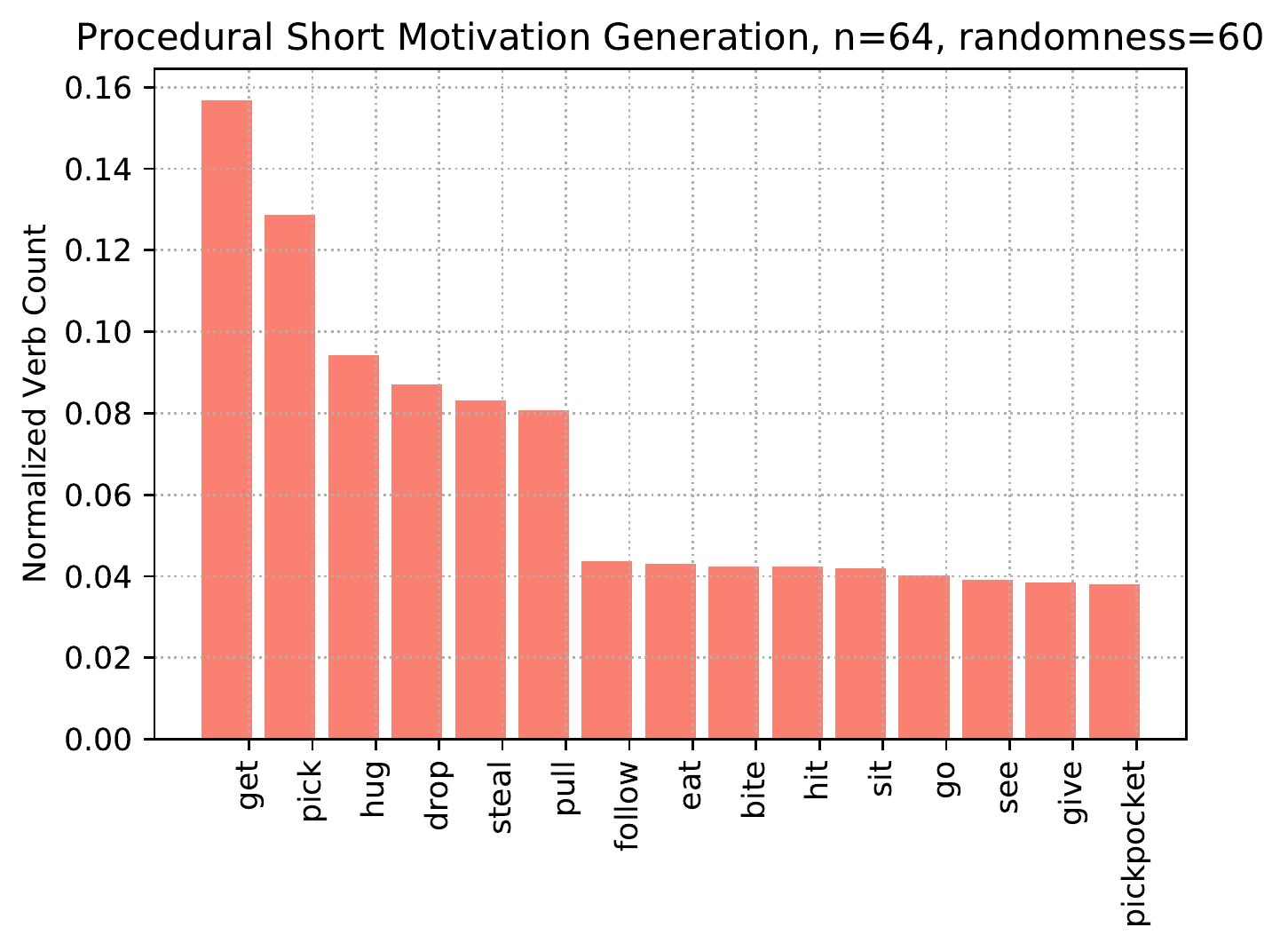}} &
\subfloat{\includegraphics[width=0.225\linewidth]{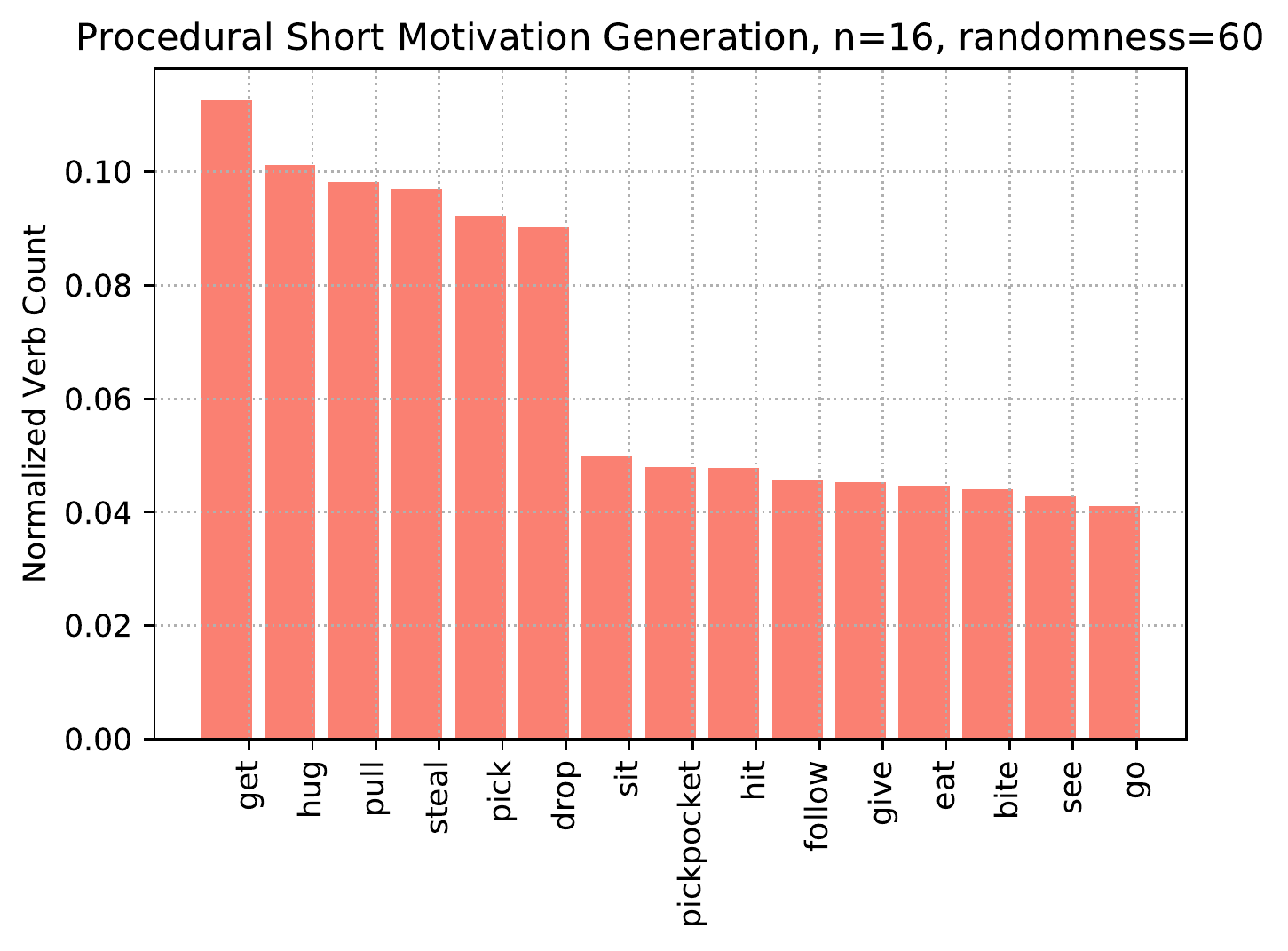}} &
\subfloat{\includegraphics[width=0.225\linewidth]{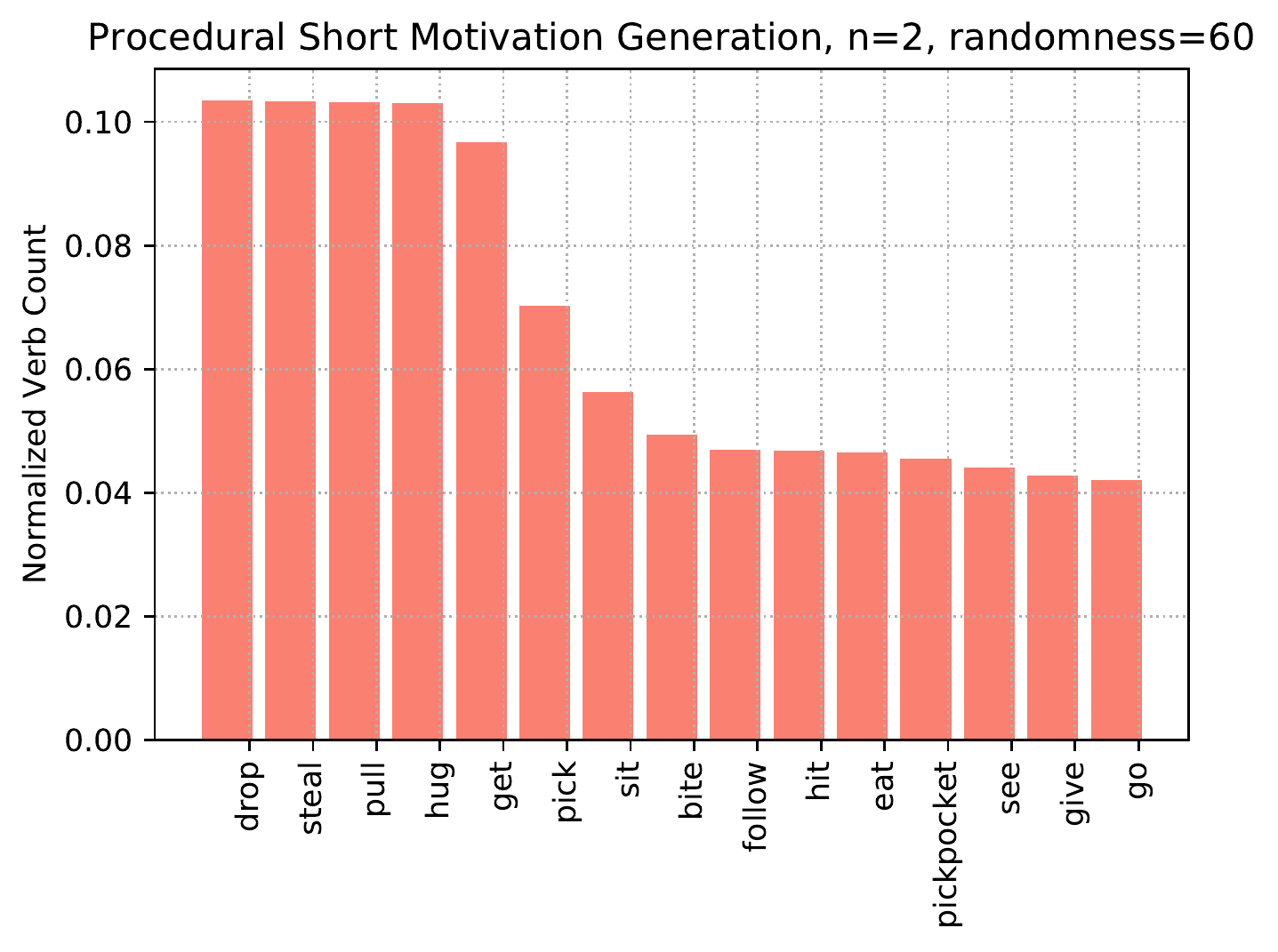}}\\
\subfloat{\includegraphics[width=0.225\linewidth]{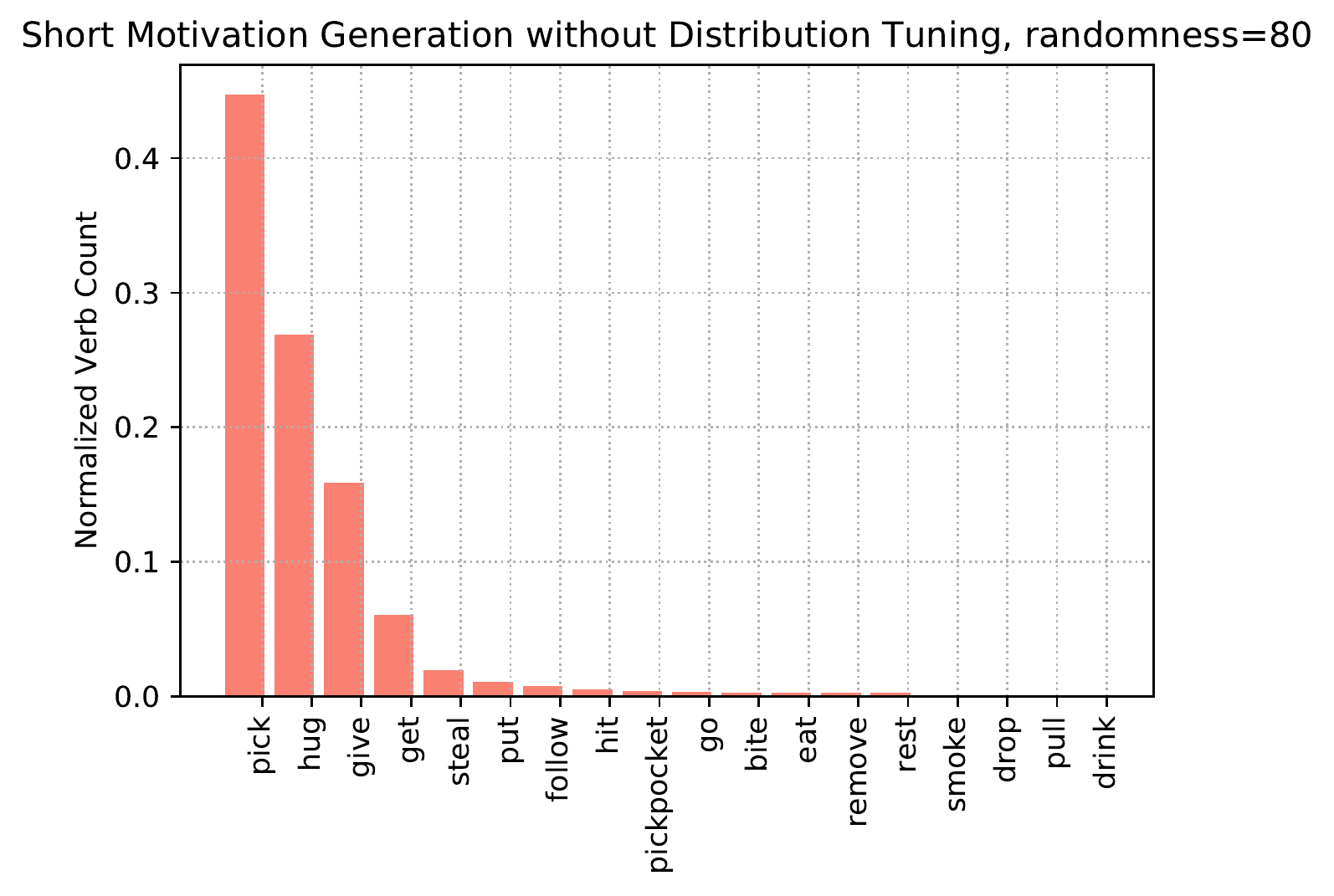}} &
\subfloat{\includegraphics[width=0.225\linewidth]{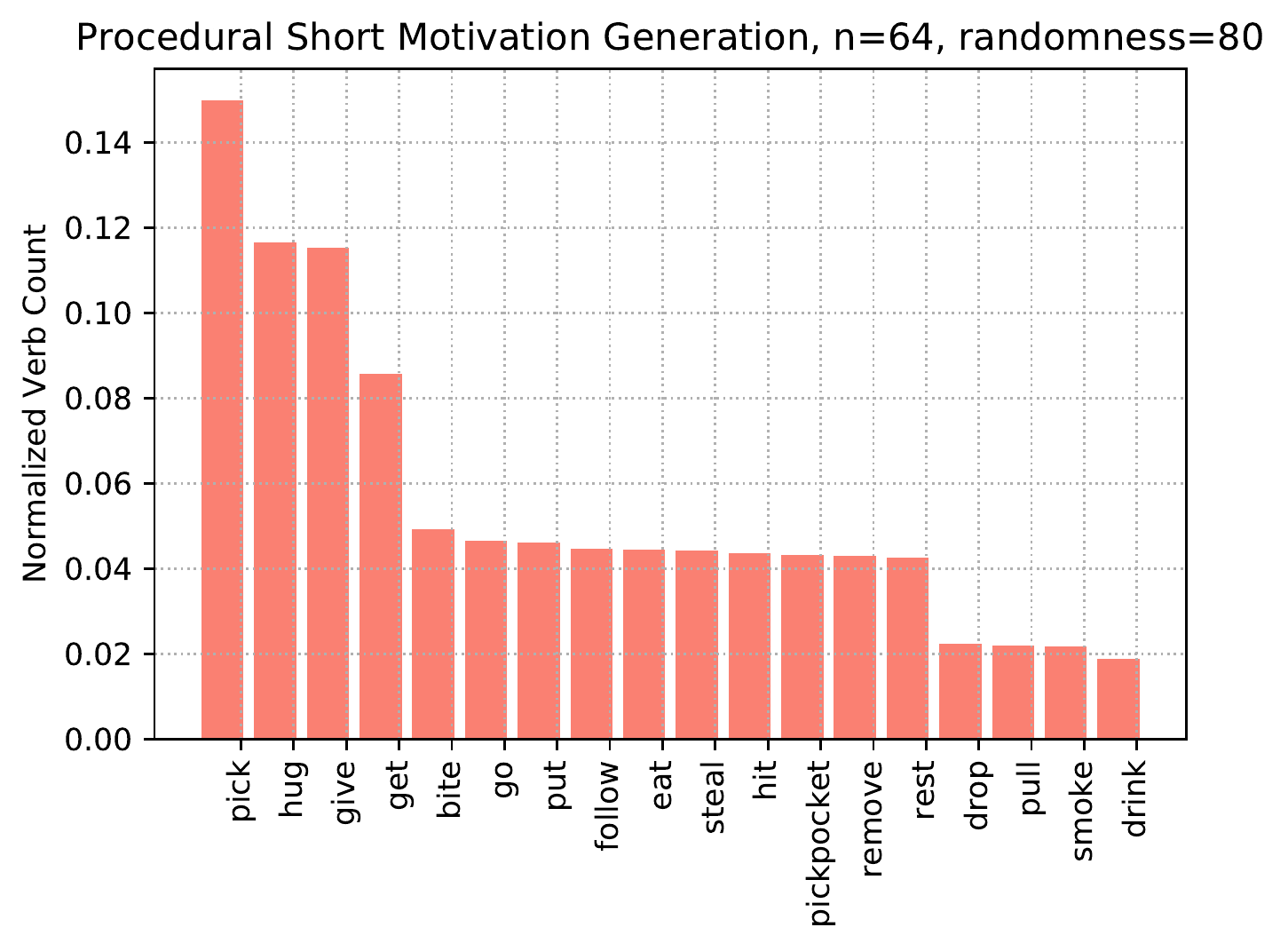}} &
\subfloat{\includegraphics[width=0.225\linewidth]{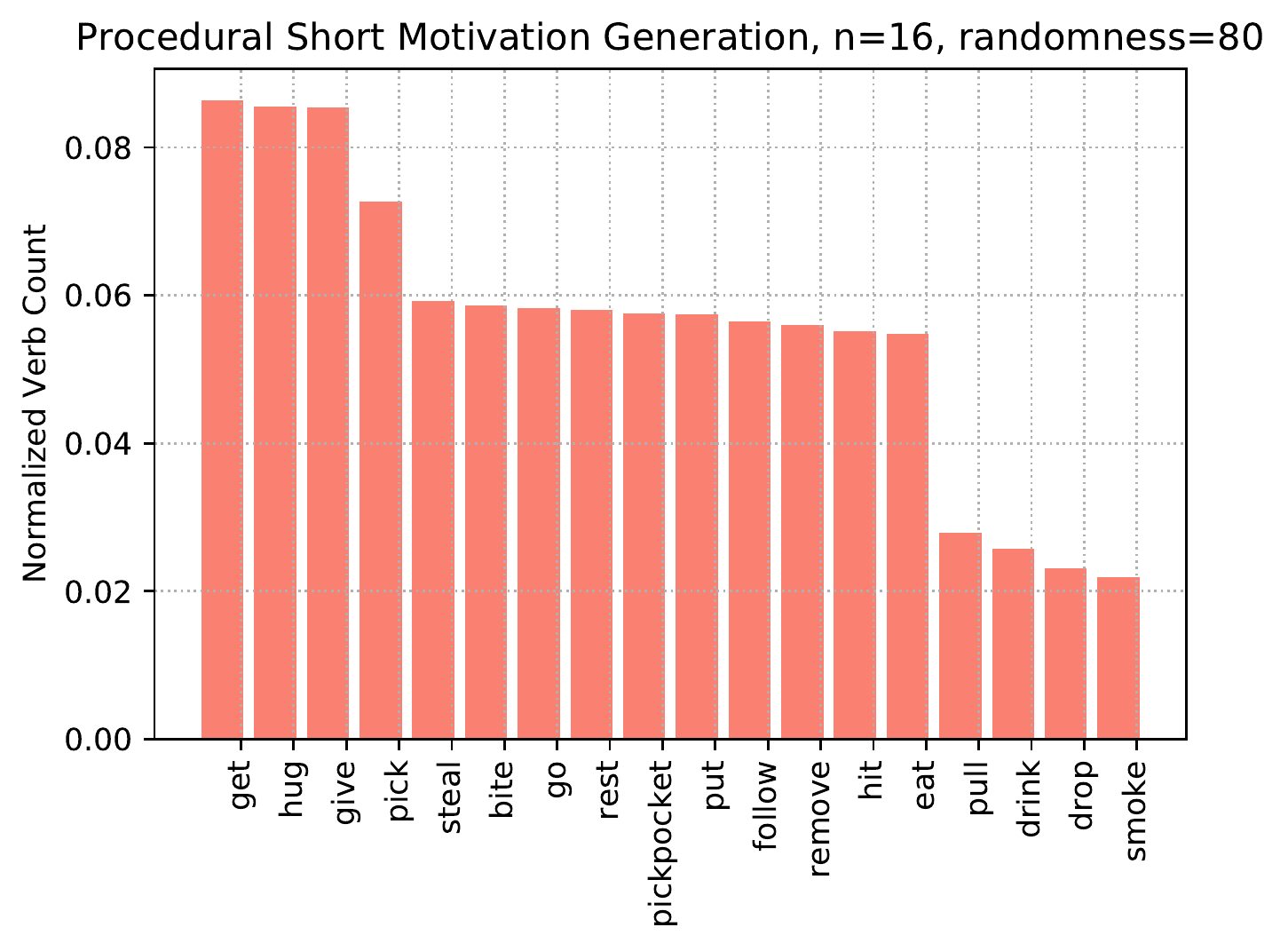}} &
\subfloat{\includegraphics[width=0.225\linewidth]{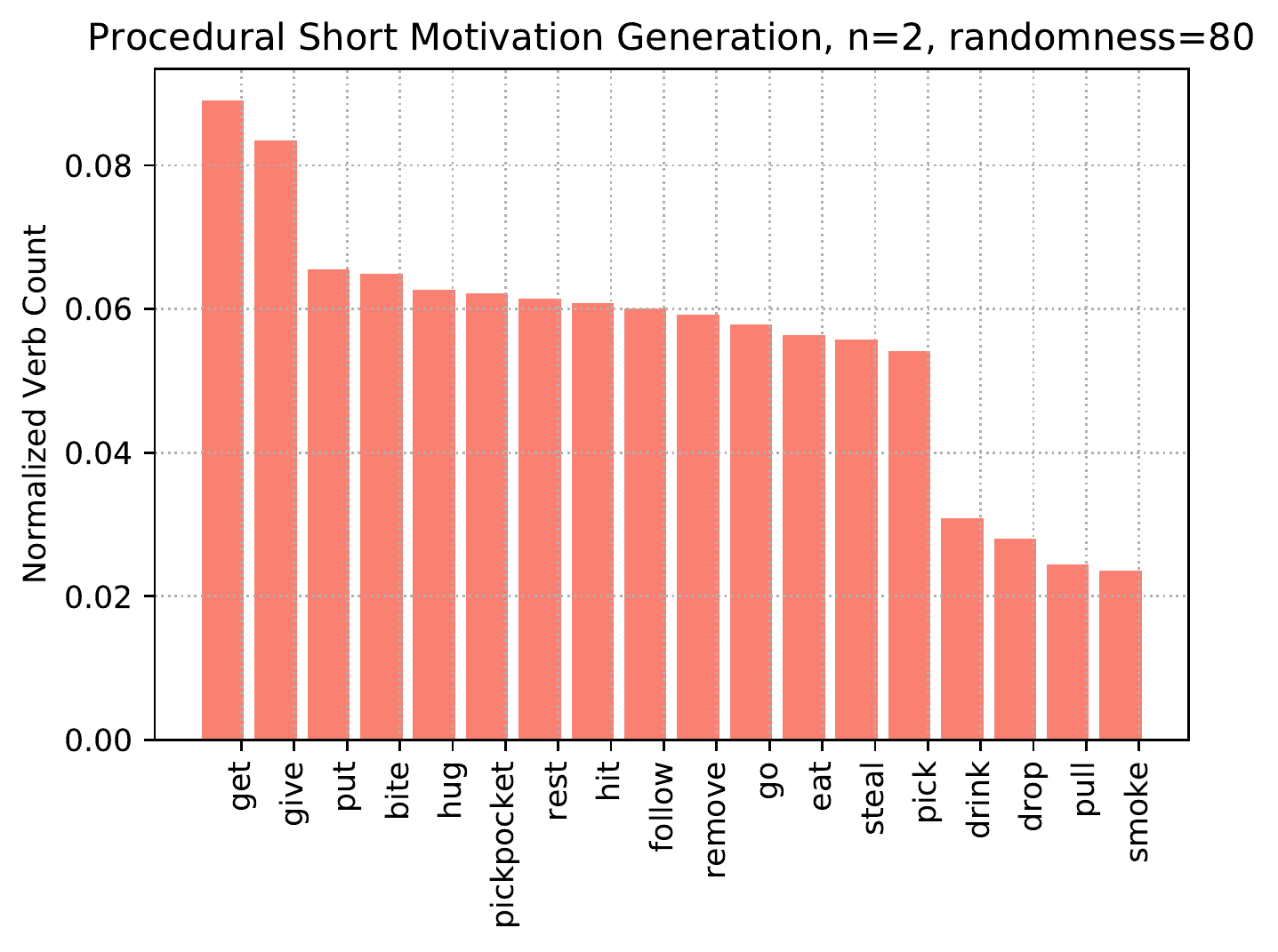}}\\
\subfloat{\includegraphics[width=0.225\linewidth]{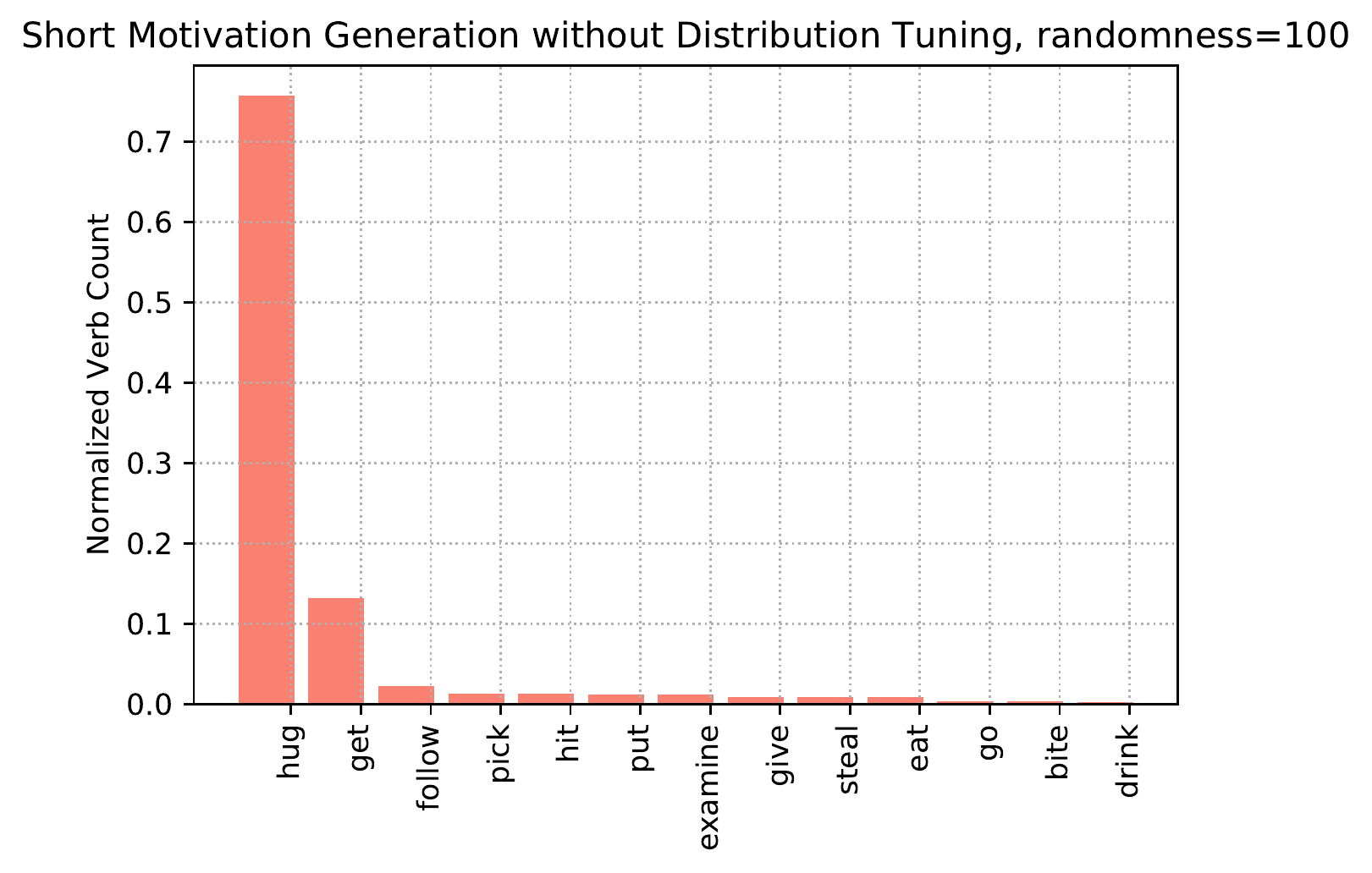}} &
\subfloat{\includegraphics[width=0.225\linewidth]{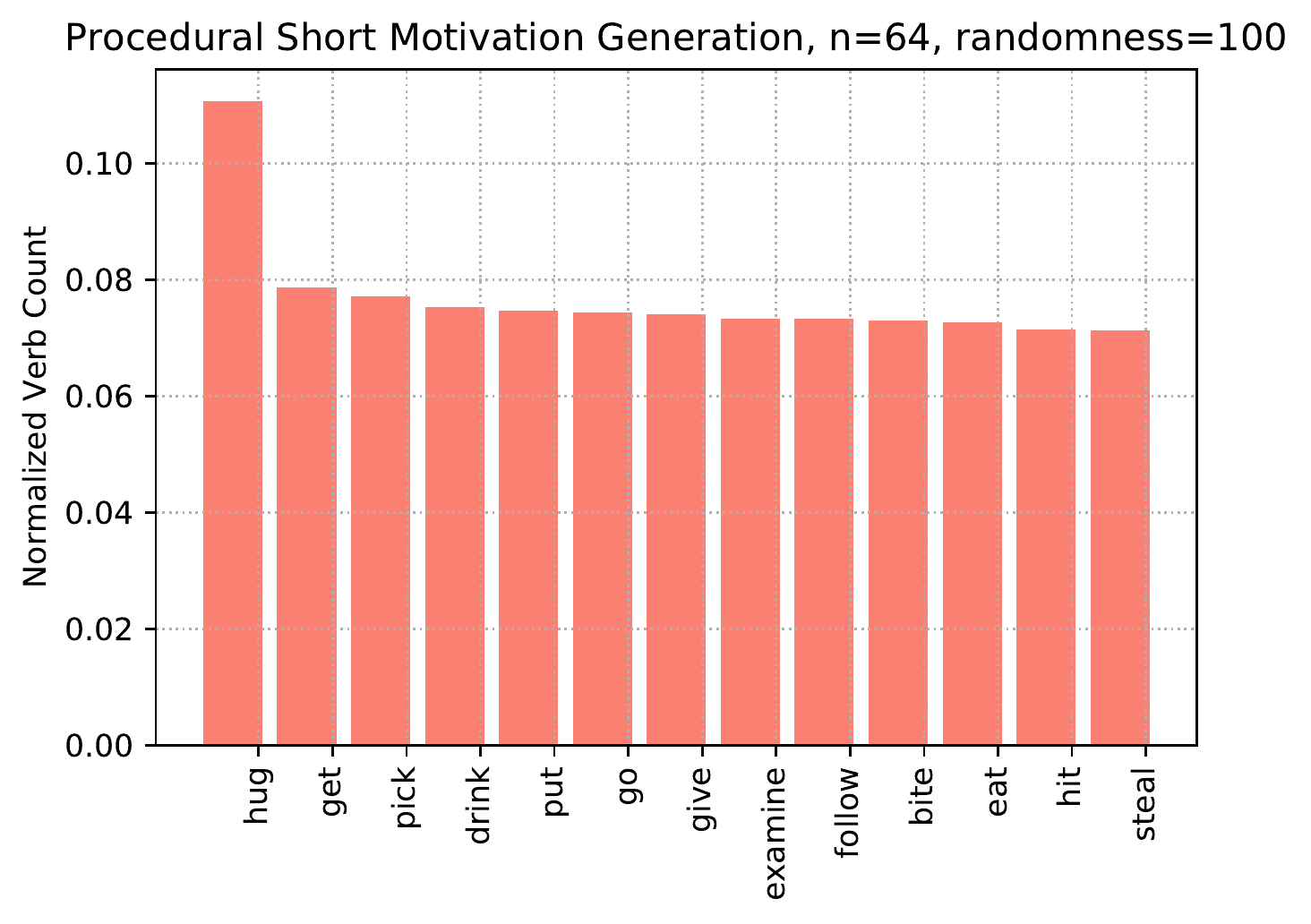}} &
\subfloat{\includegraphics[width=0.225\linewidth]{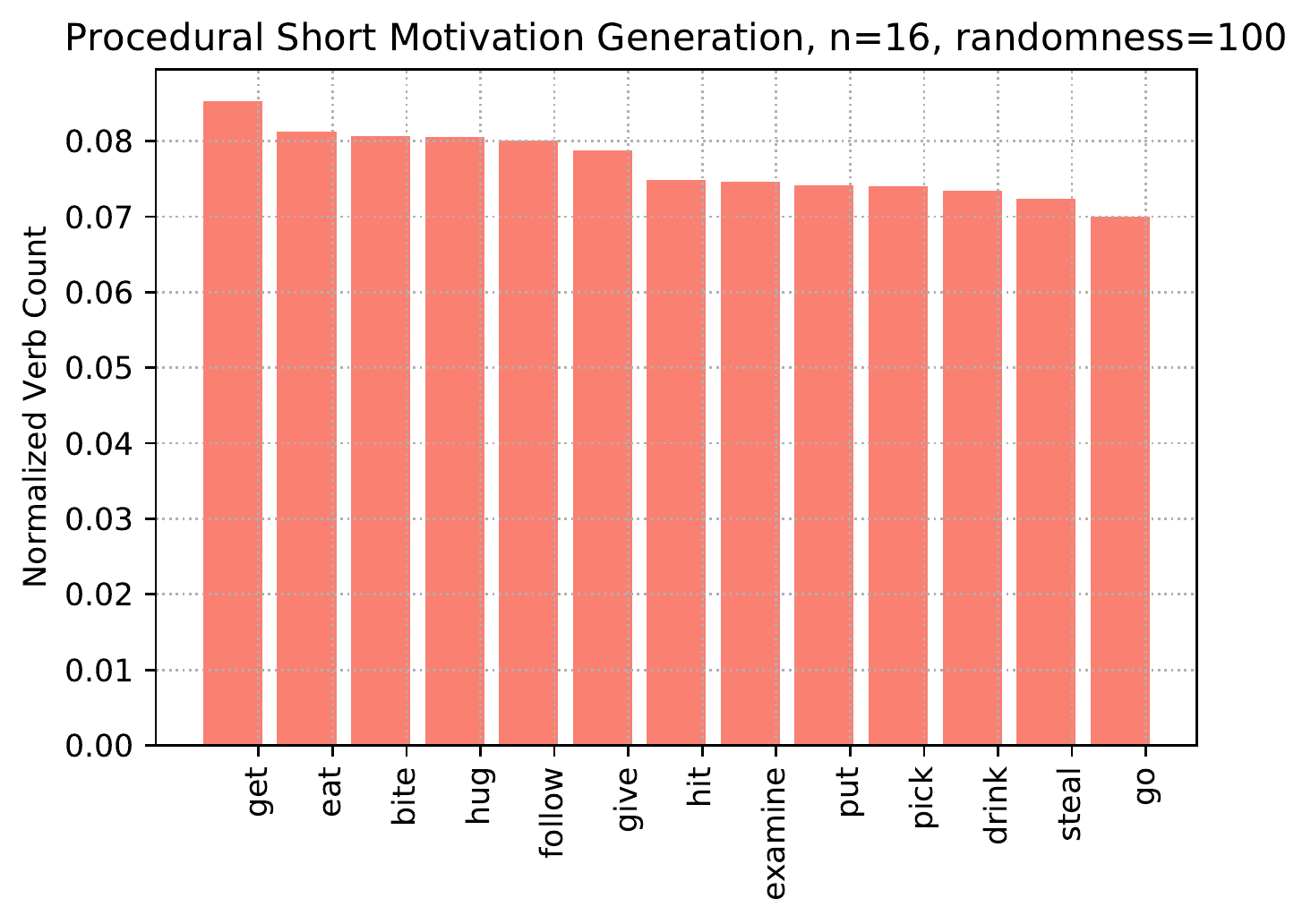}} &
\subfloat{\includegraphics[width=0.225\linewidth]{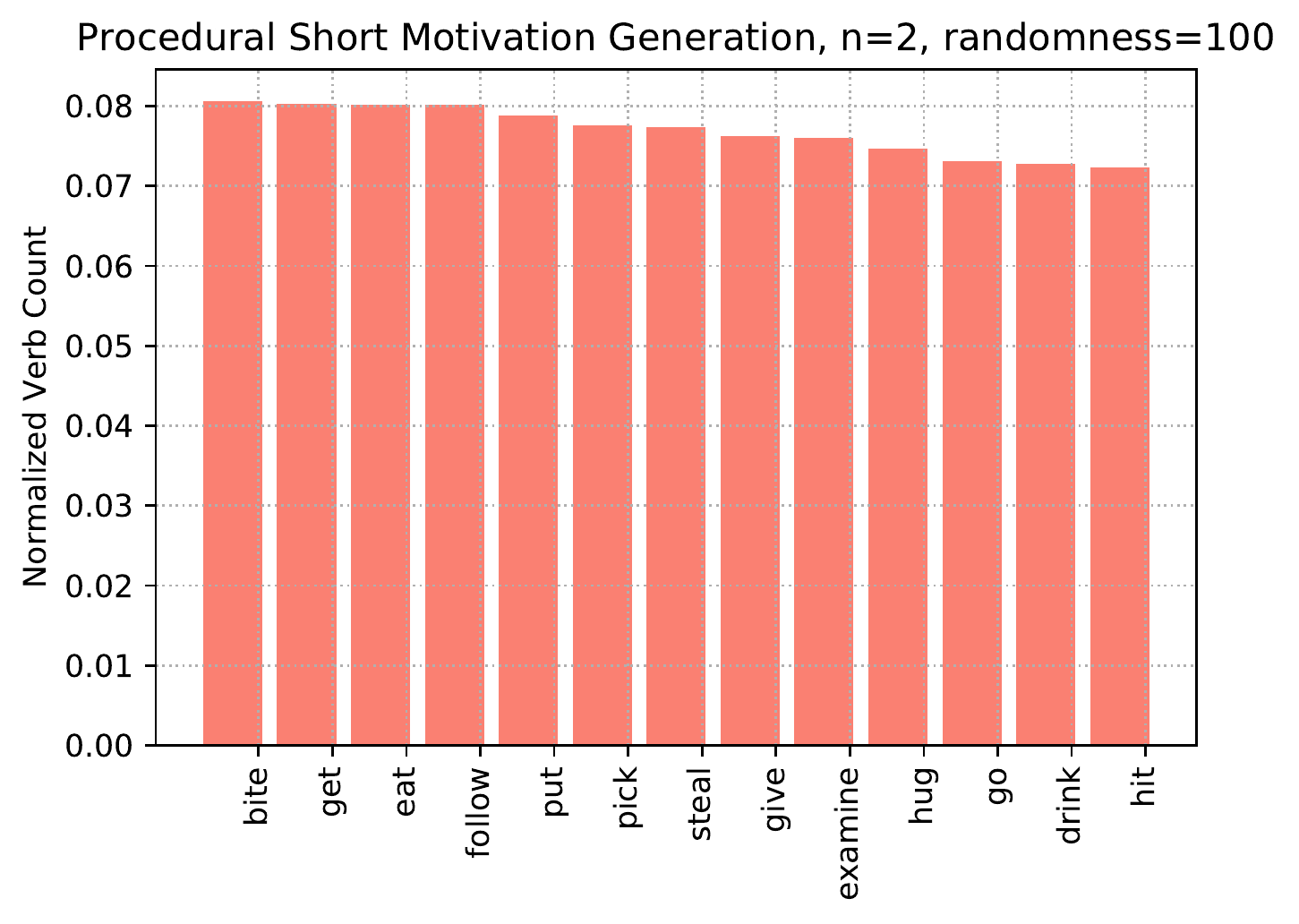}}
\end{tabular}
\caption{Distribution of verbs in the short motivation of the curriculum of quests starting from the original distribution on the left to the flattened and {\bf randomly generated} curriculum on the right as a function of $n$ (Section~\ref{sec:curriculumgen}) with the {\bf randomness percentage} tuning. The y-axis of the different verbs reflect their relative proportion in the pool of quests.}
\end{figure*}

\newpage
\clearpage


\begin{figure*}
\begin{tabular}{cccc}
\subfloat{\includegraphics[width=0.225\linewidth]{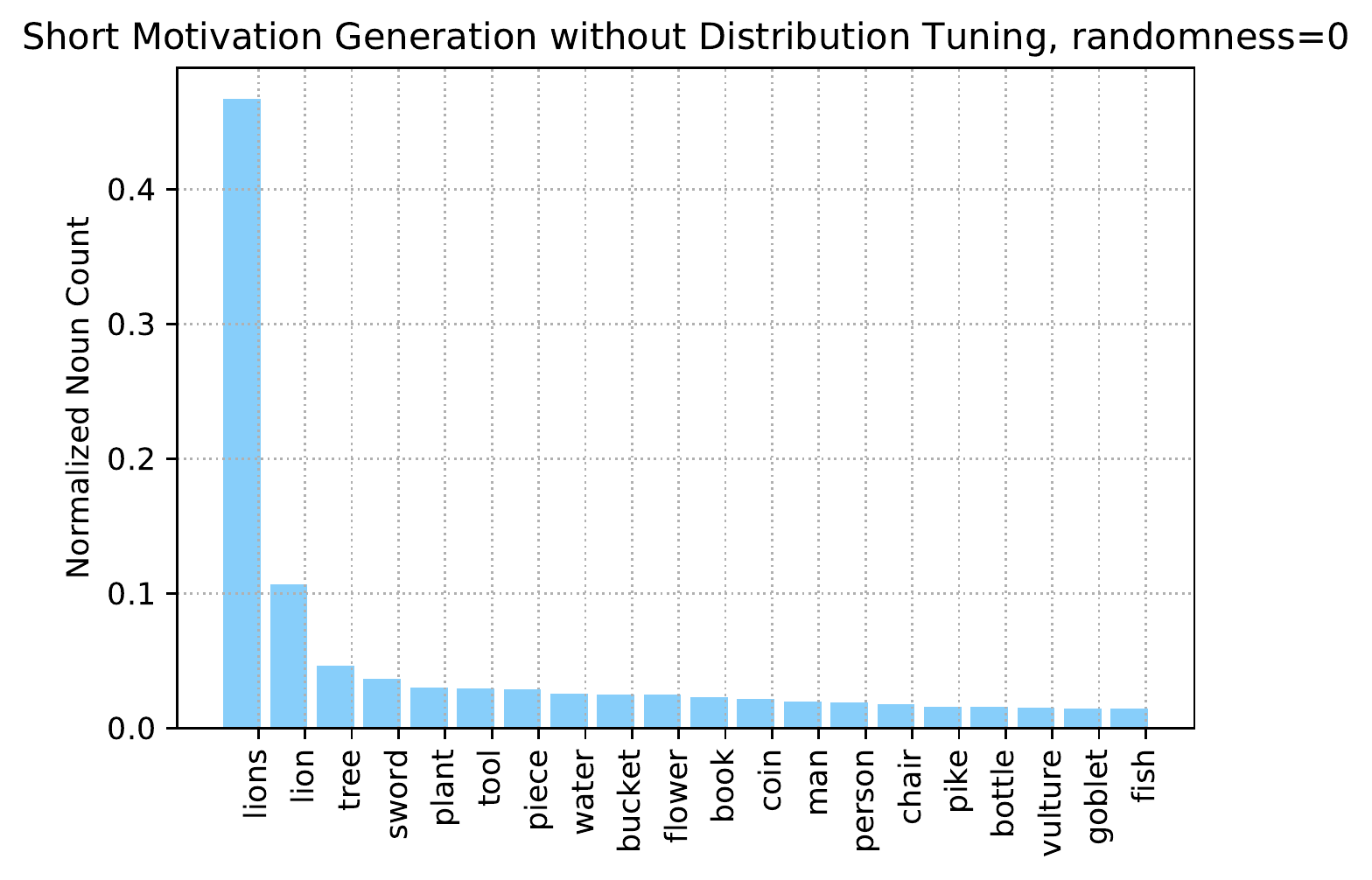}} &
\subfloat{\includegraphics[width=0.225\linewidth]{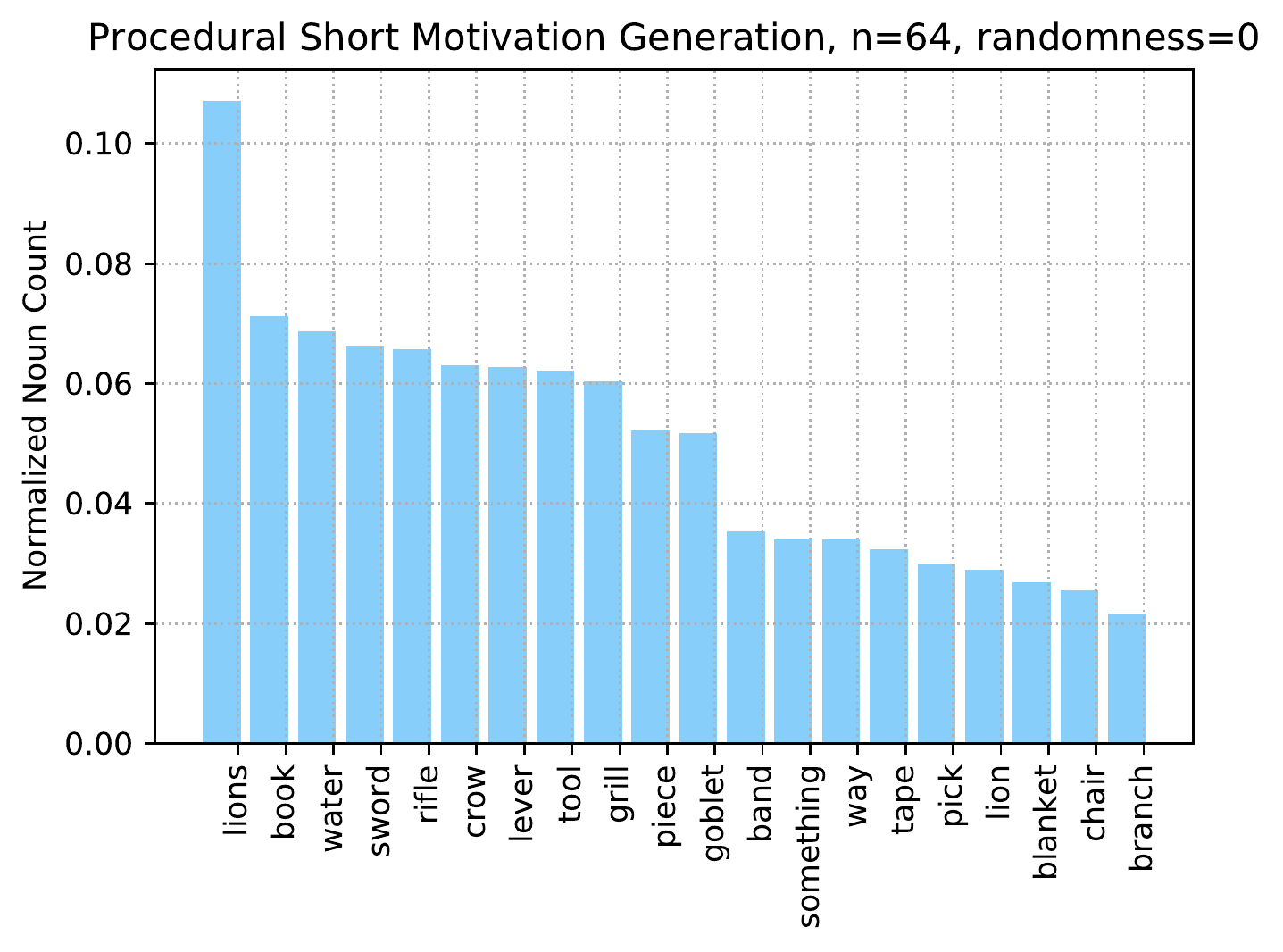}} &
\subfloat{\includegraphics[width=0.225\linewidth]{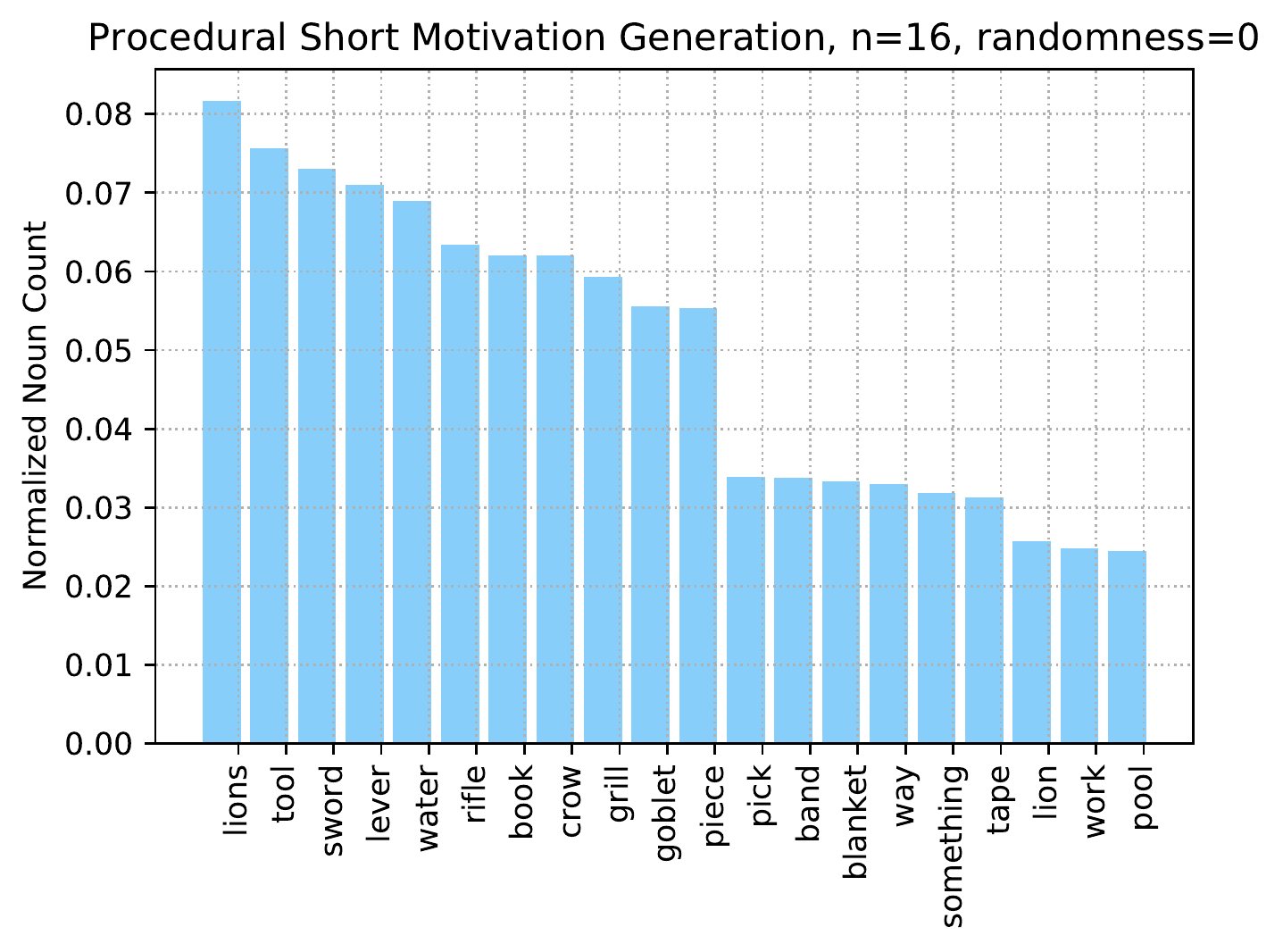}} &
\subfloat{\includegraphics[width=0.225\linewidth]{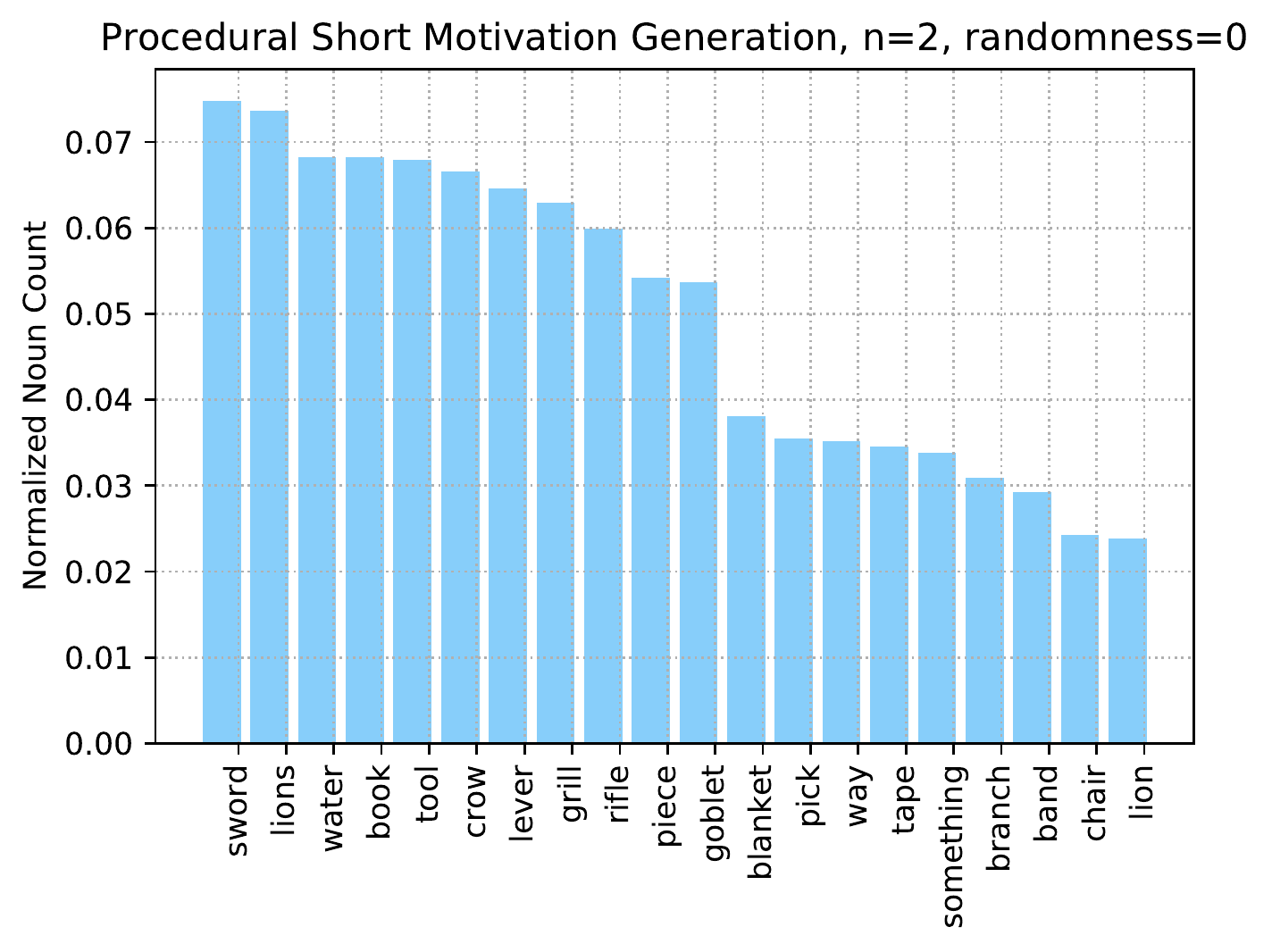}}\\
\subfloat{\includegraphics[width=0.225\linewidth]{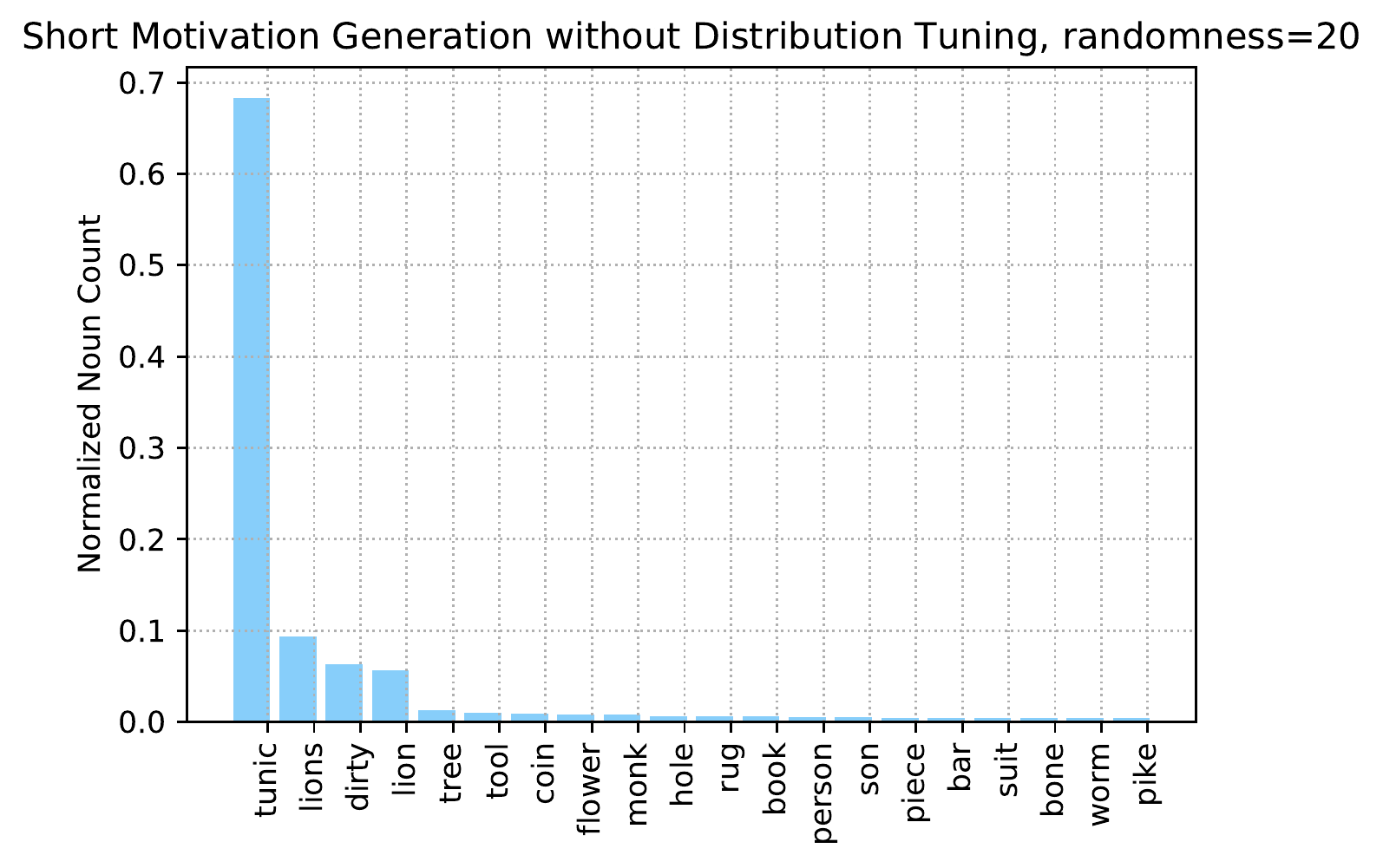}} &
\subfloat{\includegraphics[width=0.225\linewidth]{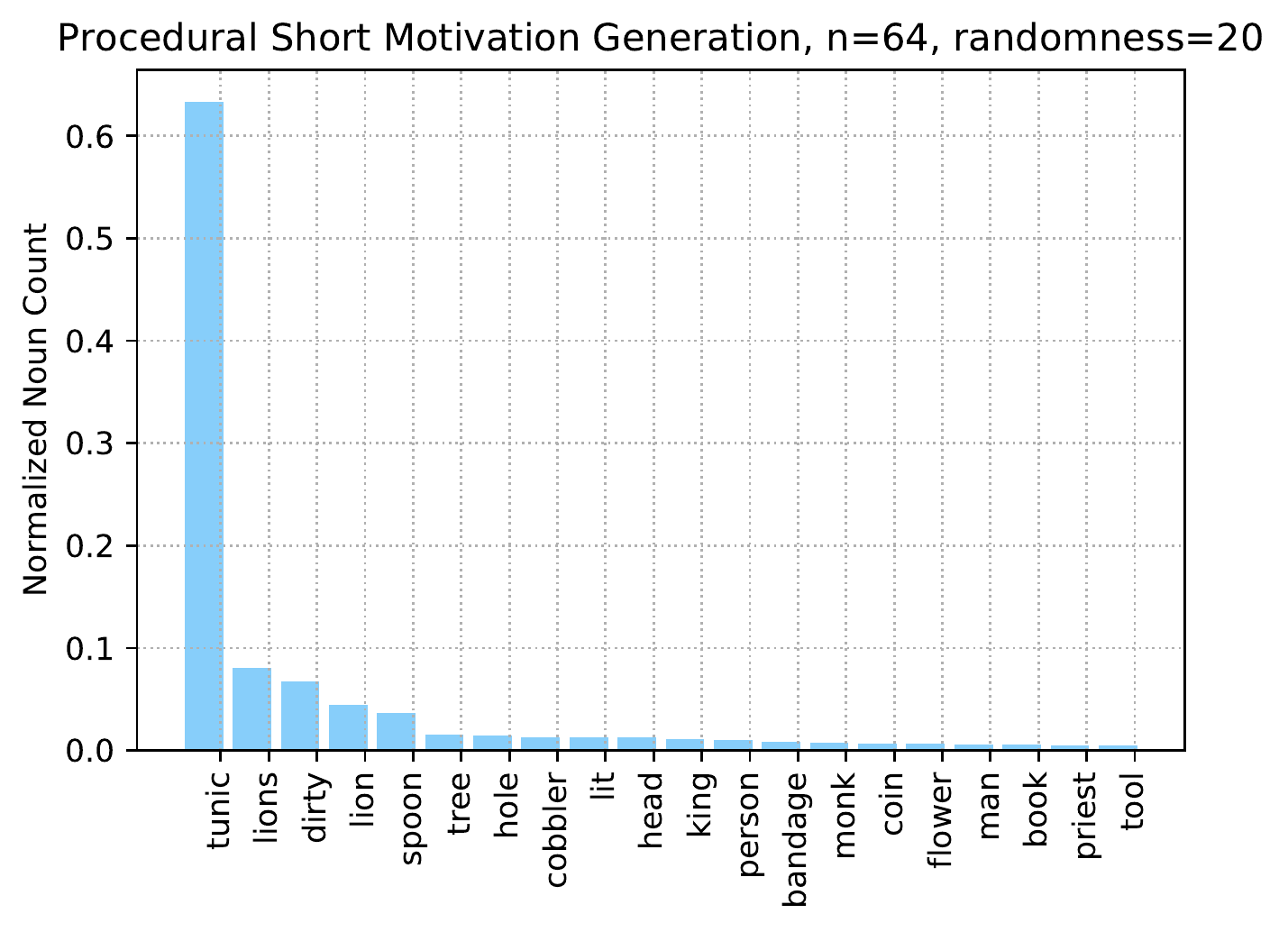}} &
\subfloat{\includegraphics[width=0.225\linewidth]{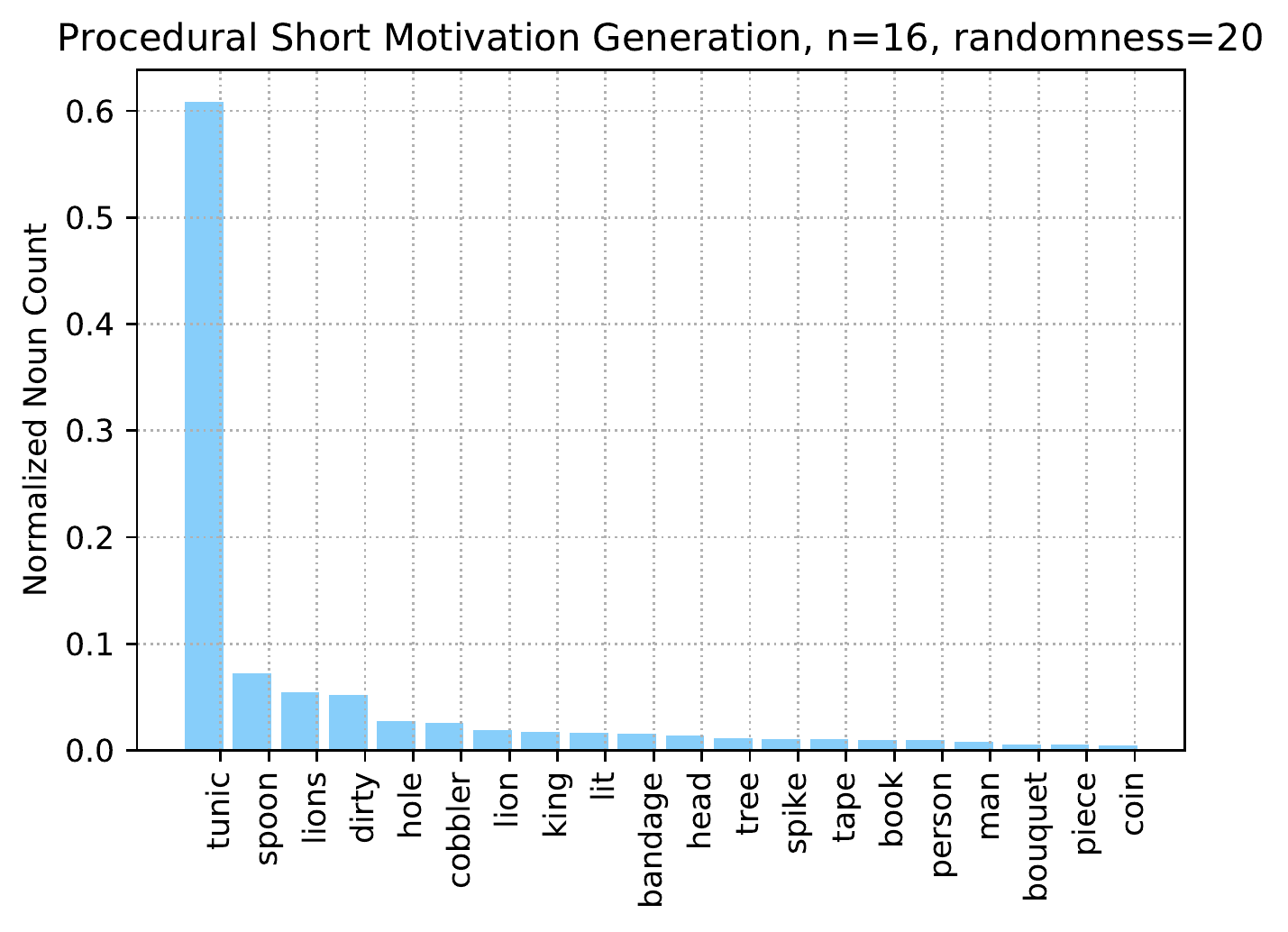}} &
\subfloat{\includegraphics[width=0.225\linewidth]{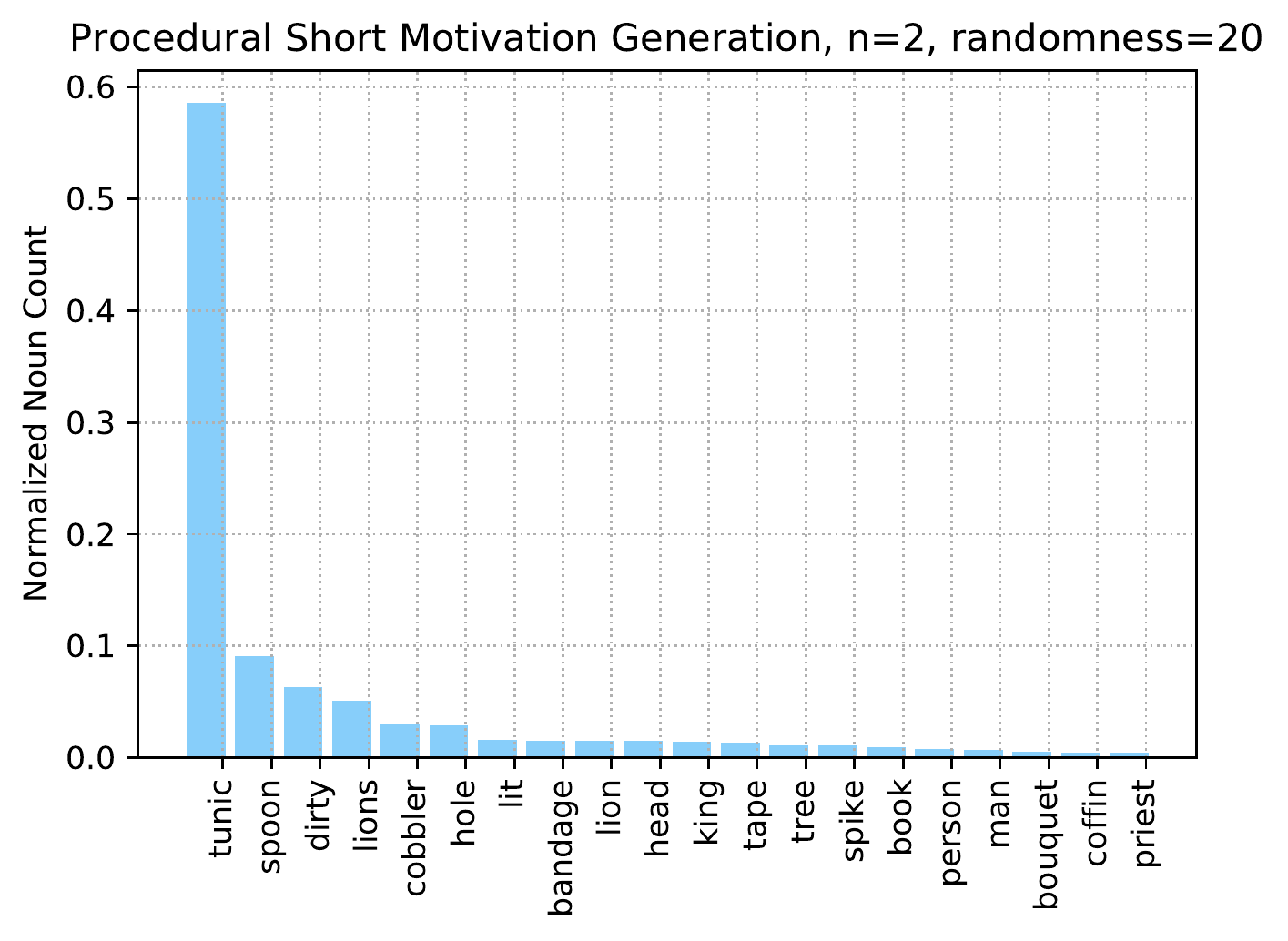}}\\
\subfloat{\includegraphics[width=0.225\linewidth]{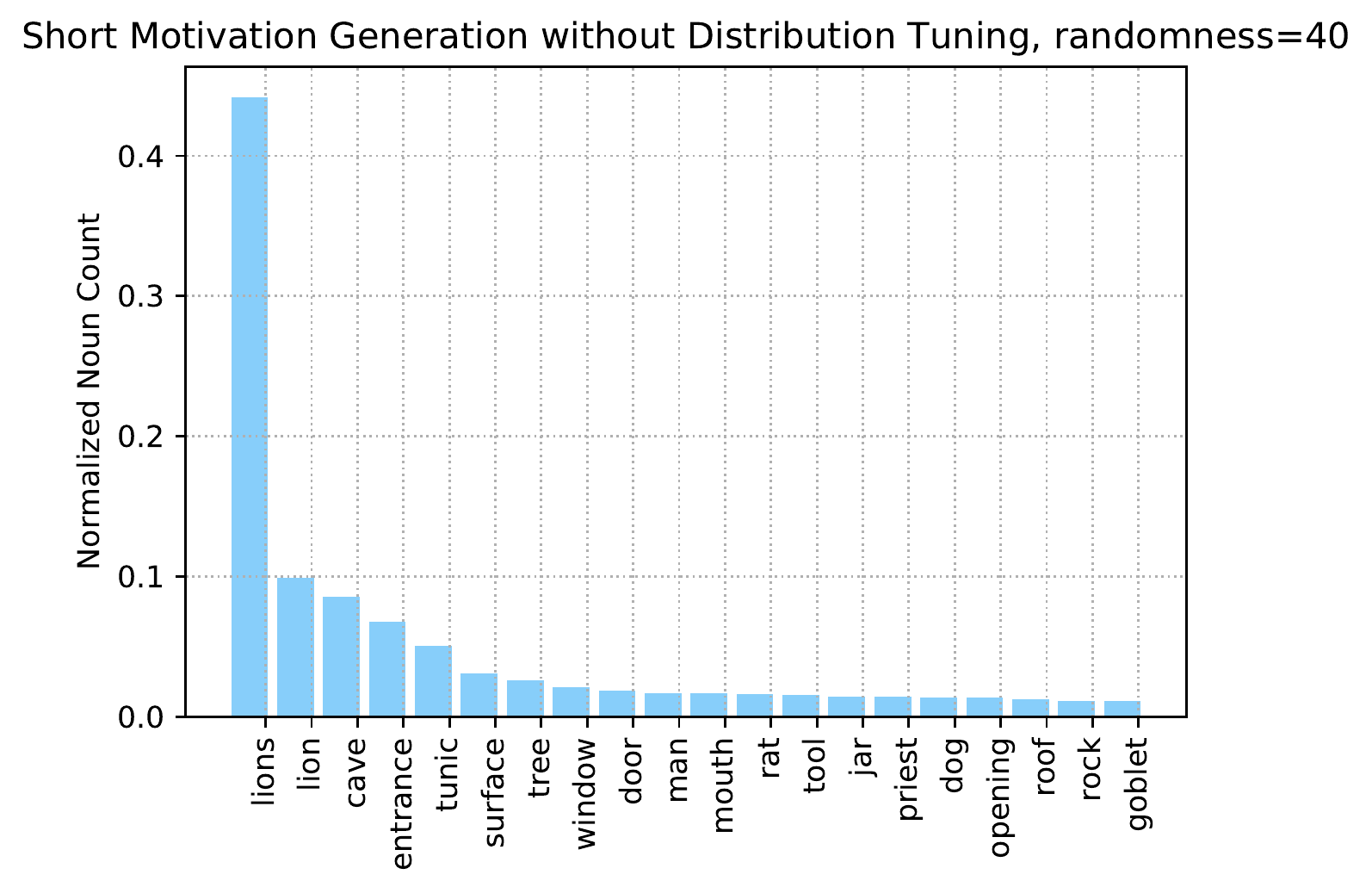}} &
\subfloat{\includegraphics[width=0.225\linewidth]{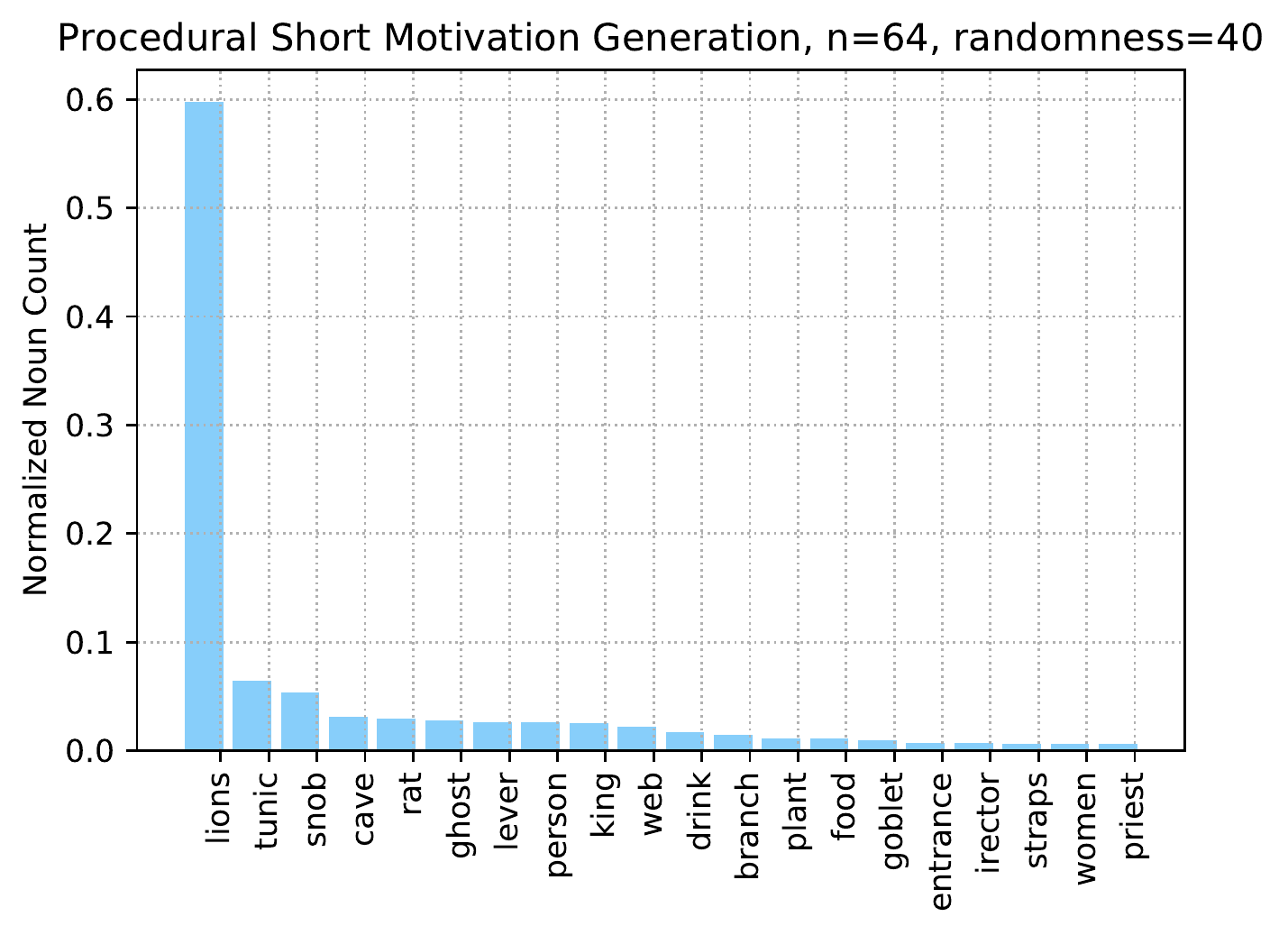}} &
\subfloat{\includegraphics[width=0.225\linewidth]{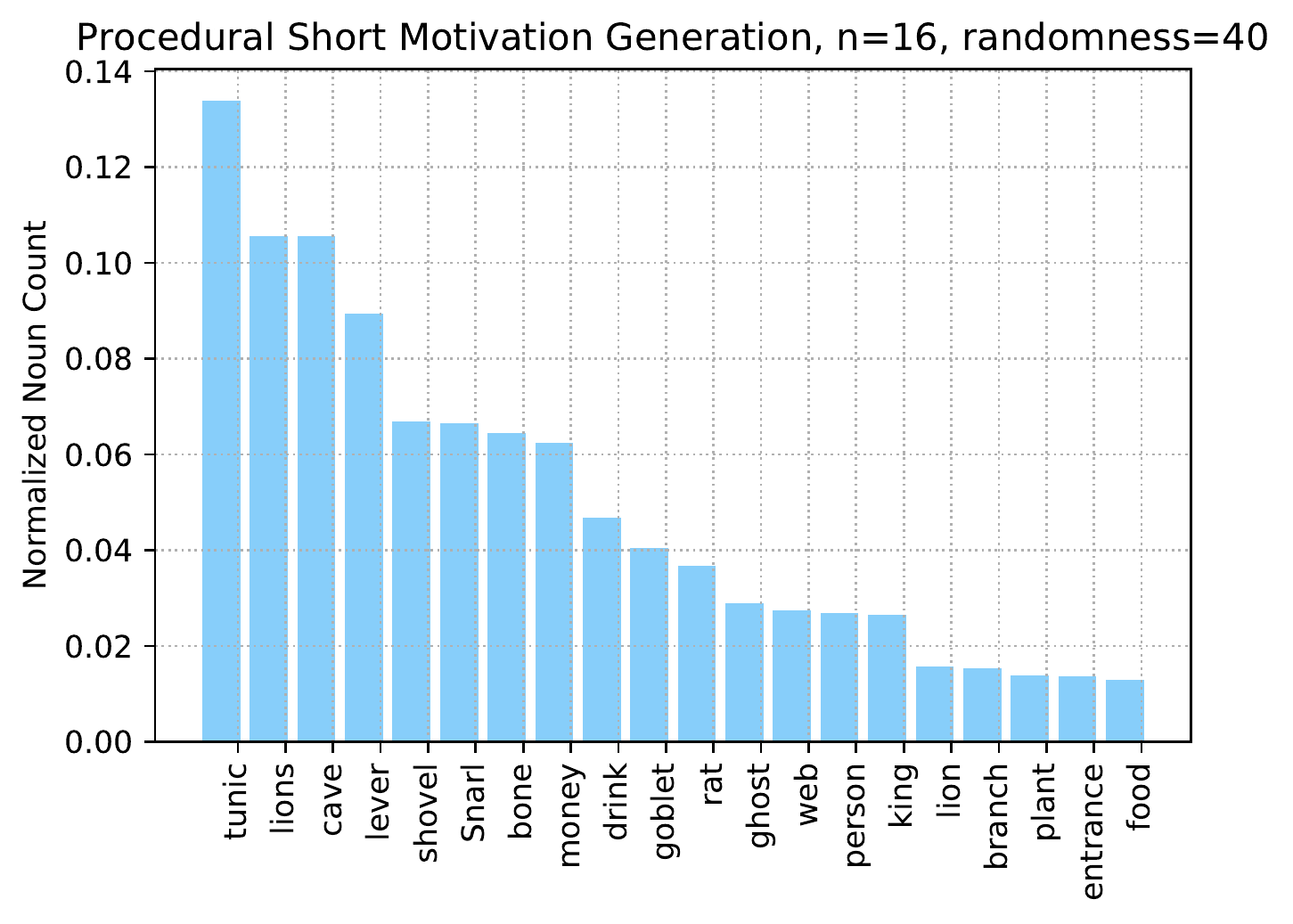}} &
\subfloat{\includegraphics[width=0.225\linewidth]{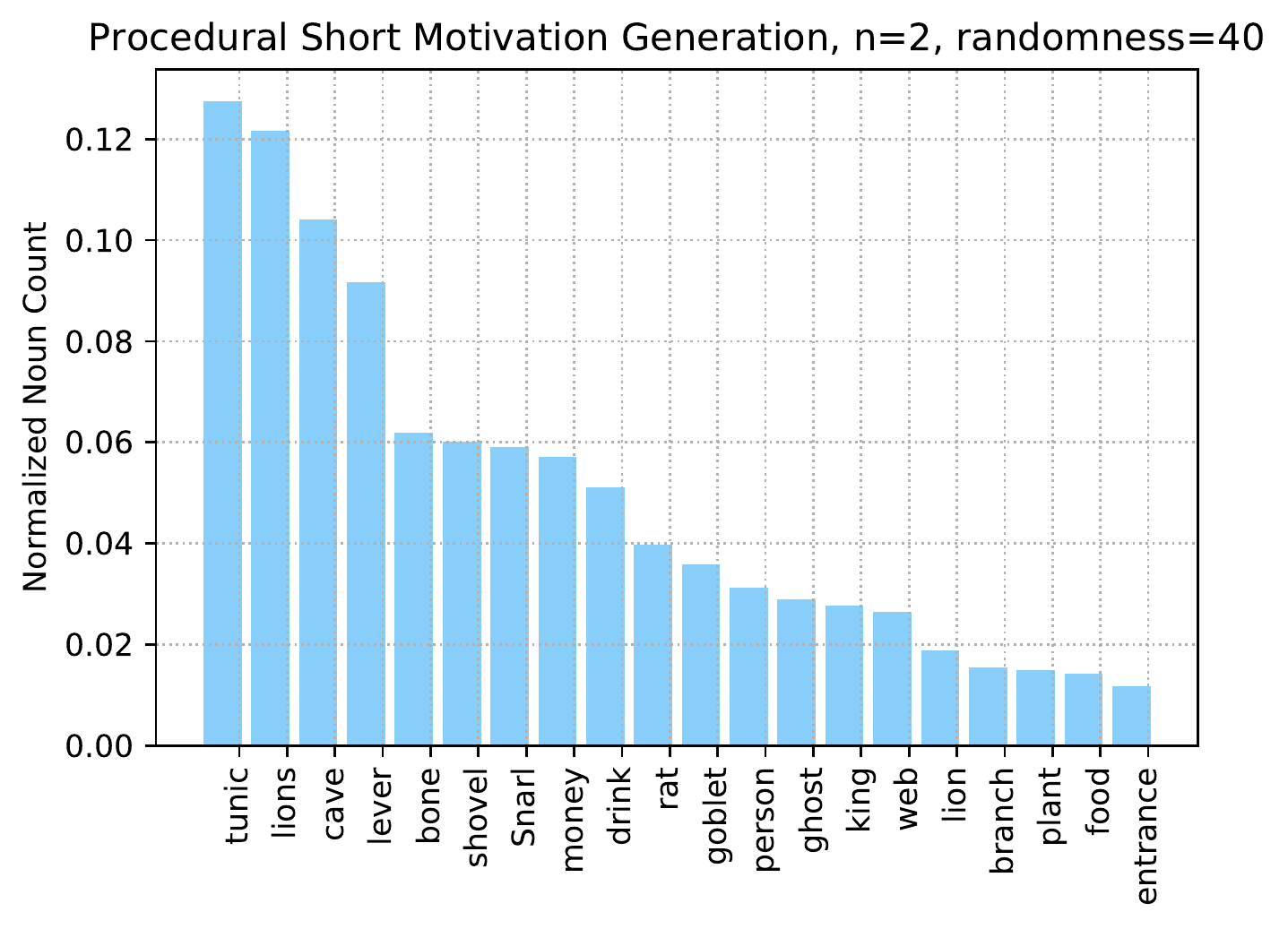}}\\
\subfloat{\includegraphics[width=0.225\linewidth]{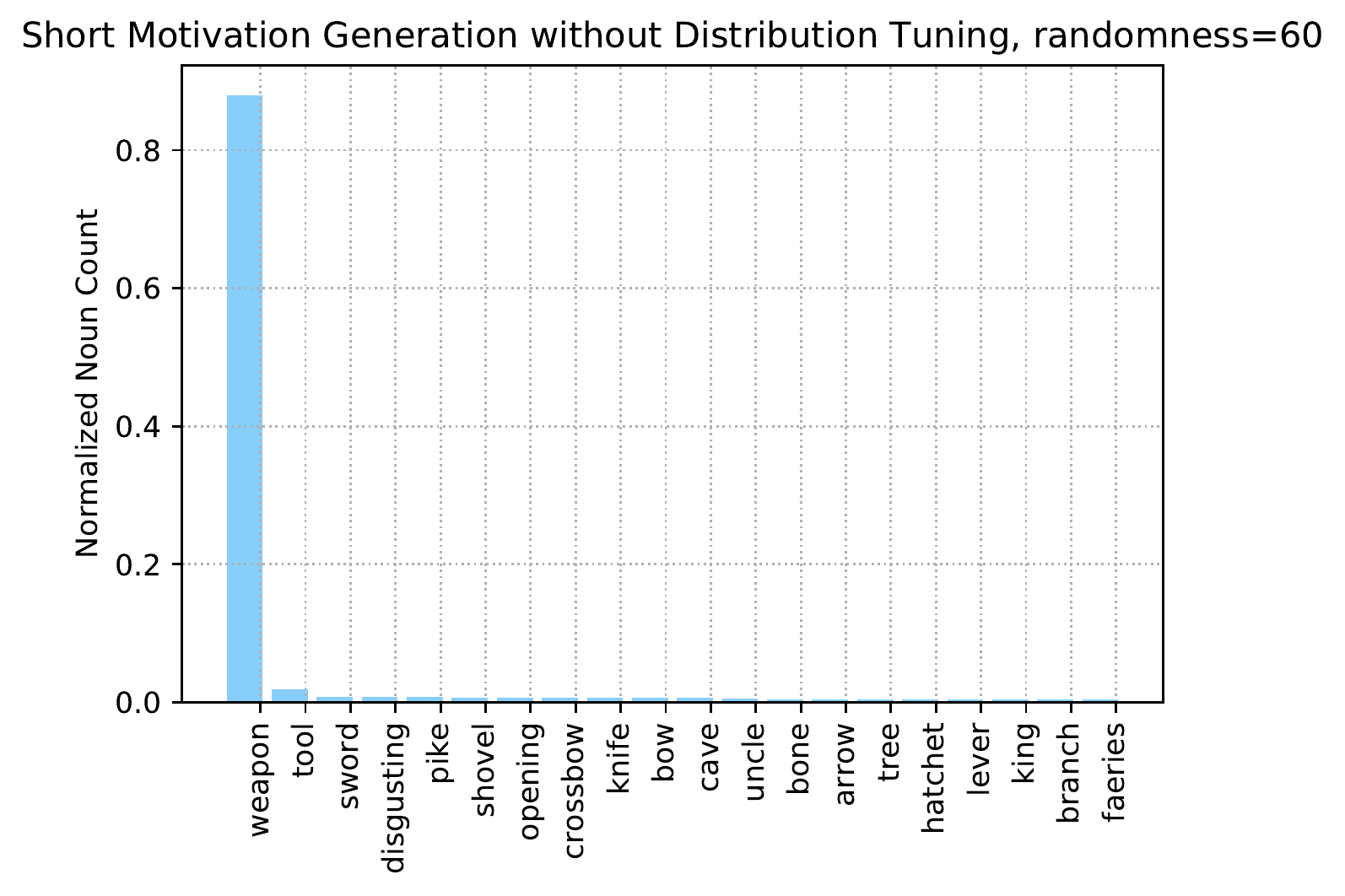}} &
\subfloat{\includegraphics[width=0.225\linewidth]{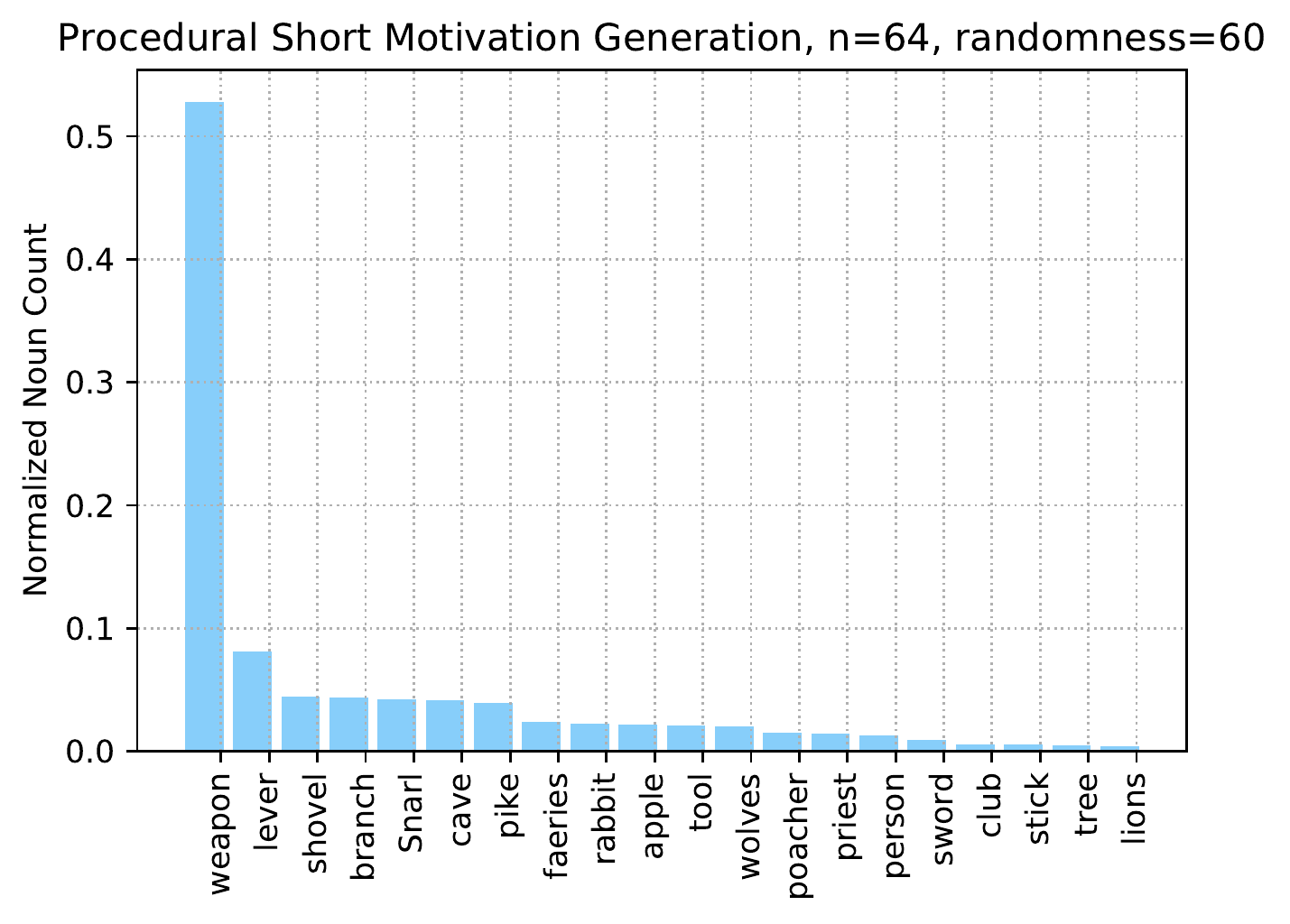}} &
\subfloat{\includegraphics[width=0.225\linewidth]{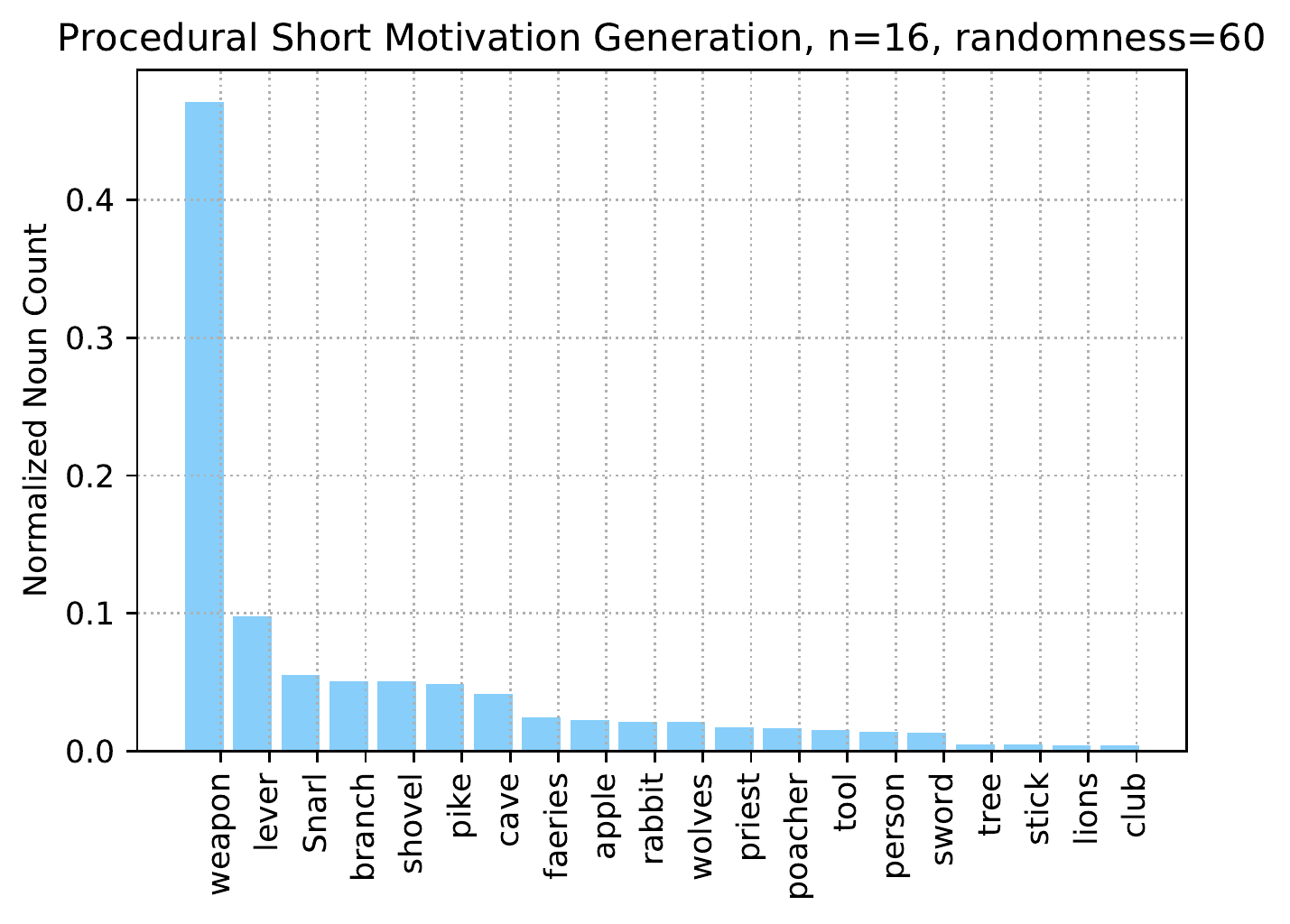}} &
\subfloat{\includegraphics[width=0.225\linewidth]{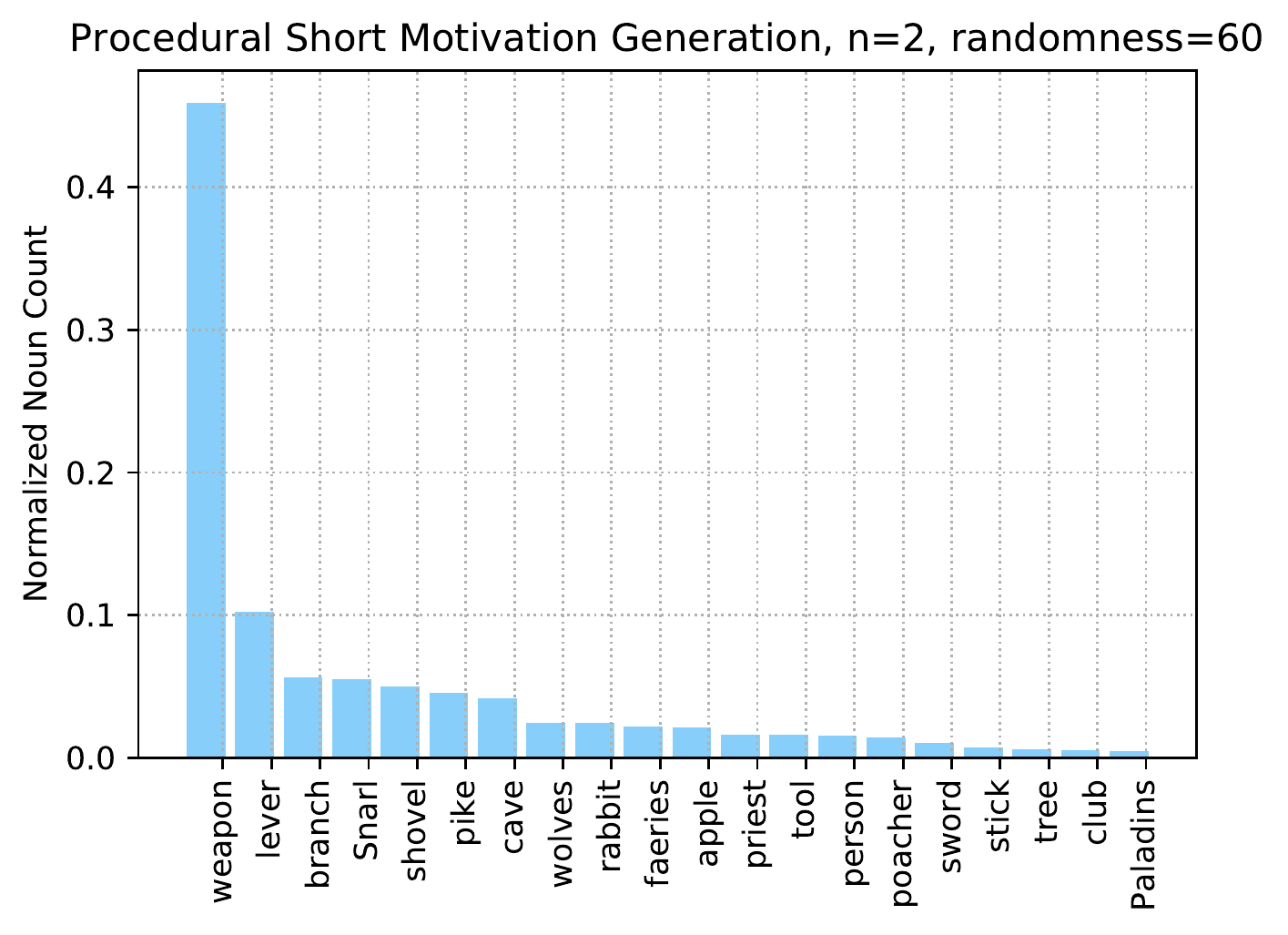}}\\
\subfloat{\includegraphics[width=0.225\linewidth]{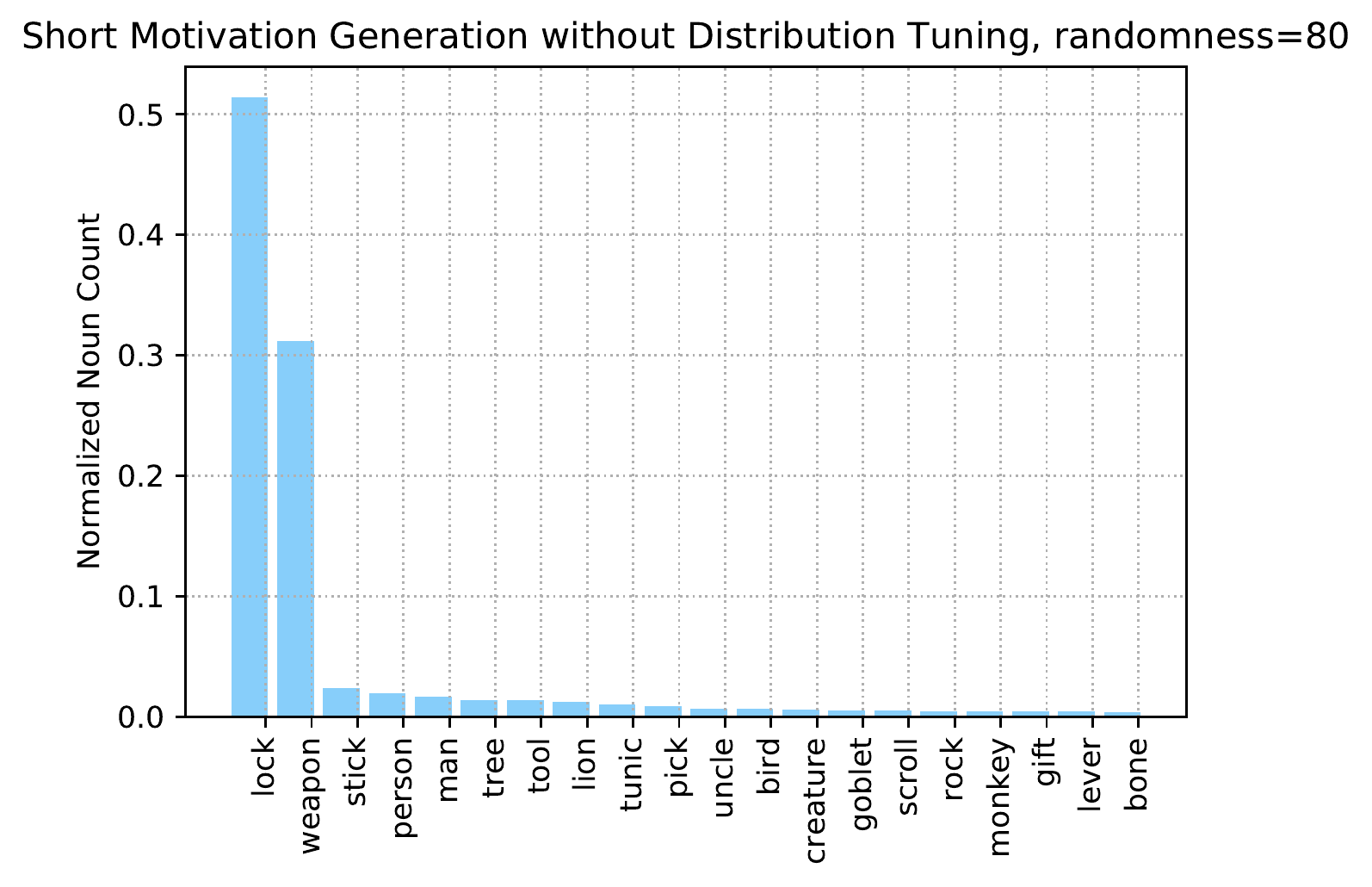}} &
\subfloat{\includegraphics[width=0.225\linewidth]{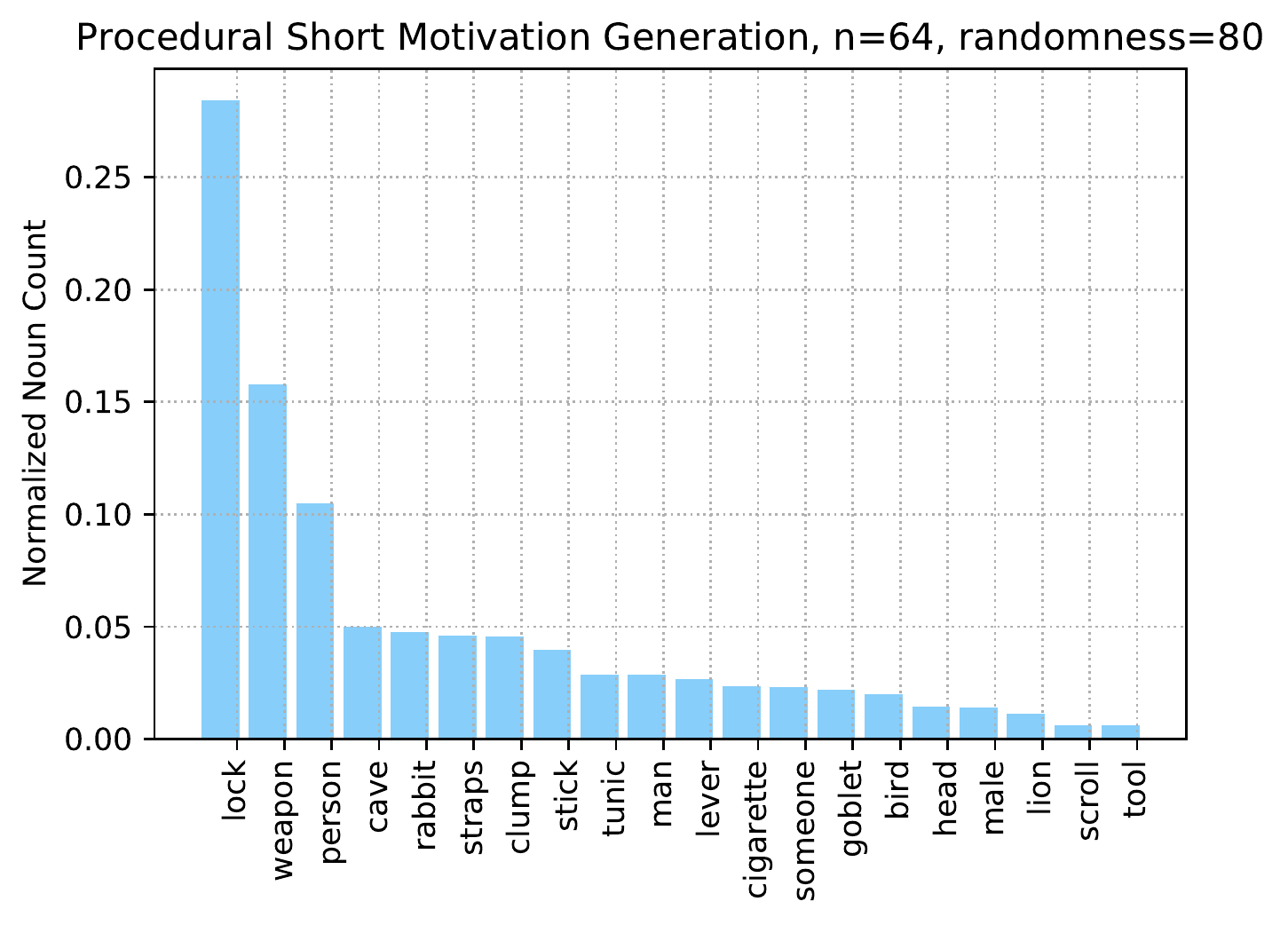}} &
\subfloat{\includegraphics[width=0.225\linewidth]{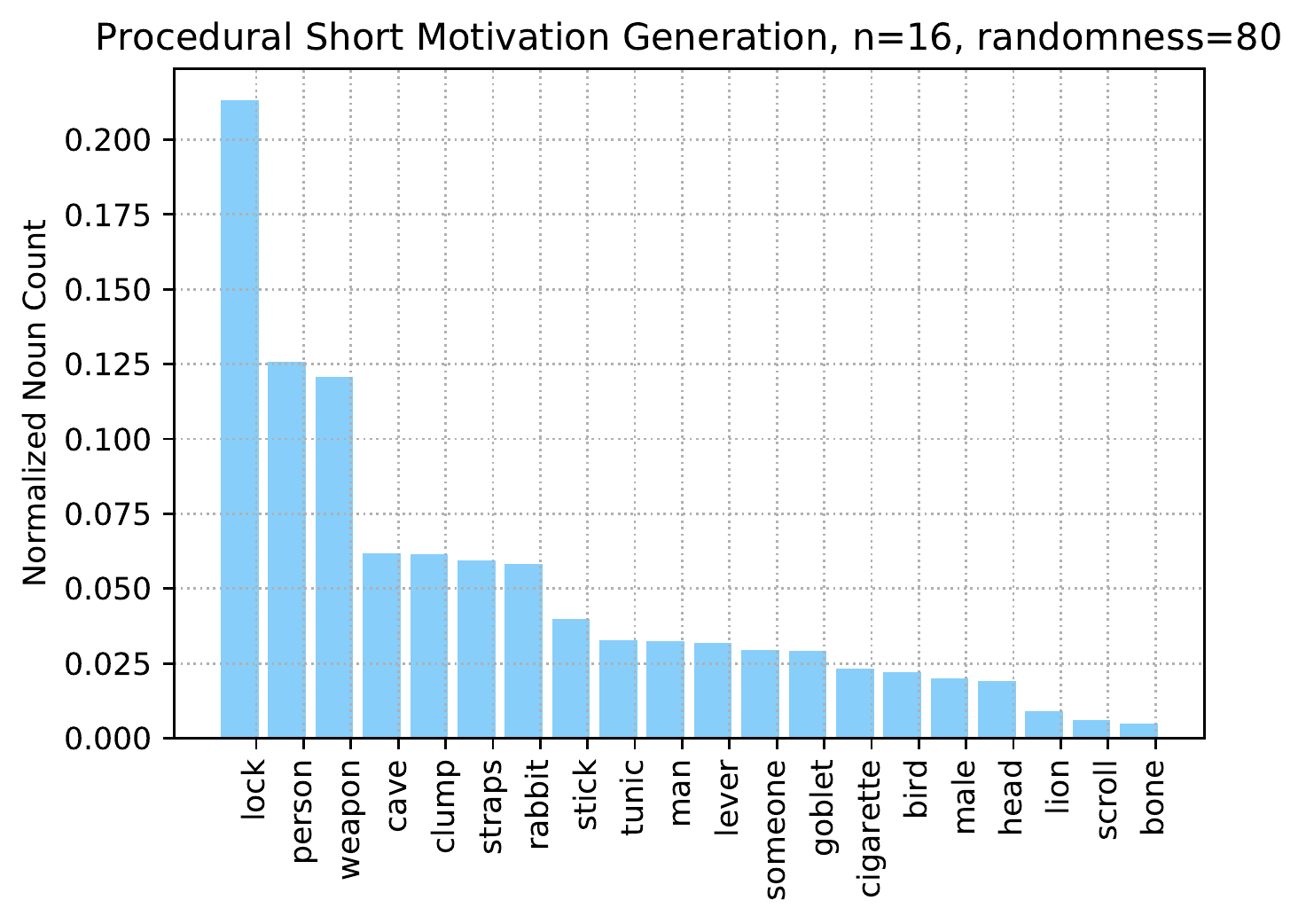}} &
\subfloat{\includegraphics[width=0.225\linewidth]{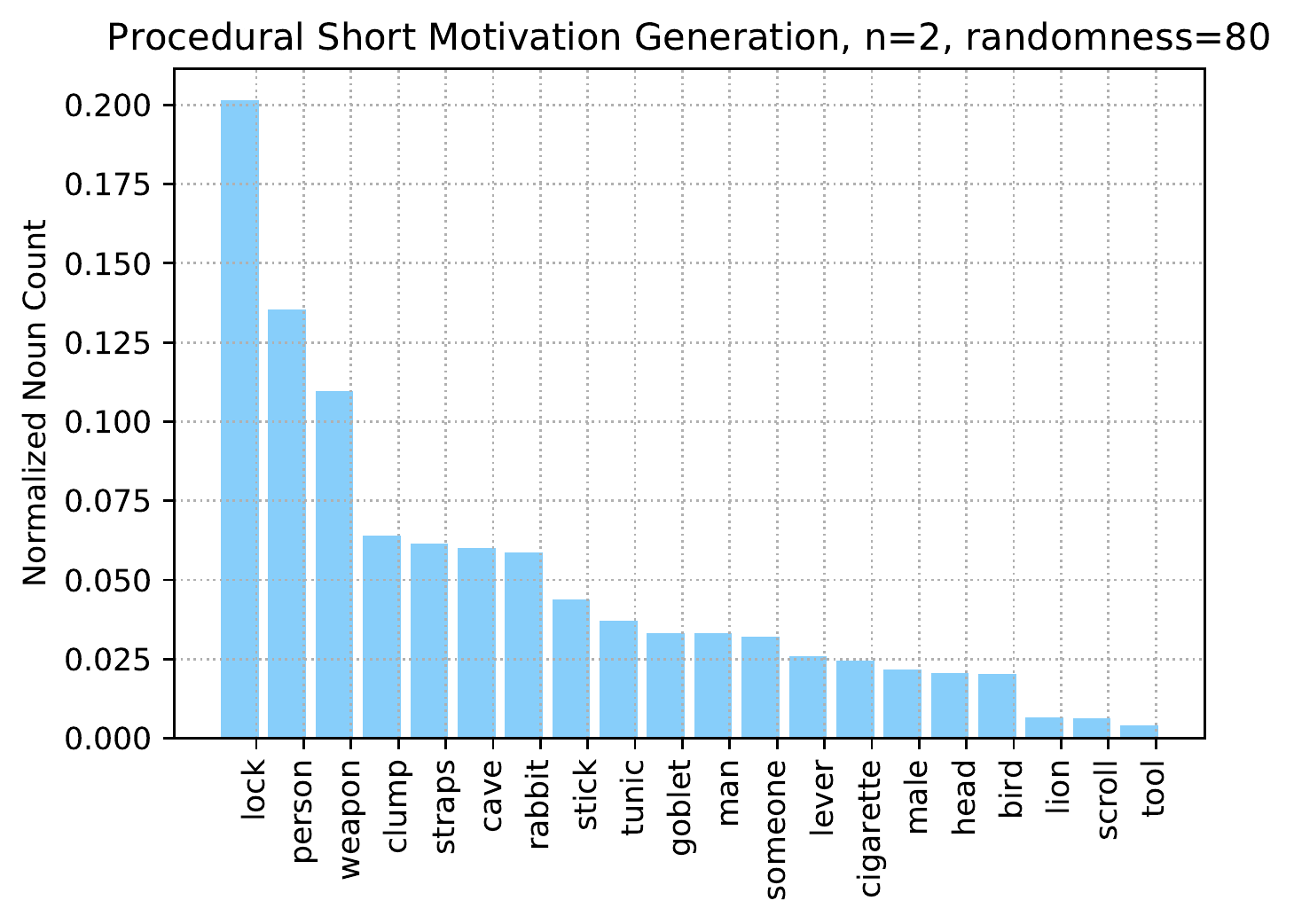}}\\
\subfloat{\includegraphics[width=0.225\linewidth]{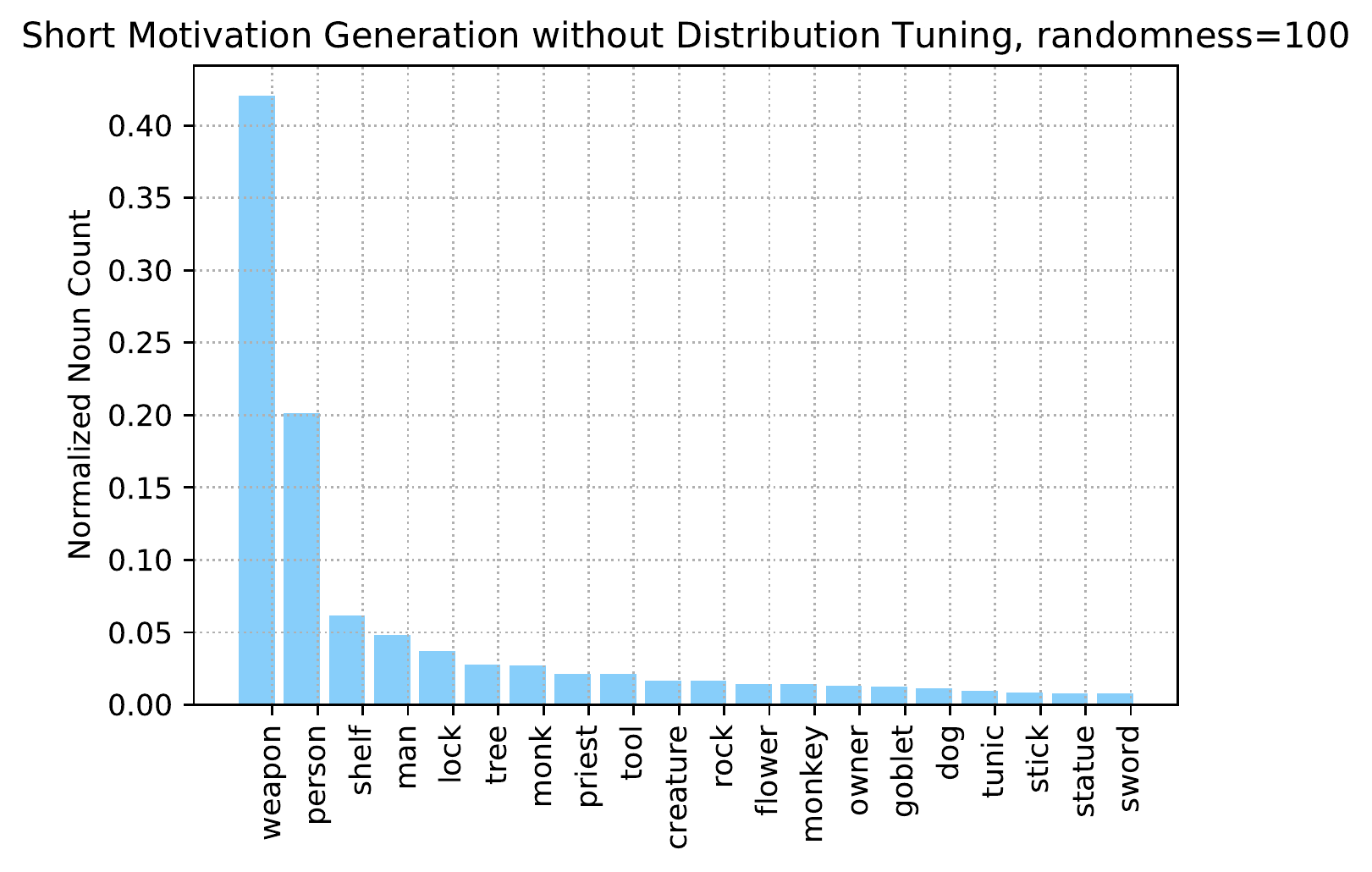}} &
\subfloat{\includegraphics[width=0.225\linewidth]{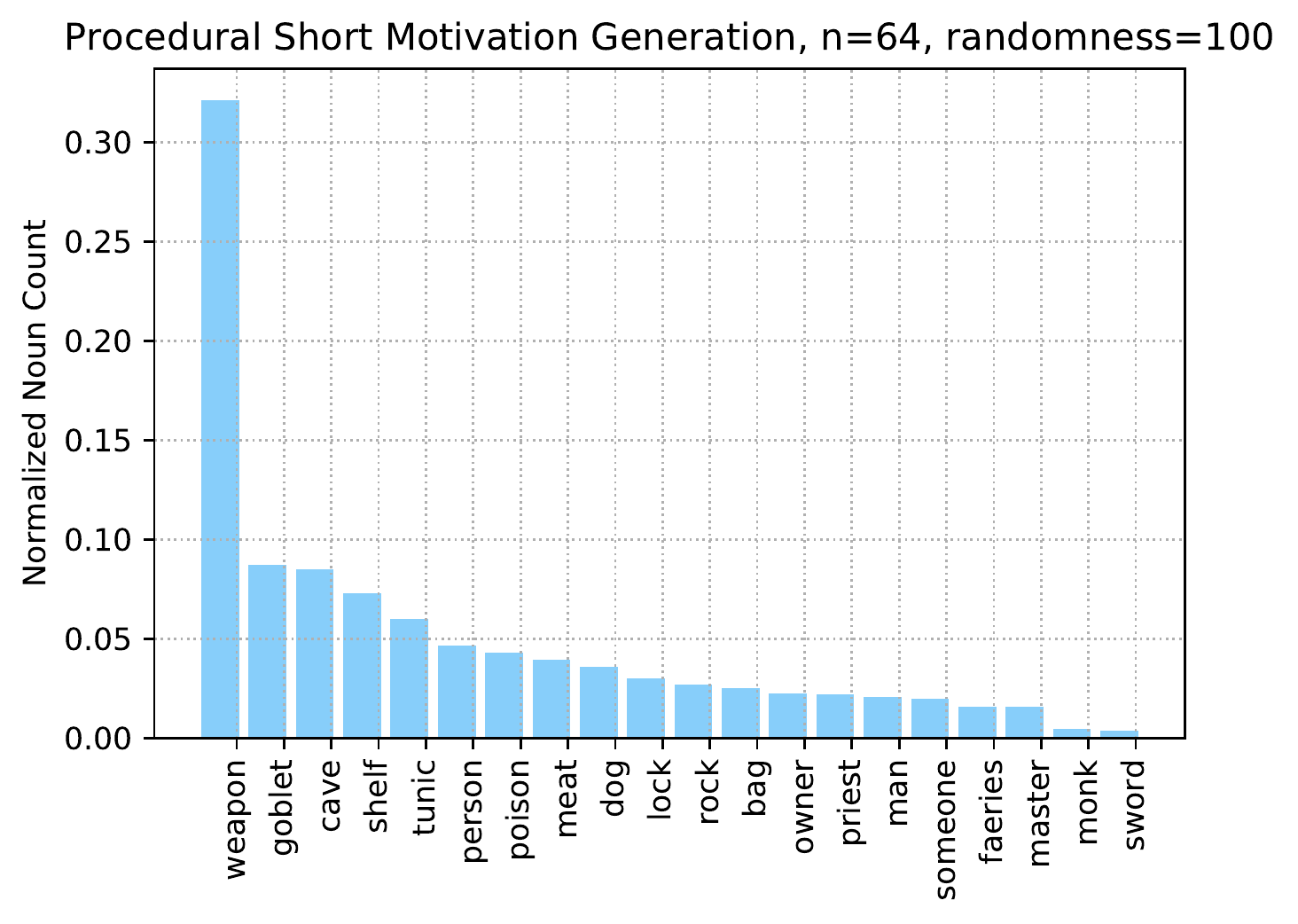}} &
\subfloat{\includegraphics[width=0.225\linewidth]{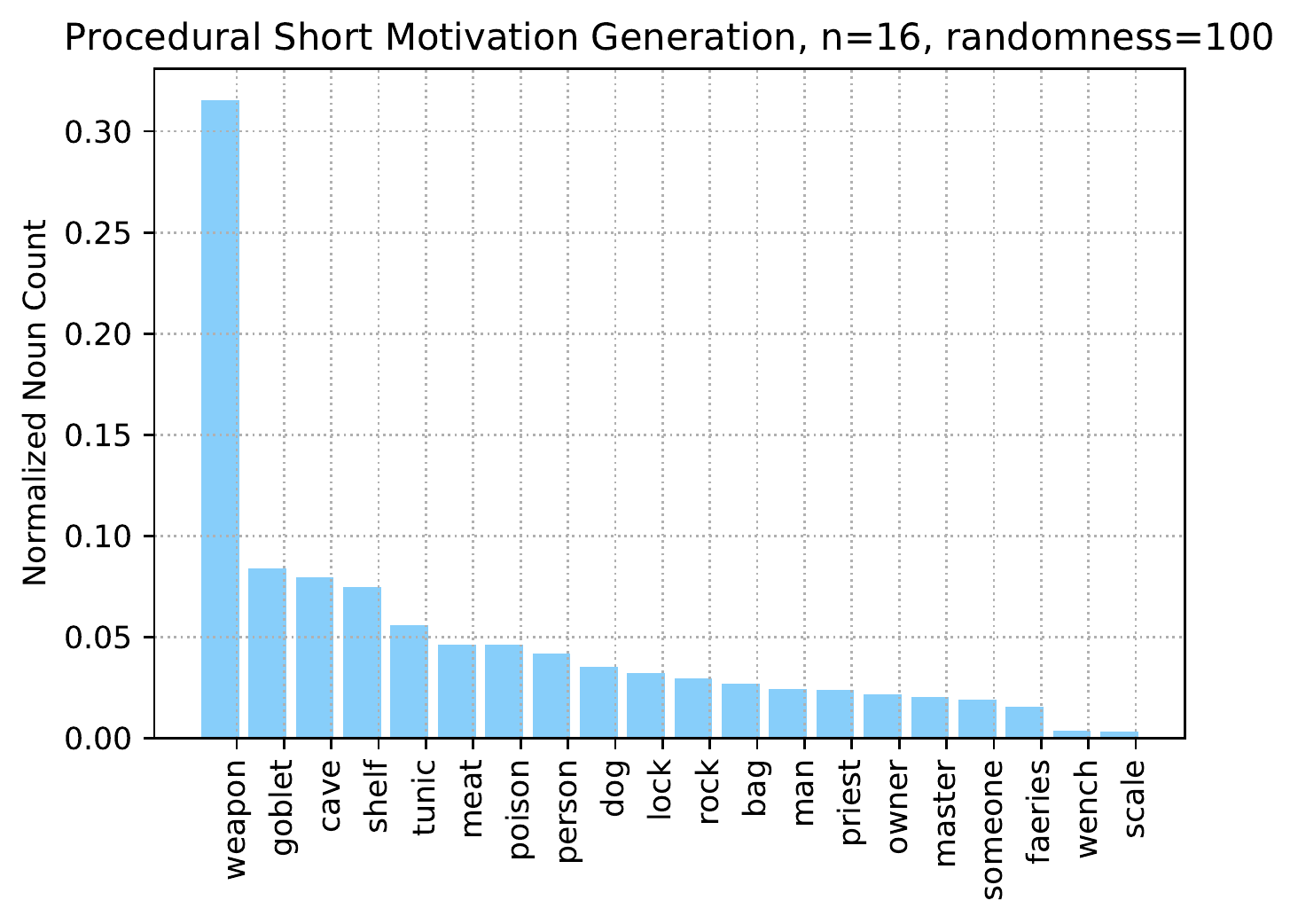}} &
\subfloat{\includegraphics[width=0.225\linewidth]{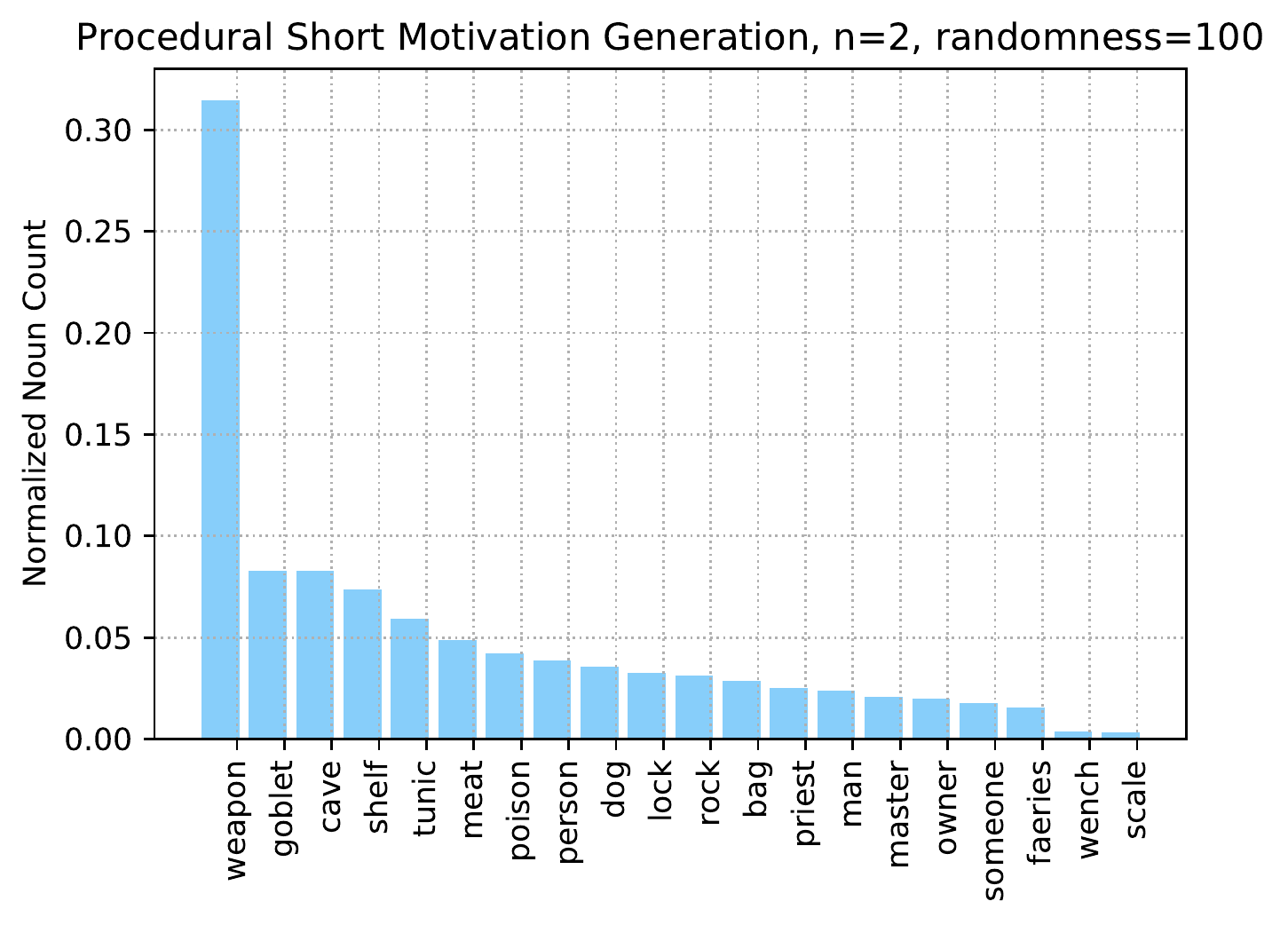}}
\end{tabular}
\caption{Distribution of nouns in the short motivation of the curriculum of quests starting from the original distribution on the left to the flattened and {\bf randomly generated} curriculum on the right as a function of $n$ (Section~\ref{sec:curriculumgen}) with the {\bf randomness percentage} tuning. The y-axis of the different nouns reflect their relative proportion in the pool of quests.}
\end{figure*}

\newpage
\clearpage

\newpage
\clearpage

\subsection{Effects of Diversity in Procedural Generation Pipeline on Curriculum Learning}
Table~\ref{app:currlearningeval} shows the results of a zero-shot evaluation as described in Section~\ref{sec:curreval} on each of the randomly generated curriculum pools.
Agents were trained on the full curriculum for each of these experiments.
One major trend stands out: the less randomness during environment generation, the greater the performance.
This shows that, while more diverse (as seen in Table~\ref{tab:curriculumhitgrams}), having potentially less coherent worlds and quests during training hurts agent performance at test time---a case of spurious diversity in training data.

\begin{table}[!h]
    \scriptsize
      \centering
        \begin{tabular}{rlll}
        \multicolumn{1}{l}{\textbf{Expt.}} & \textbf{Act Goals} & \textbf{Speech Goals} & \textbf{All Goals} \\ \hline
        \multicolumn{4}{c}{\textbf{Scratch Encoder}} \\ \hline
        \multicolumn{1}{l}{\textbf{No Curr.}} & 0.418 & 0.118 & 0.103 \\
        \multicolumn{1}{l}{\textbf{Sampled}} & 0.460 & 0.145 & 0.138 \\
\multicolumn{1}{l}{\textbf{100\% Randomly Generated}} &  0.263&	0.024&	0.017  \\
\multicolumn{1}{l}{\textbf{80\% Randomly Generated}} &  0.267&	0.080&	0.062  \\
\multicolumn{1}{l}{\textbf{60\% Randomly Generated}} &  0.379&	0.112&	0.093  \\
\multicolumn{1}{l}{\textbf{40\% Randomly Generated}} &  0.422&	0.115&	0.109  \\
\multicolumn{1}{l}{\textbf{20\% Randomly Generated}} &  0.464&	0.146&	0.143  \\
\multicolumn{1}{l}{\textbf{Procedurally Generated}} & \textbf{0.477}&	\textbf{0.163}&	\textbf{0.155}  \\
        \hline
        \multicolumn{4}{c}{\textbf{Adaptive Encoder}} \\ \hline
        \multicolumn{1}{l}{\textbf{No Curr.}} & 0.420 & 0.330 & 0.303 \\
        \multicolumn{1}{l}{\textbf{Sampled}} & 0.473 & 0.358 & 0.344 \\
        \multicolumn{1}{l}{\textbf{100\% Randomly Generated}} &  0.335&	0.221&	0.207  \\
\multicolumn{1}{l}{\textbf{80\% Randomly Generated}} &  0.364&	0.280&	0.269  \\
\multicolumn{1}{l}{\textbf{60\% Randomly Generated}} &  0.424&	0.327&	0.293  \\
\multicolumn{1}{l}{\textbf{40\% Randomly Generated}} &  0.481&	0.370&	0.330 \\
\multicolumn{1}{l}{\textbf{20\% Randomly Generated}} &  \textbf{0.508}&	0.371&	0.369  \\
\multicolumn{1}{l}{\textbf{Procedurally Generated}} &  0.506&	\textbf{0.382}&	\textbf{0.373}  \\

        \end{tabular}
        \caption{Effects of diversity in procedural generation on curriculum learning. All experiments were averaged over 3 random seeds. Standard deviations across any individual result do not exceed $0.02$. The ``All Goals'' column refers to quests where the agent has simultaneously achieved both types of goals within the allotted one episode. The parameter $n$ refers to the difference between the number of instances for the highest and lowest count quest types.}
        \label{app:currlearningeval}
\end{table}

\subsection{Curriculum Statistics}
\label{app:currstats}
This section presents statistics attempting to quantify the diversity and relative coherence of the environments in each of the curriculums we test on.
We quantify diversity in terms of the unique entities present overall in the world as well as the number of unique uni-,bi-, and tri-grams found in the generated short motivations and goal texts.

\begin{table}[]
\scriptsize
\centering
\begin{tabular}{llllll}
\hline
\multicolumn{5}{c}{\textbf{Procedural Generated Short Motivations, randomness = 0}} 
\\ 
\hline
& {\textbf{entities}} & {\textbf{hit \%age}} & {\textbf{unigrams}} & {\textbf{bigrams}} & {\textbf{trigrams}} \\
{\textbf{untuned}} & 2529 & 0.93 & 448 & 1141 & 1734 \\
{\textbf{n=64}} & 2527 & 0.91 & 446 & 1173 & 1789 \\
{\textbf{n=16}} & 2523 & 0.91 & 441 & 1139 & 1720 \\
{\textbf{n=2}}  & 2523 & 0.91 & 436 & 1146 & 1738  
\\ \hline
\multicolumn{5}{c}{\textbf{Procedural Generated Goals, randomness = 0}}    
\\ \hline
& {\textbf{entities}} & {\textbf{hit \%age}} & {\textbf{unigrams}} & {\textbf{bigrams}} & {\textbf{trigrams}} \\
{\textbf{untuned}} & 2529 & 0.93 & 955  & 5148  & 8348  \\
{\textbf{n=64}} & 2527 & 0.93 & 1061 & 5032 & 8126  \\
{\textbf{n=16}} & 2523 & 0.93 & 992 & 4749 & 7853    \\
{\textbf{n=2}}  & 2523 & 0.94  & 935 & 4594 & 7693  \\ \hline
\multicolumn{5}{c}{\textbf{Randomly Generated Short Motivations, randomness = 20}} \\ \hline
& {\textbf{entities}} & {\textbf{hit \%age}} & {\textbf{unigrams}} & {\textbf{bigrams}} & {\textbf{trigrams}} \\
{\textbf{untuned}} & 2633 & 0.10  & 389 & 1007 & 1617 \\
{\textbf{n=64}} & 2626 & 0.12 & 378  & 1013  & 1641  \\
{\textbf{n=16}} & 2607 & 0.16 & 372  & 985 & 1593  \\
{\textbf{n=2}} & 2614 & 0.17 & 349  & 917& 1475 \\ \hline
\multicolumn{5}{c}{\textbf{Randomly Generated Goals, randomness = 20}} 
\\ \hline
& {\textbf{entities}} & {\textbf{hit \%age}} & {\textbf{unigrams}} & {\textbf{bigrams}} & {\textbf{trigrams}} \\
{\textbf{untuned}} & 2633 & 0.17  & 846 & 3061 & 5824  \\
{\textbf{n=64}}  & 2626 & 0.15 & 919  & 3450  & 6530  \\
{\textbf{n=16}}  & 2607 & 0.17 & 827  & 3311   & 6422 \\
{\textbf{n=2}}  & 2614 & 0.17  & 724 & 2998  & 5926   \\ 
\hline
\multicolumn{5}{c}{\textbf{Randomly Generated Short Motivations, randomness = 40}}  \\ \hline
& {\textbf{entities}} & {\textbf{hit \%age}} & {\textbf{unigrams}} & {\textbf{bigrams}} & {\textbf{trigrams}} \\
{\textbf{untuned}} & 2604 & 0.37 & 478   & 1239 & 1968  \\
{\textbf{n=64}} & 2590 & 0.21  & 762 & 1695 & 2458 \\
{\textbf{n=16}} & 2586 & 0.61  & 490 & 1251 & 1984 \\
{\textbf{n=2}} & 2584 & 0.60 & 476 & 1237 & 1972 \\ \hline
\multicolumn{5}{c}{\textbf{Randomly Generated Goals, randomness = 40}} \\ \hline
& {\textbf{entities}} & {\textbf{hit \%age}} & {\textbf{unigrams}} & {\textbf{bigrams}} & {\textbf{trigrams}} \\
{\textbf{untuned}} & 2604 & 0.13  & 837 & 4302  & 7444   \\
{\textbf{n=64}} & 2590 & 0.12 & 970 & 4870 & 7750 \\
{\textbf{n=16}} & 2586 & 0.37 & 901 & 4617 & 7551 \\
{\textbf{n=2}} & 2584 & 0.36  & 879 & 4643 & 7570 \\ 
\hline
\multicolumn{5}{c}{\textbf{Randomly Generated Short Motivations, randomness = 60}}  
\\ \hline
& {\textbf{entities}} & {\textbf{hit \%age}} & {\textbf{unigrams}} & {\textbf{bigrams}} & {\textbf{trigrams}} \\
{\textbf{untuned}} & 2582 & 0.10 & 346 & 831 & 1262 \\
{\textbf{n=64}} & 2578 & 0.27 & 383 & 910 & 1384 \\
{\textbf{n=16}} & 2576 & 0.31 & 390 & 920 & 1395 \\
{\textbf{n=2}}  & 2573 & 0.31 & 378 & 893 & 1356 \\ \hline
\multicolumn{5}{c}{\textbf{Randomly Generated Goals, randomness = 60}}   \\ \hline
& {\textbf{entities}} & {\textbf{hit \%age}} & {\textbf{unigrams}} & {\textbf{bigrams}} & {\textbf{trigrams}} \\
{\textbf{untuned}} & 2582 & 0.09  & 468 & 1565  & 3054 \\
{\textbf{n=64}} & 2578 & 0.27 & 612 & 2862 & 5549 \\
{\textbf{n=16}} & 2576 & 0.30 & 571 & 2834 & 5612 \\
{\textbf{n=2}} & 2573 & 0.31 & 556 & 2842 & 5631   \\ \hline
\multicolumn{5}{c}{\textbf{Randomly Generated Short Motivations, randomness = 80}} \\ \hline
& {\textbf{entities}} & {\textbf{hit \%age}} & {\textbf{unigrams}} & {\textbf{bigrams}} & {\textbf{trigrams}} \\
{\textbf{untuned}} & 2541 & 0.08 & 409 & 1046 & 1636 \\
{\textbf{n=64}} & 2541 & 0.17 & 409 & 1110 & 1771 \\
{\textbf{n=16}} & 2540 & 0.19 & 406 & 1075 & 1710 \\
{\textbf{n=2}} & 2540 & 0.18 & 402 & 1070 & 1691  \\ \hline
\multicolumn{5}{c}{\textbf{Randomly Generated Goals, randomness = 80}} \\ \hline
& {\textbf{entities}} & {\textbf{hit \%age}} & {\textbf{unigrams}} & {\textbf{bigrams}} & {\textbf{trigrams}} \\
{\textbf{untuned}} & 2541 & 0.11 & 516 & 2781 & 5804 \\
{\textbf{n=64}} & 2541 & 0.26 & 786 & 4171 & 7534  \\
{\textbf{n=16}} & 2540 & 0.28 & 757 & 4153 & 7576  \\
{\textbf{n=2}}  & 2540 & 0.28  & 719 & 3979 & 7372  \\ \hline
\multicolumn{5}{c}{\textbf{Randomly Generated Short Motivations, randomness = 100}} \\ \hline
& {\textbf{entities}} & {\textbf{hit \%age}} & {\textbf{unigrams}} & {\textbf{bigrams}} & {\textbf{trigrams}} \\
{\textbf{untuned}} & 2537 & 0.11 & 321  & 779  & 1204  \\
{\textbf{n=64}} & 2537 & 0.30 & 321 & 765 & 1166 \\
{\textbf{n=16}} & 2527 & 0.29 & 314 & 744 & 1141 \\
{\textbf{n=2}} & 2527 & 0.30 & 314 & 739 & 1127  \\ \hline
\multicolumn{5}{c}{\textbf{Randomly Generated Goals, randomness = 100}} \\ \hline
& {\textbf{entities}} & {\textbf{hit \%age}} & {\textbf{unigrams}} & {\textbf{bigrams}} & {\textbf{trigrams}} \\
{\textbf{untuned}} & 2537 & 0.07 & 397  & 2363 & 5232 \\
{\textbf{n=64}} & 2537 & 0.13 & 477 & 3156  & 6263 \\
{\textbf{n=16}} & 2527 & 0.13 & 434 & 3039 & 6154  \\
{\textbf{n=2}}  & 2527 & 0.14  & 427 & 2993  & 6114 
\end{tabular}
\caption{Curriculum learning hit analysis and unique n-grams counts. The tables show the hit percentage of the procedually generated entities in short motivations/ goals among the retrieved entities (objects + character).The count of unique uni-grams /bi-grams/ tri-grams represent the n-grams counts changing with the {\bf procedurally generated} curriculum as a function of $n$ (Section~\ref{sec:curriculumgen}) with the {\bf randomness percentage} tuning for the generated short motivations or goals using BART model.}
\label{tab:curriculumhitgrams}
\end{table}

Specifically, unique entities were calculated by using the count of all the unique objects and character which are generated in the procedural generated short motivations / goals. 
In addition, the count of the unique uni-grams /bi-grams /tri-grams represent the n-grams counts changing with the {\bf procedurally generated} curriculum as a function of $n$ (Section~\ref{sec:curriculumgen}) with the {\bf randomness percentage} tuning for both the short motivations and goals generated by BART.
As a sanity check on how coherent an environment is, we attempt to see if the entities required to finish a quest even exist within the world---i.e. a hit percentage that roughly estimates what proportion of quests in a pool are achievable end to end.
The hit percentage are calculated by checking if the \emph{NOUN} extracted from the short motivations/goals exists in the procedually generated entities (objects + character) in the same environment. 
Counting as 1/0 to represent as existing/not and divided by the total number of quests to get the hit percentage in the table.

\subsection{Hyperparameters}
\label{app:lightprocegenhyper}

\begin{table}[h]
\scriptsize

\centering
\begin{tabular}{lll} \toprule
Hyperparameter type & Value  \\ \midrule
Num. layers & 2 \\
Num. attention heads & 2 \\
Embedding size & 300 \\
Dropout ratio & 0.0 \\
Gradient clip & 0.1  \\
Optimizer & Adam  \\
Learning rate & \num{1e-4} \\ \bottomrule
\end{tabular}
\caption{Hyperparameters used to train the Biencoder model to retrieve objects for generating the LIGHT world. The same trained models were then frozen and used for further experiments.}
\label{tab:para_objects}
\end{table}

\begin{table}[h]
\footnotesize

\centering
\begin{tabular}{lll} \toprule
Hyperparameter type & Value  \\ \midrule
Embedding size & 128 \\
Embedding norm & 10 \\
Dropout ratio & 0.0 \\
Gradient clip & 0.1  \\
Optimizer & SGD  \\
Learning rate & 0.1 \\ \bottomrule
\end{tabular}
\caption{Hyperparameters used to train the Starspace model to retrieve character for generating the LIGHT world.}
\label{tab:para_character}
\end{table}

\begin{table}[]
\footnotesize
\centering
\begin{tabular}{lll} \toprule
Hyperparameter type & Value  \\ \midrule
Num. encoder layers & 12 \\
Num. decoder layers & 12 \\
Num. attention heads & 16 \\
Batchsize & 8 \\
Activation & gelu \\
Beam size & 1 \\
Beam decay & 30 \\
Beam length penalty & 0.65 \\
Num. attention heads & 2 \\
Embedding size & 1024 \\
Dropout ratio & 0.1 \\
Gradient clip & 0.1  \\
Optimizer & SGD  \\
Learning rate & \num{1e-4} \\ \bottomrule
\end{tabular}
\caption{Hyperparameters used to train and test the BART model for generating short motivations and goals.}
\label{tab:para_short_mot}
\end{table}

\begin{table}[h]
\centering
\footnotesize
\begin{tabular}{l|l} \toprule
\multicolumn{1}{c}{{Hyperparameter type}} & \multicolumn{1}{c}{{Value}} \\ \midrule
Dictionary Tokenizer                             & Byte-pair encoding                 \\
Num. layers                                      & 12                                 \\
Num. attention heads                             & 12                                 \\
Feedforward network hidden size                  & 3072                               \\
Input length                                     & 1024                               \\
Embedding size                                   & 768                                \\
Batch size                                       & 32                                 \\
Dropout ratio                                    & 0.1                                \\
Poly-n-codes                                     & 64                                 \\
Gradient clip                                    & 1.0                                \\
Optimizer                                        & Adam                               \\
Learning rate                                    &   $\num{1e-6}$                                 \\ \bottomrule
\end{tabular}
\caption{Hyperparameters used to pre-train the adaptive encoder as described in \citet{Humeau2020}.}
\label{tab:suphyper}
\end{table}

\begin{table}[h]
\centering
\footnotesize
\begin{tabular}{l|l} \toprule
\multicolumn{1}{c}{{Hyperparameter type}} & \multicolumn{1}{c}{{Value}} \\ \midrule
\multicolumn{2}{l}{General}                      \\ \hline
Discount $\gamma$                      & 0.99           \\
Valid Action loss coefficient   & 10             \\
Action entropy coefficient      & 0.01           \\
Valid Speech loss coefficient   & 40             \\
Speech entropy coefficient      & 0.04           \\
Batch size                      & 32             \\
Gradient clip                   & 1.0            \\
Steps per episode               & 100            \\ \hline
Policy Networks (Actors)        &                \\ \hline
Num. Layers                     & 3              \\
Feedforward network hidden size & 768            \\
GRU hidden size                 & 768            \\ \hline
\multicolumn{2}{l}{Value Predictor (Critic)}     \\ \hline
Num. Layers                     & 2              \\
Feedforward network hidden size & 768            \\ \hline
\multicolumn{2}{l}{Appended Encoder}             \\ \hline
Num. layers                     & 3              \\
Num. attention heads            & 3              \\
Feedforward network hidden size & 768    \\ \bottomrule
\end{tabular}
\caption{RL experiments hyperparameters unchanged from \citet{Ammanabrolu2021}.}
\label{tab:rlhyper}
\end{table}

\end{document}